\documentclass[acmsmall,table,xcdraw]{acmart}

\usepackage{rotating}
\usepackage{makecell}
\usepackage{multirow}
\usepackage{amsmath,amssymb,amsfonts}
\usepackage{algorithmic}
\usepackage{graphicx}
\usepackage{lscape}
\usepackage{afterpage}
\usepackage{textcomp}
\usepackage{array}
\usepackage{url}
\usepackage{enumitem}
\usepackage{xcolor}
\usepackage{booktabs}
\usepackage{lscape}
\usepackage{anyfontsize}
\usepackage{caption}
\usepackage{subcaption}
\usepackage{natbib}

\usepackage[subtle,tracking=normal]{savetrees}
\definecolor{lightgray}{rgb}{0.929, 0.929, 0.929}

\newcommand{\blue}[1]{{\color{black}{#1}}}

\setcopyright{acmcopyright}
\copyrightyear{2020}
\acmYear{2020}
\acmDOI{XX.XXXX/XXXXXXX.XXXXXXX}

\acmJournal{CSUR}

\begin{document}
	
	\title{The Creation and Detection of Deepfakes: A Survey}
	
	\author{Yisroel Mirsky}\authornote{Corresponding Author}
	\email{yisroel@gatech.edu}	\email{yisroel@post.bgu.ac.il}
	\affiliation{%
	\institution{Georgia Institute of Technology}
	\streetaddress{756 W Peachtree St NW}
	\city{Atlanta}
	\state{Georgia}
	\postcode{30308}
	}
	\affiliation{%
	\institution{Ben-Gurion University}
	\streetaddress{P.O.B. 653}
	\city{Beer-Sheva}
	\state{Israel}
	\postcode{8410501}
	}
	\author{Wenke Lee}
	\email{wenke@cc.gatech.edu}
	\affiliation{%
		\institution{Georgia Institute of Technology}
		\streetaddress{756 W Peachtree St NW}
		\city{Atlanta}
		\state{Georgia}
		\postcode{30308}
	}

	\renewcommand{\shortauthors}{Mirsky, et al.}
	
	\begin{abstract}
		Generative deep learning algorithms have progressed to a point where it is difficult to tell the difference between what is real and what is fake. In 2018, it was discovered how easy it is to use this technology for unethical and malicious applications, such as the spread of misinformation, impersonation of political leaders, and the defamation of innocent individuals. Since then, these `deepfakes' have advanced significantly. 

		In this paper, we explore the creation and detection of deepfakes an provide an in-depth view how these architectures work. The purpose of this survey is to provide the reader with a deeper understanding of (1) how deepfakes are created and detected, (2) the current trends and advancements in this domain, (3) the shortcomings of the current defense solutions, and (4) the areas which require further research and attention.
	\end{abstract}

	\ccsdesc[500]{Security and privacy~Social engineering attacks}
	\ccsdesc[500]{Computing methodologies~Machine learning}
	\ccsdesc[300]{Security and privacy~Human and societal aspects of security and privacy}

	\keywords{Deepfake, Deep fake, reenactment, replacement, face swap, generative AI, social engineering, impersonation}
	
	\maketitle
	
	\section{Introduction}

A deepfake is content, generated by an artificial intelligence, that is authentic in the eyes of a human being. The word \textit{deepfake} is a combination of the words \textit{`deep learning'} and \textit{`fake'} and primarily relates to content generated by an artificial neural network, a branch of machine learning.

The most common form of deepfakes involve the generation and manipulation of human imagery. 	
This technology has creative and productive applications. For example, realistic video dubbing of foreign films,\footnote{https://variety.com/2019/biz/news/\\ai-dubbing-david-beckham-multilingual-1203309213/} education though the reanimation of historical figures \cite{Deepfake69:online}, and virtually trying on clothes while shopping.\footnote{\url{https://www.forbes.com/sites/forbestechcouncil/2019/05/21/gans-and-deepfakes-could-revolutionize-the-fashion-industry/}} There are also numerous online communities devoted to creating deepfake memes for entertainment,\footnote{https://www.reddit.com/r/SFWdeepfakes/} such as music videos portraying the face of actor Nicolas Cage.

However, despite the positive applications of deepfakes, the technology is infamous for its unethical and malicious aspects. 
At the end of 2017, a Reddit user by the name of `deepfakes' was using deep learning to swap faces of celebrities into pornographic videos, and was posting them online\footnote{https://www.vice.com/en\_us/article/gydydm/gal-gadot-fake-ai-porn}.
The discovery caused a media frenzy and a large number of new deepfake videos began to emerge thereafter. In 2018, BuzzFeed released a deepfake video of former president Barak Obama giving a talk on the subject. The video was made using the Reddit user's software (FakeApp), and raised concerns over identity theft, impersonation, and the spread of misinformation on social media. Fig. presents an information trust chart for deepfakes, inspired by \cite{Thebigge1:online}.
\looseness=-1

Following these events, the subject of deepfakes gained traction in the academic community, and the technology has been rapidly advancing over the last few years. Since 2017, the number of papers published on the subject rose from 3 to over 250 (2018-20).

To understand where the threats are moving and how to mitigate them, we need a clear view of the technology's, challenges, limitations, capabilities, and trajectory. 
Unfortunately, to the best of our knowledge, there are no other works which present the techniques, advancements, and challenges, in a technical and encompassing way. 
Therefore, the goals of this paper are (1) to provide the reader with an understanding of how modern deepfakes are created and detected, (2) to inform the reader of the recent advances, trends, and challenges in deepfake research, (3) to serve as a guide to the design of deepfake architectures, and (4) to identify the current status of the attacker-defender game, the attacker's next move, and future work that may help give the defender a leading edge.

\begin{figure}[t]
\centering
\begin{minipage}[b]{.6\textwidth}
	\centering
	\includegraphics[width=\textwidth]{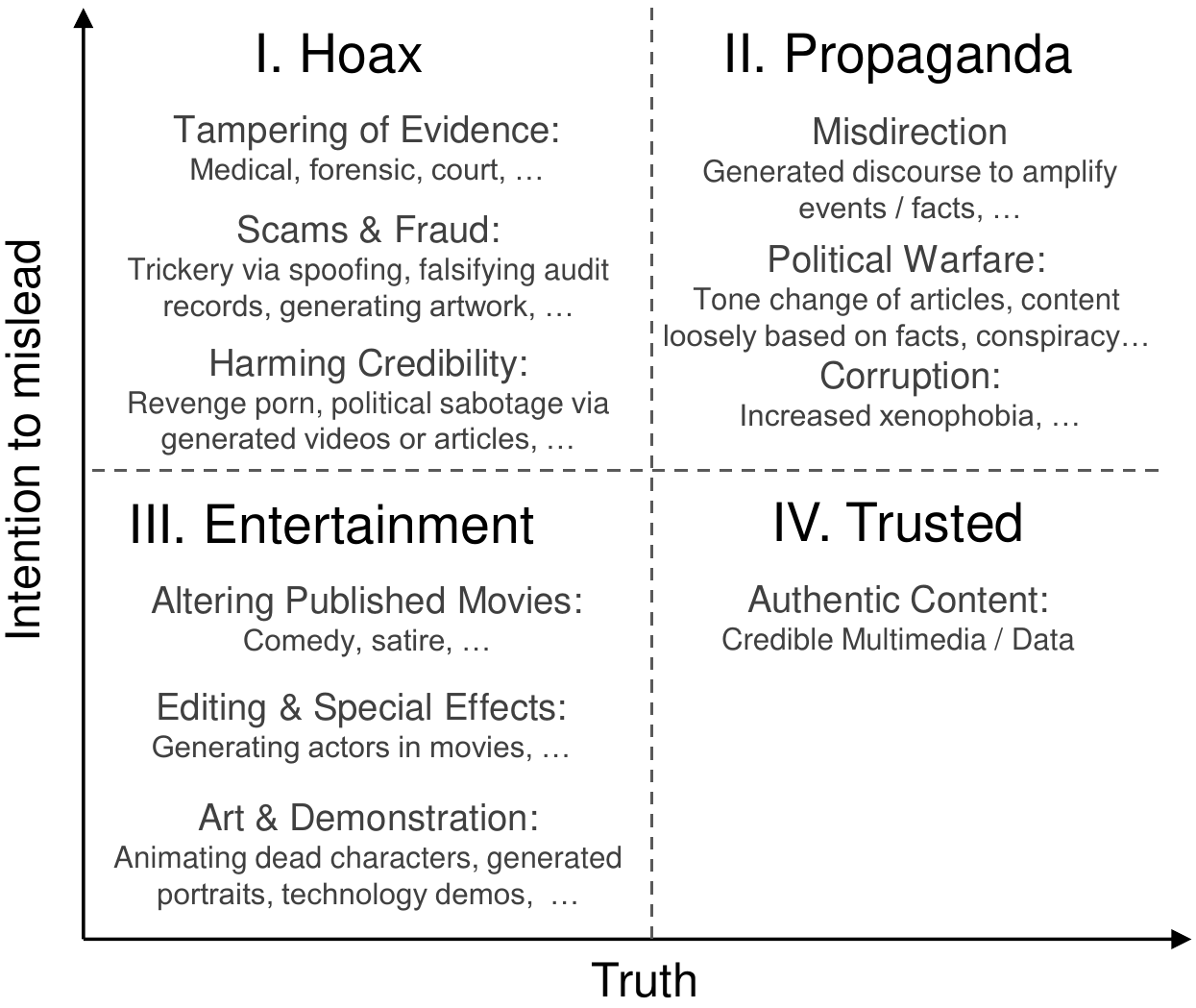}
	\caption{A deepfake information trust chart.}\label{fig:infochart}
\end{minipage}
\hfill
\begin{minipage}[b]{.39\textwidth}
	\centering
	\includegraphics[width=.88\textwidth]{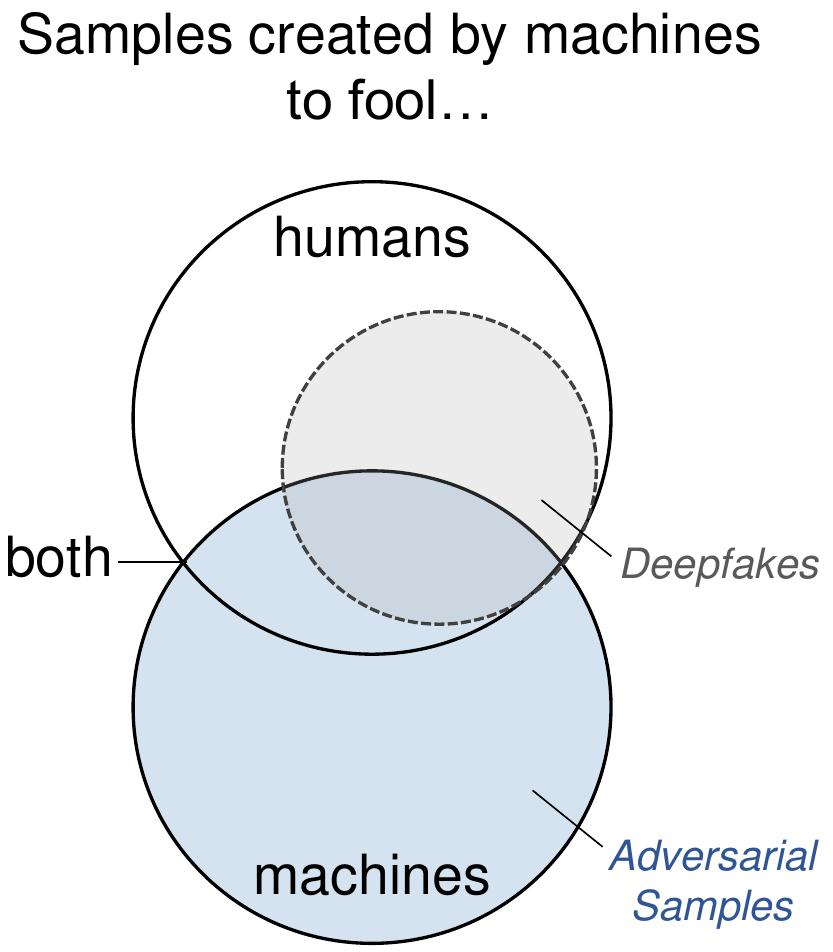}
	\begin{flushleft}
	\vspace{-.5em}
	\scriptsize 
	Examples:\\
	...\textbf{humans}: entertainment, impersonation, art fraud. \\
	...\textbf{machines}: hiding a stop sign, evading face recog.\\
	...\textbf{both}: tampering medical scans, malware evasion.
	\end{flushleft}
	\caption{The difference between \textit{adversarial machine learning} and \textit{deepfakes}.}\label{fig:adv_df}
\end{minipage}
\vspace{-2em}
\end{figure}

We achieve these goals through an overview of human visual deepfakes (Section \ref{sec:overview}), followed by a technical background which identifies technology's basic building blocks and challenges (Section \ref{sec:background}). We then provide a chronological and systematic review for each category of deepfake, and provide the networks' schematics to give the reader a deeper understanding of the various approaches (Sections \ref{sec:reen} and \ref{sec:repl}). Finally, after reviewing the countermeasures (Section \ref{sec:countermeasures}), we discuss their weaknesses, note the current limitations of deepfakes, suggest alternative research, consider the adversary's next steps, and raise awareness to the spread of deepfakes to other domains (Section \ref{sec:discussion}).   

\emph{Scope.} In this survey we will focus on deepfakes pertaining to the human face and body. We will not be discussing the synthesis of new faces or the editing of facial features because they do not have a clear attack goal associated with them. In Section \ref{subsec:df_otherdomains} we will discuss deepfakes with a much broader scope, note the future trends, and exemplify how deepfakes have spread to other domains and media such as forensics, finance, and healthcare. 

We note to the reader that deepfakes should not be confused with adversarial machine learning, which is the subject of fooling machine learning algorithms with maliciously crafted inputs (Fig. \ref{fig:adv_df}). The difference being that for deepfakes, the objective of the generated content is to fool a human and not a machine.

	\section{Overview \& Attack Models}\label{sec:overview}

We define a deepfake as 
\vspace{-.2em}
\begin{center}
\fcolorbox{lightgray}{lightgray}{\textit{``Believable media generated by a deep neural network''}}
\end{center}
\vspace{-.2em}

\begin{figure}[b]
	\centering
	\includegraphics[width=\textwidth]{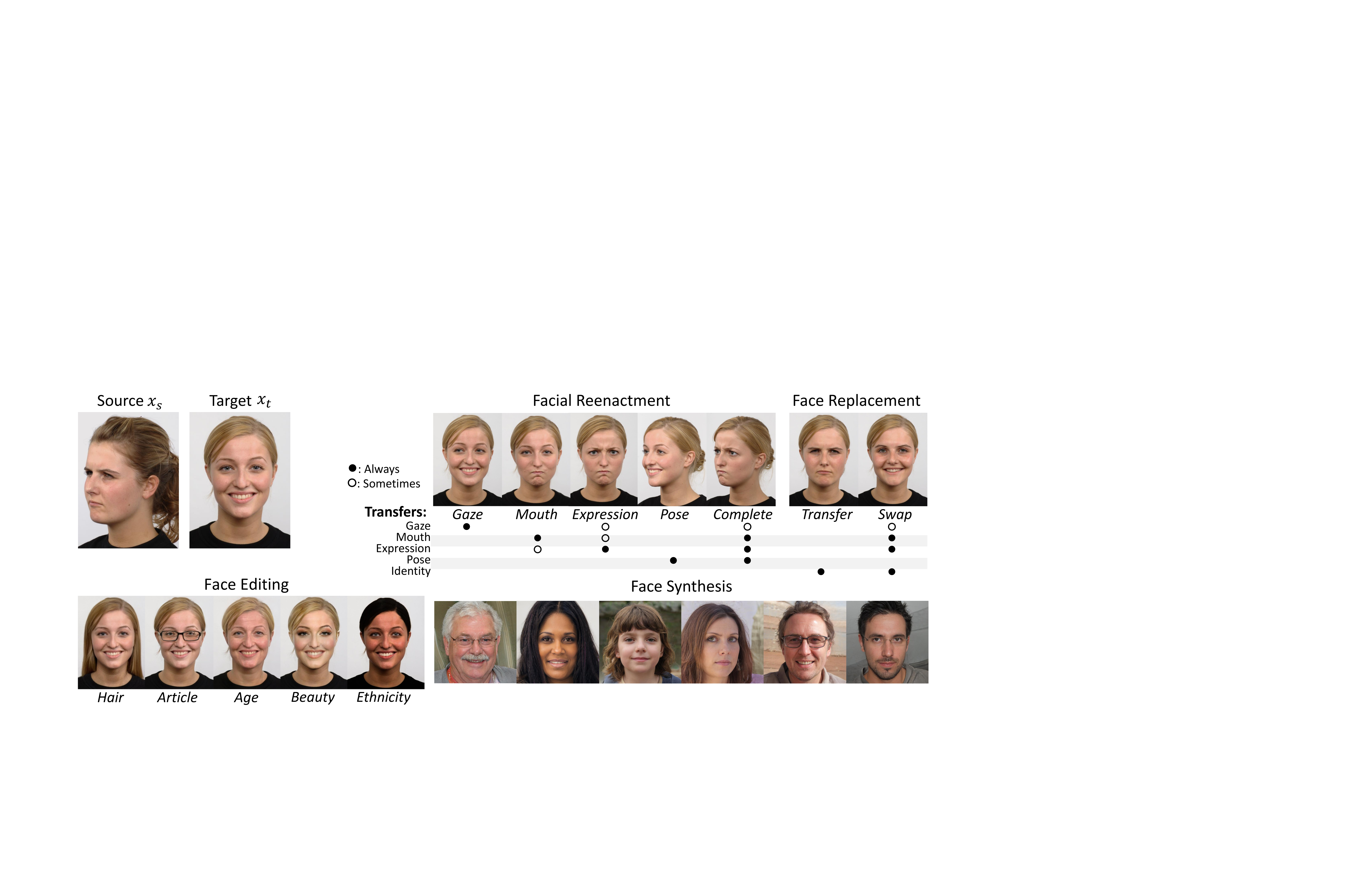}
	\vspace{-2em}
	\caption{Examples of reenactment, replacement, editing, and synthesis deepfakes of the human face.}
	\label{fig:examples}
	\vspace{-1em}
\end{figure}

\noindent In the context of human visuals, we identify four categories: reenactment, replacement, editing, and synthesis. Fig. \ref{fig:examples} illustrates some examples facial deepfakes in each of these categories and their sub-types.
		Throughout this paper we denote $s$ and $t$ as the source and the target identities. We also denote $x_s$ and $x_t$ as images of these identities and $x_g$ as the deepfake generated from $s$ and $t$.

	\subsection{Reenactment}
	
	A reenactment deepfake is where $x_s$ is used to drive the expression, mouth, gaze, pose, or body of $x_t$:\looseness=-1

	\begin{description} 
		\item[Expression] reenactment is where $x_s$ drives the expression of $x_t$. It is the most common form of reenactment since these technologies often drive target's mouth and pose as well, providing a wide range of flexibility. Benign uses are found in the movie and video game industry where the performances of actors are tweaked in post, and in educational media where historical figures are reenacted. 
		\item[Mouth] reenactment, also known as `dubbing', is where the mouth of $x_t$ is driven by that of $x_s$, or an audio input $a_s$ containing speech. Benign uses of the technology includes realistic voice dubbing into another language and editing.
		\item[Gaze] reenactment is where direction of $x_t$'s eyes, and the position of the eyelids, are driven by those of $x_s$. This is used to improve photographs or to automatically maintain eye contact during video interviews \cite{kononenko2017photorealistic}.
		\item[Pose] reenactment is where the head position of $x_t$ is driven by $x_s$. This technology has primarily been used for face frontalization of individuals in security footage, and as a means for improving facial recognition software \cite{tran2018representation}.	
		\item[Body] reenactment, a.k.a. pose transfer and human pose synthesis, is similar to the facial reenactments listed above except that's its the pose of $x_t$'s body being driven.  
	\end{description}

	\noindent\textbf{The Attack Model.} Reenactment deep fakes give attackers the ability to impersonate an identity, controlling what he or she says or does. This enables an attacker to perform acts of defamation, cause discredability, spread misinformation, and tamper with evidence. For example, an attacker can impersonate $t$ to gain trust the of a colleague, friend, or family member as a means to gain access to money, network infrastructure, or some other asset. An attacker can also generate embarrassing content of $t$ for blackmailing purposes or generate content to affect the public's opinion of an individual or political leader. The technology can also be used to tamper surveillance footage or some other archival imagery in an attempt to plant false evidence in a trial. Finally, the attack can either take place online (e.g., impersonating someone in a \textit{real-time} conversation) or offline (e.g., fake media spread on the Internet).

	\subsection{Replacement}
	A replacement deepfake is where the content of $x_t$ is replaced with that of $x_s$, preserving the identity of $s$. \looseness=-1
	\begin{description}
		\item[Transfer] is where the content of $x_t$ is replaced with that of $x_s$. A common type of transfer is facial transfer, used in the fashion industry to visualize an individual in different outfits.\looseness=-1
		\item[Swap] is where the content transferred to $x_t$ from $x_s$ is driven by $x_t$. The most popular type of swap replacement is `face swap', often used to generate memes or satirical content by swapping the identity of an actor with that of a famous individual. Another benign use for face swapping includes the anonymization of one's identity in public content in-place of blurring or pixelation.
	\end{description}

	\noindent\textbf{The Attack Model.} Replacement deepfakes are well-known for their harmful applications. For example, revenge porn is where an attacker swaps a victim's face onto the body of a porn actress to humiliate, defame, and blackmail the victim. Face replacement can also be used as a short-cut to fully reenacting $t$ by transferring  $t$'s face onto the body of a look-alike. This approach has been used as a tool for disseminating political opinions in the past \cite{Youthoug41:online}.
	
	\subsection{Editing \& Synthesis}
	An enchantment deepfake is where the attributes of $x_t$ are added, altered, or removed. Some examples include the changing a target's clothes, facial hair, age, weight, beauty, and ethnicity. Apps such as FaceApp enable users to alter their appearance for entertainment and easy editing of multimedia. The same process can be used by and attacker to build a false persona for misleading others. For example, a sick leader can be made to look healthy \cite{Thebigge0:online}, and child or sex predators can change their age and gender to build dynamic profiles online. A known unethical use of editing deepfakes is the removal of a victim's clothes for humiliation or entertainment \cite{Youthoug42:online}.

	Synthesis is where the deepfake $x_g$ is created with no target as a basis. Human face and body synthesis techniques such as \cite{karras2019analyzing} (used in Fig. \ref{fig:examples}) can create royalty free stock footage or generate characters for movies and games. However, similar to editing deepfakes, it can also be used to create fake personas online. 
	
	 Although human image editing and synthesis are active research topics, reenactment and replacement deepfakes are the greatest concern  because they give an attacker control over one's identity\cite{hall2018deepfake,chesney2018deep,antinori2019terrorism}.  Therefore, in this survey we will be focusing on reenactment and replacement deepfakes.

	
	
	
	\section{Technical background}\label{sec:background}

	Although there are a wide variety of neural networks, most deepfakes are created using variations or combinations of generative networks and encoder decoder networks. In this section we provide a brief introduction to these networks, how they are trained, and the notations which we will be using throughout the paper.
	
	\subsection{Neural Networks}
	Neural networks are non-linear models for predicting or generating content based on an input. They are made up of layers of neurons, where each layer is connected sequentially via synapses. The synapses have associated weights which collectively define the concepts learned by the model. To execute a network on an $n$-dimensional input $x$, a process known as \textit{forward-propagation} is performed where $x$ propagated through each layer and an activation function is used to summarize a neuron's output (e.g., the Sigmoid or ReLU function).
	
	Concretely, let $l^{(i)}$ denote the $i$-th layer in the network $M$, and let $\|l^{(i)}\|$ denote the number of neurons in $l^{(i)}$. Finally, let the total number of layers in $M$ be denoted as $L$. The weights which connect $l^{(i)}$ to $l^{(i+1)}$ are denoted as the $\|l^{(i)}\|$-by-$\|l^{(i+1)}\|$ matrix $W^{(i)}$ and $\|l^{(i+1)}\|$ dimensional bias vector $\vec{b}^{(i)}$. Finally, we denote the collection of all parameters $\theta$ as the tuple $\theta \equiv (W,b)$, where $W$ and $b$ are the weights of each layer respectively. Let $a^{(i+1)}$ denote the output (activation) of layer $l^{(i)}$ obtained by computing $f \left( W^{(i)} \cdot \vec{a}^{(i)} +\vec{b^{(i)}} \right)$ where $f$ is often the Sigmoid or ReLU function. To execute a network on an $n$-dimensional input $x$, a process known as \textit{forward-propagation} is performed where $x$ is used to activate $l^{(1)}$ which activates $l^{(2)}$ and so on until the activation of $l^{(L)}$ produces the $m$-dimensional output $y$.
	
	To summarize this process, we consider $M$ a black box and denote its execution as $M(x)=y$. To train $M$ in a supervised setting, a dataset of paired samples with the form $(x_i,y_i)$ is obtained and an objective loss function $\mathcal{L}$ is defined. The loss function is used to generate a signal at the output of $M$ which is \textit{back-propagated} through $M$ to find the errors of each weight. An optimization algorithm, such as gradient descent (GD), is then used to update the weights for a number of epochs. The function $\mathcal{L}$ is often a measure of error between the input $x$ and predicted output $y'$. As a result the network learns the function $M(x_i)\approx y_i$ and can be used to make predictions on unseen data. 
	
	Some deepfake networks use a technique called one-shot or few-shot learning which enables a pre-trained network to adapt to a new dataset $X'$ similar to $X$ on which it was trained. 
	Two common approaches for this are to (1) pass information on $x'\in X'$ to the inner layers of $M$ during the feed-forward process, and (2) perform a few additional training iterations on a few samples from $X'$. \looseness=-1

		\setlength{\abovedisplayskip}{3pt}
\setlength{\belowdisplayskip}{3pt}
	\subsection{Loss Functions}
	In order to update the weights with an optimization algorithm, such as GD, the loss function must be differentiable. There are various types of loss functions which can be applied in different ways depending on the learning objective. For example, when training a $M$ as an $n$-class classifier, the output of $M$ would be the probability vector $y\in \mathbb{R}^n$. To train $M$, we perform \textit{forward-propagation} to obtain $y^{'}=M(x)$, compute the cross-entropy loss ($\mathcal{L}_{CE}$) by comparing $y'$ to the ground truth label $y$, and then perform \textit{back-propagation} and to update the weights with the training signal. The loss $\mathcal{L}_{CE}$ over the entire training set $X$ is calculated as 
	\begin{equation}
	\mathcal{L}_{CE} = 	-\sum_{i=1}^{|X|}\sum_{c=1}^{n} y_i[c] \log(y_{i}'[c])
	\end{equation}
	where $y'[c]$ is the predicted probability of $x_i$ belonging to the $c$-th class. 
	
	Other popular loss functions used in deepfake networks include the L1 and L2 norms $\mathcal{L}_{1}=|x-x_g|^1$	and $\mathcal{L}_{2}=|x-x_g|^2$. However, L1 and L2 require paired images (e.g., of $s$ and $t$ with same expression) and perform poorly when there are large offsets between the images such as different poses or facial features. This often occurs in reenactment when $x_t$ has a different pose than $x_s$ which is reflected in $x_g$, and ultimately we'd like $x_g$ to match the appearance of $x_t$.
	
	One approach to compare two unaligned images is to pass them through another network (a perceptual model) and measure the difference between the layer's activations (feature maps). This loss is called the perceptual loss ($\mathcal{L}_{perc}$) and is described in \cite{johnson2016perceptual} for image generation tasks. In the creation of deepfakes, $\mathcal{L}_{perc}$ is often computed using a face recognition network such as VGGFace. The intuition behind $\mathcal{L}_{perc}$ is that the feature maps (inner layer activations) of the perceptual model act as a normalized representation of $x$ in the context of how the model was trained. Therefore, by measuring the distance between the feature maps of two different images, we are essentially measuring their semantic difference (e.g., how similar the noses are to each other and other finer details.)	
	Similar to $\mathcal{L}_{perc}$, there is a feature matching loss ($\mathcal{L}_{FM}$) \cite{salimans2016improved} which uses the last output of a network. The idea behind $\mathcal{L}_{FM}$ is to consider the high level semantics captured by the last layer of the perceptual model (e.g., the general shape and textures of the head). 
	
	Another common loss is a type of content loss ($\mathcal{L}_{C}$) \cite{gatys2015neural} which is used to help the generator create realistic features, based on the perspective of a perceptual model. In $\mathcal{L}_{C}$, only $x_g$ is passed through the perceptual model and the difference between the network's feature maps are measured.  
	
	\subsection{Generative Neural Networks (for deepfakes)}
		
\begin{figure}
		\centering
		\includegraphics[width=\columnwidth]{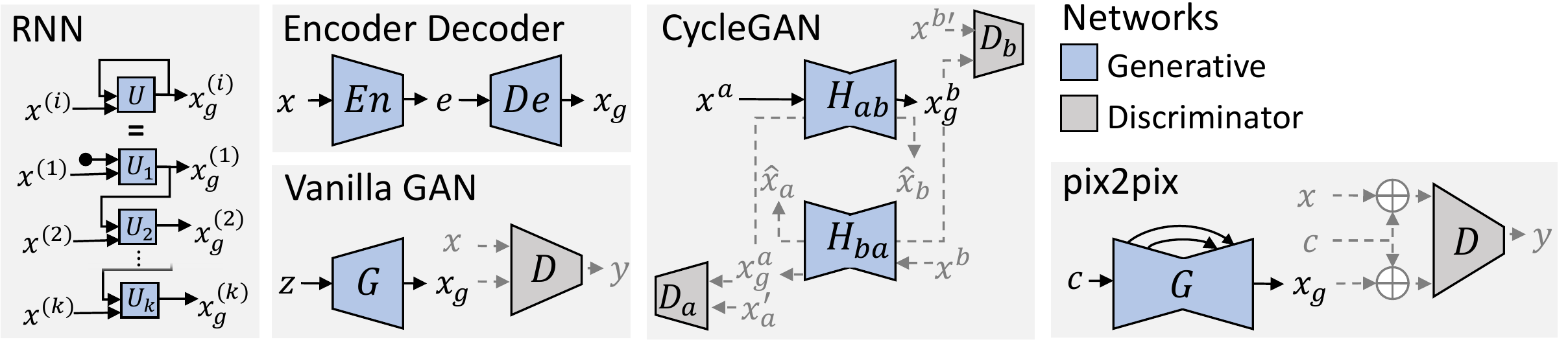}
		\caption{Five basic neural network architectures used to create deepfakes. 
		The lines indicate dataflows used during deployment (black) and training (grey).}\label{fig:basic_nns}
		\vspace{-1em}
	\end{figure}

	Deep fakes are often created using combinations or variations of six different networks, five of which are illustrated in Fig. \ref{fig:basic_nns}.  
	
\begin{description}	
	\item[Encoder-Decoder Networks (ED).] An ED consists of at least two networks, an encoder $En$ and decoder $De$. The ED has narrower layers towards its center so that when it's trained as $De(En(x))=x_g$, the network is forced to summarize the observed concepts. The summary of $x$, given its distribution $X$, is $En(x)=e$, often referred to as an encoding or embedding and $E=En(X)$ is referred to as the `latent space'. Deepfake technologies often use multiple encoders or decoders and manipulate the encodings to influence the output $x_g$. If an encoder and decoder are symmetrical, and the network is trained with the objective $De(En(x))=x$, then the network is called an autoencoder and the output is the reconstruction of $x$ denoted $\hat{x}$. Another special kind of ED is the variational autorencoder (VAE) where the encoder learns the posterior distribution of the decoder given $X$. VAEs are better at generating content than autoencoders because the concepts in the latent space are disentangled, and thus encodings respond better to interpolation and modification.
	
	\item[Convolutional Neural Network (CNN).] In contrast to a fully connected (dense) network, a CNN learns pattern hierarchies in the data and is therefore much more efficient at handling imagery. A convolutional layer in a CNN learns filters which are shifted over the input forming an abstract feature map as the output. Pooling layers are used to reduce the dimensionality as the network gets deeper and up-sampling layers are used to increase it. With convolutional, pooling, and upsampling layers, it is possible to build an ED CNNs for imagery. 
	\looseness=-1

	\item[Generative Adversarial Networks (GAN)] The GAN was first proposed in 2014 by Goodfellow et al. in \cite{goodfellow2014generative}. A GANs consist of two neural networks which work against each other: the generator $G$ and the discriminator $D$. $G$ creates fake samples $x_g$ with the aim of fooling $D$, and $D$ learns to differentiate between real samples ($x\in X$) and fake samples ($x_g=G(z)$ where $z\sim N$). Concretely, there is an adversarial loss used to train $D$ and $G$ respectively:
	\begin{gather}
		\mathcal{L}_{adv}(D)=\max \log D(x) + \log(1 - D(G(z))) \\
		\mathcal{L}_{adv}(G)=\min \log(1 - D(G(z)))
	\end{gather}
	This zero-sum game leads to $G$ learning how to generate samples that are indistinguishable from the original distribution. After training, $D$ is discarded and $G$ is used to generate content. When applied to imagery, this approach produces photo realistic images. 
	
	\vspace{.5em}
	Numerous of variations and improvements of GANs have been proposed over the years. In the creation of deepfakes, there are two popular image translation frameworks which use the fundamental principles of GANs: 
	
	\begin{description}
		\item[Image-to-Image Translation (pix2pix).] The pix2pix framework enables paired translations from one image domain to another \cite{isola2017image}. In pix2pix, $G$ tries to generate the image $x_g$ given a visual context $x_c$ as an input, and $D$ discriminates between $(x,x_c)$ and $(x_g,x_c)$. Moreover, $G$ is a an ED CNN with skip connections from $En$ to $De$ (called a U-Net) which enables $G$ to produce high fidelity imagery by bypassing the compression layers when needed. Later, pix2pixHD was proposed \cite{wang2018high} for generating high resolution imagery with better fidelity. 

		\item[CycleGAN.] An improvement of pix2pix which enables image translation through unpaired training \cite{zhu2017unpaired}. The network forms a cycle consisting of two GANs used to convert images from one domain to another, and then back again to ensure consistency with a cycle consistency loss ($\mathcal{L}_{cyc}$). 
	\end{description}

	\item[Recurrent Neural Networks (RNN)] An RNN is type of neural network that can handle sequential and variable length data. The network remembers is internal state after processing $x^{(i-1)}$ and can use it to process $x^{(i)}$ and so on. In deepfake creation, RNNs are often used to handle audio and sometimes video. More advanced versions of RNNs include long short-term memory (LSTM) and gate reccurent units (GRU).
	
\end{description}
\vspace{-1em}
\subsection{Feature Representations}
	Most deep fake architectures use some form of intermediate representation to capture and sometimes manipulate $s$ and $t$'s facial structure, pose, and expression. One way is to use the facial action coding system (FACS) and measure each of the face's taxonomized action units (AU) \cite{ekman2002facial}. Another way is to use monocular reconstruction to obtain a 3D morphable model (3DMM) of the head from a 2D image, where the pose and expression are parameterized by a set of vectors and matrices. Then use the parameters or a 3D rendering of the head itself. Some use a UV map of the head or body to give the network a better understanding of the shape's orientation. 
	
	Another approach is to use image segmentation to help the network separate the different concepts (face, hair, etc). The most common representation is landmarks (a.k.a. key-points) which are a set of defined positions on the face or body which can be efficiently tracked using open source CV libraries. The landmarks are often presented to the networks as a 2D image with Gaussian points at each landmark. Some works separate the landmarks by channel to make it easier for the network to identity and associate them. Similarly, facial boundaries and body skeletons can also used.
	\looseness=-1
	
	For audio (speech), the most common approach is to split the audio into segments, and for each segment, measure the Mel-Cepstral Coefficients (MCC) which captures the dominant voice frequencies.\looseness=-1

	\subsection{Deepfake Creation Basics}
	\begin{figure}[t]
		\centering
		\includegraphics[width=\columnwidth]{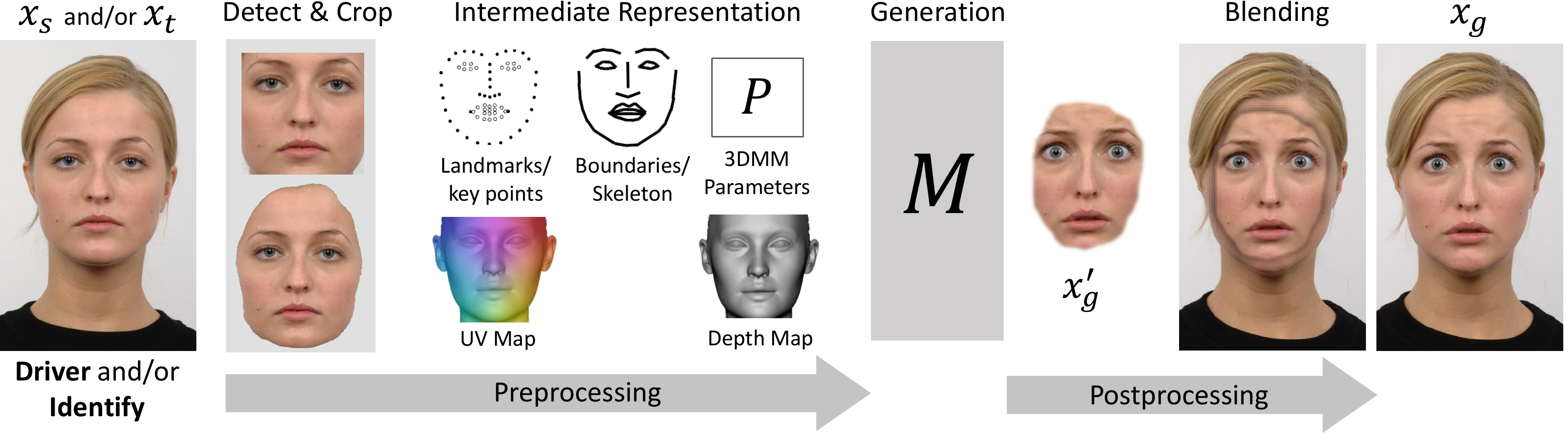}
		\caption{The processing pipeline for making reenactment and face swap deepfakes. Usually only a subset of these steps are performed.}\label{fig:pipeline}
		\vspace{-.5em}
	\end{figure}
	To generate $x_g$, reenactment and face swap networks follow some variation of this process (illustrated in Fig. \ref{fig:pipeline}): Pass $x$ through a pipeline that (1) detects and crops the face, (2) extracts intermediate representations, (3) generates a new face based on some driving signal (e.g., another face), and then (4) blends the generated face back into the target frame. 

\vspace{.5em}
\noindent In general there are six approaches to driving an image:
\begin{enumerate}
\item Let a network work directly on the image and perform the mapping itself. 
\item Train an ED network to disentangle the identity from the expression, and then modify/swap the encodings of the target the before passing it through the decoder.
\item Add an additional encoding (e.g., AU or embedding) before passing it to the decoder.
\item Convert the intermediate face/body representation to the desired identity/expression before generation (e.g., transform the boundaries with a secondary network or render a 3D model of the target with the desired expression).
\item Use the optical flow field from subsequent frames in a source video to drive the generator.
\item Create composite of the original content (hair, scene, etc) with a combination of the 3D rendering, warped image, or generated content, and pass the composite through another network (such as pix2pix) to refine the realism.
\end{enumerate}

	\subsection{Generalization}
	A deepfake network may be trained or designed to work with only a specific set of target and source identities. An identity agnostic model is sometimes hard to achieve due to correlations learned by the model between $s$ and $t$ during training. 
	
	Let $E$ be some model or process for representing or extracting features from $x$, and let $M$ be a \textit{trained} model for performing replacement or reenactment. We identify three primary categories in regard to generalization:
	\begin{description}
		\item[one-to-one:] A model that uses a specific identity to drive a specific identity: $x_g = M_t(E_s(x_s))$	\looseness=-1
		\item[many-to-one:]  A model that uses any identity to drive a specific identity: $x_g = M_t(E(x_s))$
		\item[many-to-many:]  A model that uses any identity to drive any identity: $x_g = M(E_1(x_s),E_2(x_t))$
	\end{description}

	\subsection{Challenges}\label{subsec:challenges}
	The following are some challenges in creating realistic deepfakes:
	\begin{description}
		\item[Generalization.] Generative networks are data driven, and therefore reflect the training data in their outputs. This means that high quality images of a specific identity requires a large number of samples of that identity. Moreover, access to a large dataset of the driver is typically much easier to obtain than the victim. As a result, over the last few years, researchers have worked hard to minimize the amount of training data required, and to enable the execution of a trained model on new target and source identities (unseen during training).
		\item[Paired Training.] One way to train a neural network is to present the desired output to the model for each given input. This process of \textit{data pairing} is a laborious an sometimes impractical when training on multiple identities and actions. To avoid this issue, many deepfake networks either (1) train in a self-supervised manner by using frames selected from the same video of $t$, (2) use unpaired networks such as Cycle-GAN, or (3) utilize the encodings of an ED network.  
		\item[Identity Leakage.] Sometimes the identity of the driver (e.g., $s$ in reenactment) is partially transferred to $x_g$. This occurs when training on a single input identity, or when the network is trained on many identities but data pairing is done with the same identity. Some solutions proposed by researchers include attention mechanisms, few-shot learning,  disentanglement, boundary conversions, and AdaIN or skip connections to carry the relevant information to the generator.
		\item[Occlusions.] Occlusions are where part of $x_s$ or $x_t$ is obstructed with a hand, hair, glasses, or any other item. Another type of obstruction is the eyes and mouth region that may be hidden or dynamically changing. As a result, artifacts appear such as cropped imagery or inconsistent facial features. To mitigate this, works such as \cite{nirkin2019fsgan,pumarola2019ganimation,NIPS2019_8935} perform segmentation and in-painting on the obstructed areas. 
		\item[Temporal Coherence.] Deepfake videos often produce more obvious artifacts such as flickering and jitter \cite{vougioukas2019end}. This is because most deepfake networks process each frame individually with no context of the preceding frames. To mitigate this, some researchers either provide this context to $G$ and $D$, implement temporal coherence losses, use RNNs, or perform a combination thereof.
	\end{description}

	\section{Reenactment}\label{sec:reen}

\bgroup
\setlength\tabcolsep{.1em}
\scriptsize
\begin{table*}[hbt!]
	\centering
	\caption{Summary of Deep Learning Reenactment Models (Body and Face)}
	\vspace{-.5em}
	\label{tab:reenactment}
	\resizebox{\textwidth}{!}{%
\begin{tabular}{@{}ccccccccccccccccccccccccccccccccccc@{}}
			\toprule
			&  &  &  & \multicolumn{5}{c}{Reenactment} & \multicolumn{3}{c}{Retraining for new...} & \multicolumn{4}{c}{Model} & \multicolumn{6}{c}{Representation} & \multicolumn{8}{c}{Model Training} & \multicolumn{2}{c}{Model Execution} & \multicolumn{3}{c}{Model Output} \\
\end{tabular}%
	}
\vspace{-2em}
\end{table*}
\egroup

	In this section we present a chronological review of deep learning based reenactment, organized according to their class of identity generalization. Table \ref{tab:reenactment} provides a summary and systematization of all the works mentioned in this section. Later, in Section \ref{sec:discussion}, we contrast the various methods and identify the most significant approaches.
	
	\subsection{Expression Reenactment}\label{subsec:expression}
	Expression reenactment turns an identity into a puppet, giving attackers the most flexibility to achieve their desired impact. Before we review the subject, we note that expression reenactment has been around long before deepfakes were popularized. In 2003, researchers morphed models of 3D scanned heads \cite{blanz2003reanimating}. In 2005, it was shown how this can be done without a 3D model \cite{chang2005transferable}, and through warping with matching similar textures \cite{garrido2014automatic}. Later, between 2015 and 2018, Thies et al. demonstrated how 3D parametric models can be used to achieve high quality and real-time results with depth sensing and ordinary cameras (\cite{thies2015real} and \cite{thies2016face2face,thies2018headon}). 
	
\begin{figure}	
	
	\begin{subfigure}[t]{.49\textwidth}	
	\vspace{1em}
	\includegraphics[width=.9\textwidth]{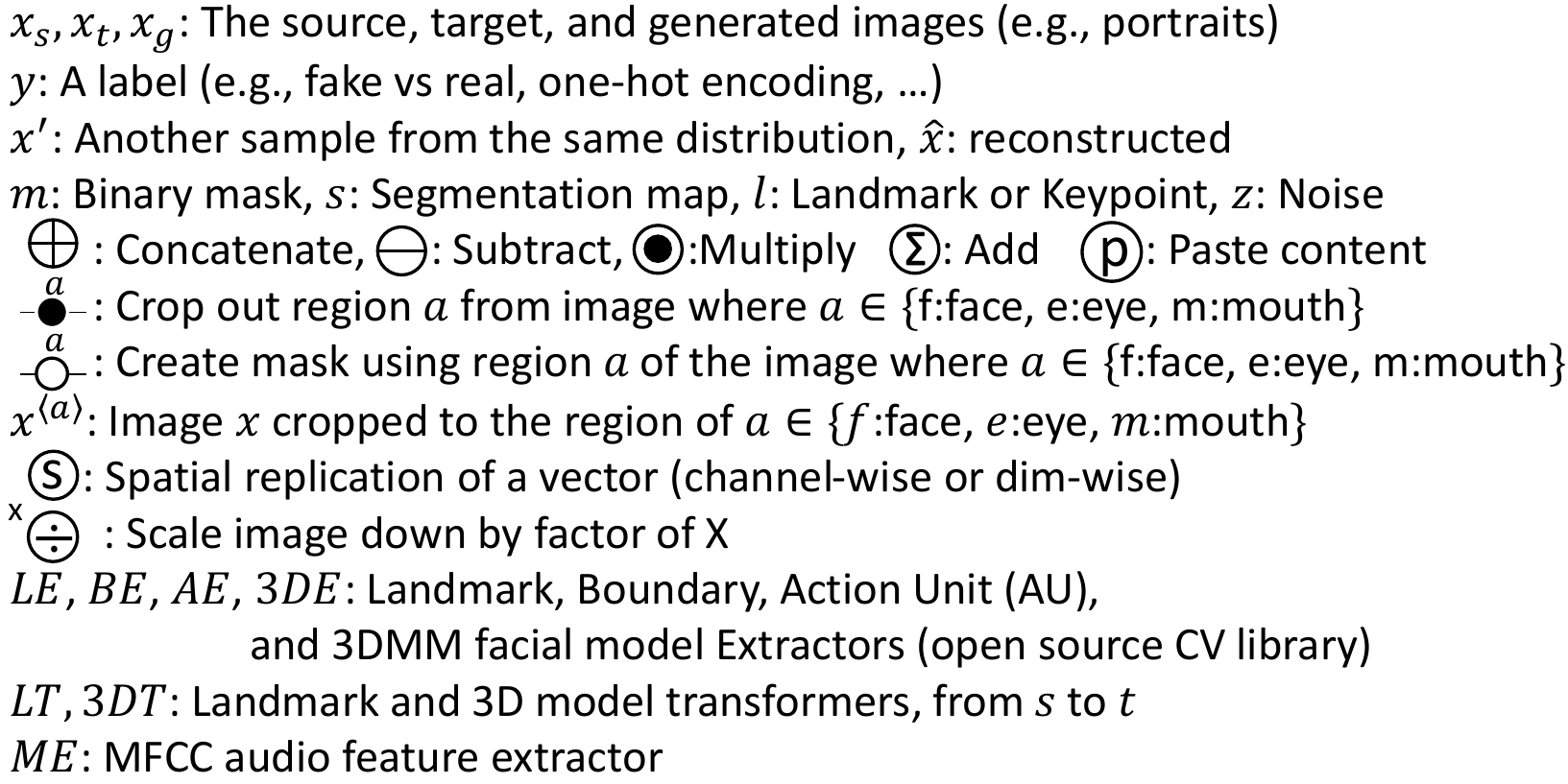}\\
	\tiny \textbf{Losses}: $\mathcal{L}_{1}:$ L1, $\mathcal{L}_{2}:$ L2, $\mathcal{L}_{CE}:$ Cross Entropy, $\mathcal{L}_{adv}:$ Adversarial, $\mathcal{L}_{FM}:$ Feature Matching, $\mathcal{L}_{perc}:$ Perceptual, $\mathcal{L}_{cyc}:$ Cycle Consistency, $\mathcal{L}_{att}:$ Attention, $\mathcal{L}_{trip}:$ Triplet, $\mathcal{L}_{tv}:$ Total Variance, $\mathcal{L}_{KL}:$ KL Divergence
\end{subfigure}		
		\begin{subfigure}[t]{.49\textwidth}	
		\centering
		\caption{\textbf{\cite{wu2018reenactgan} Reenact GAN:}}
		\includegraphics[width=\textwidth]{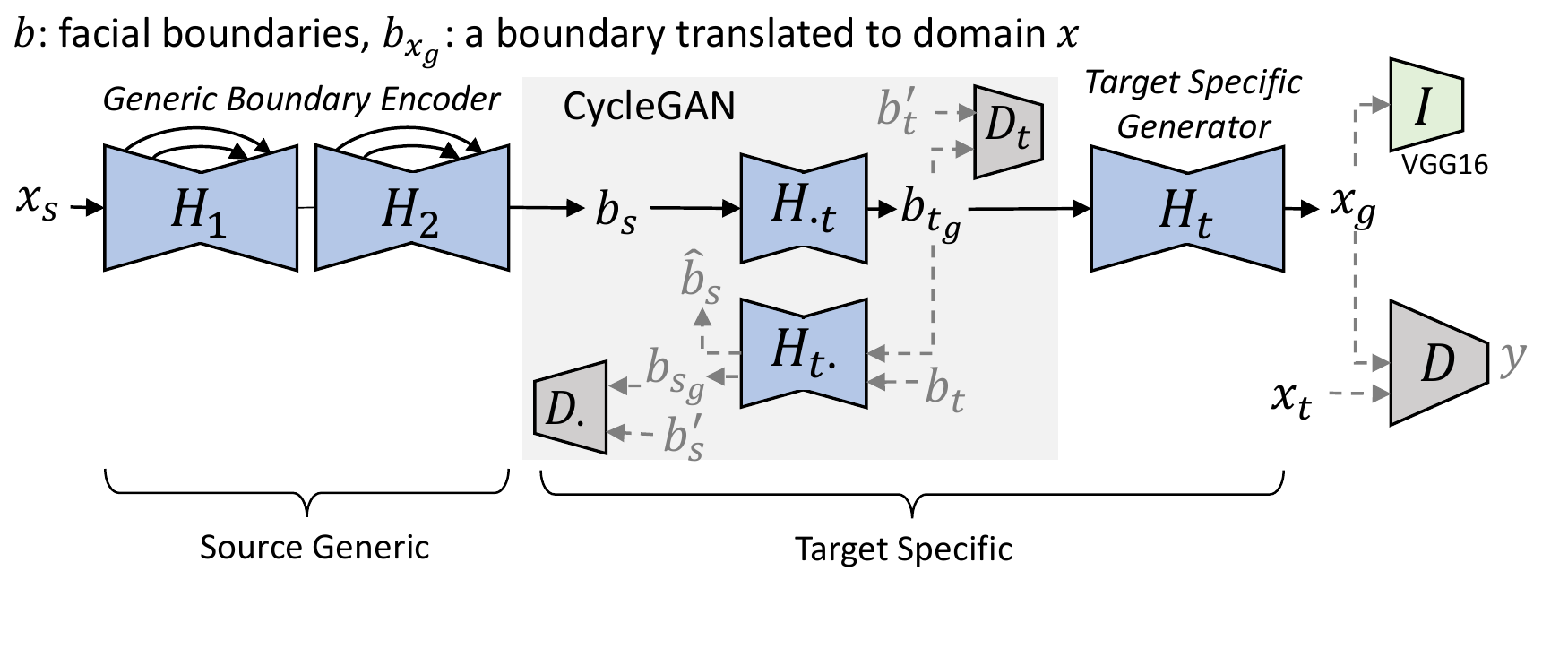}\vspace{-.5em}
		\includegraphics[width=\textwidth]{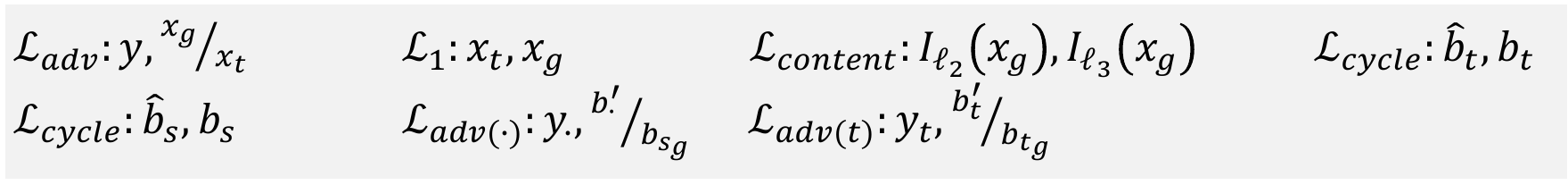}
	\end{subfigure}
	\begin{subfigure}[t]{.49\textwidth}	
	\centering
	\caption{\textbf{\cite{tulyakov2018mocogan} MocoGAN:} }
	\includegraphics[width=\textwidth]{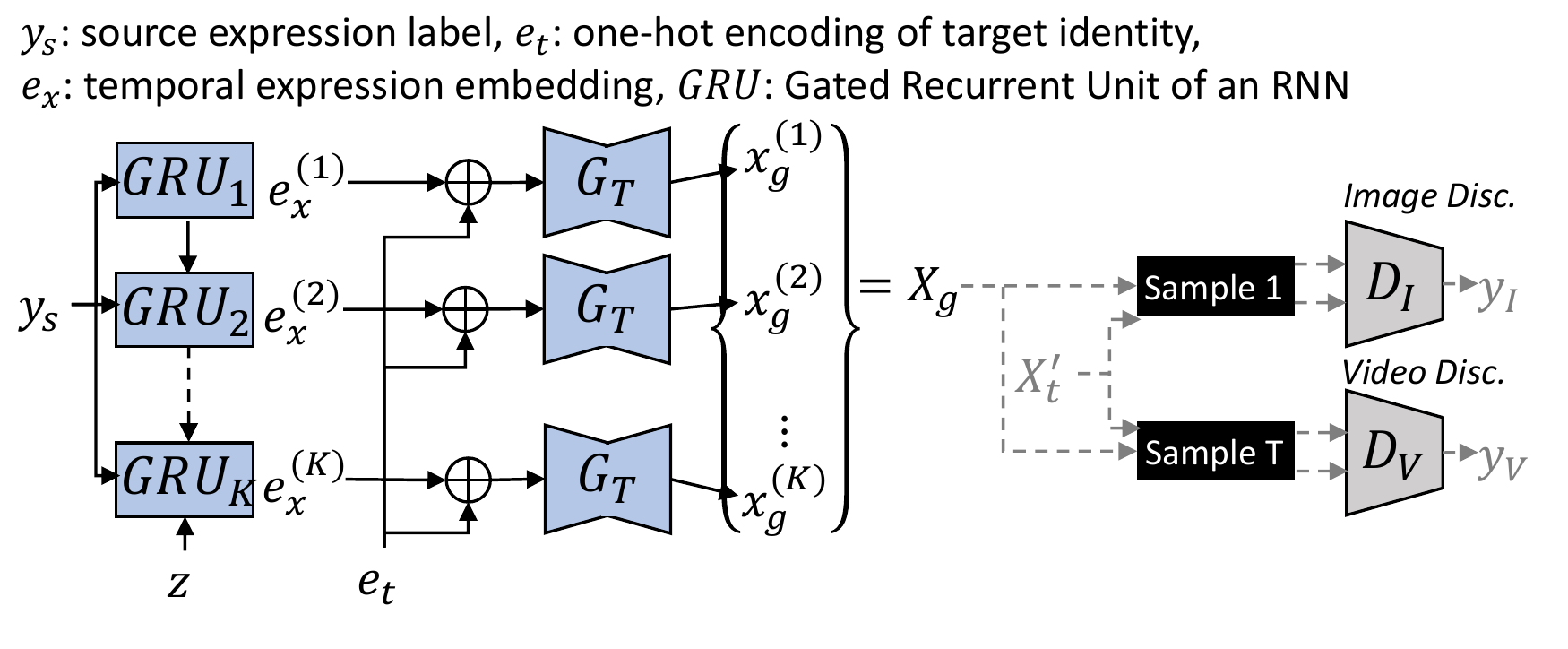}\vspace{-.5em}
		\includegraphics[width=\textwidth]{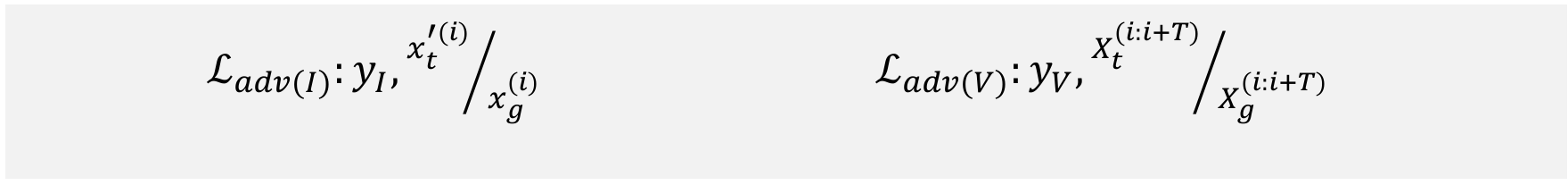}
\end{subfigure}
	\begin{subfigure}[t]{.49\textwidth}	
	\centering
	\caption{\textbf{\cite{wang2018vid2vid} Vid2Vid:}}
	\includegraphics[width=\textwidth]{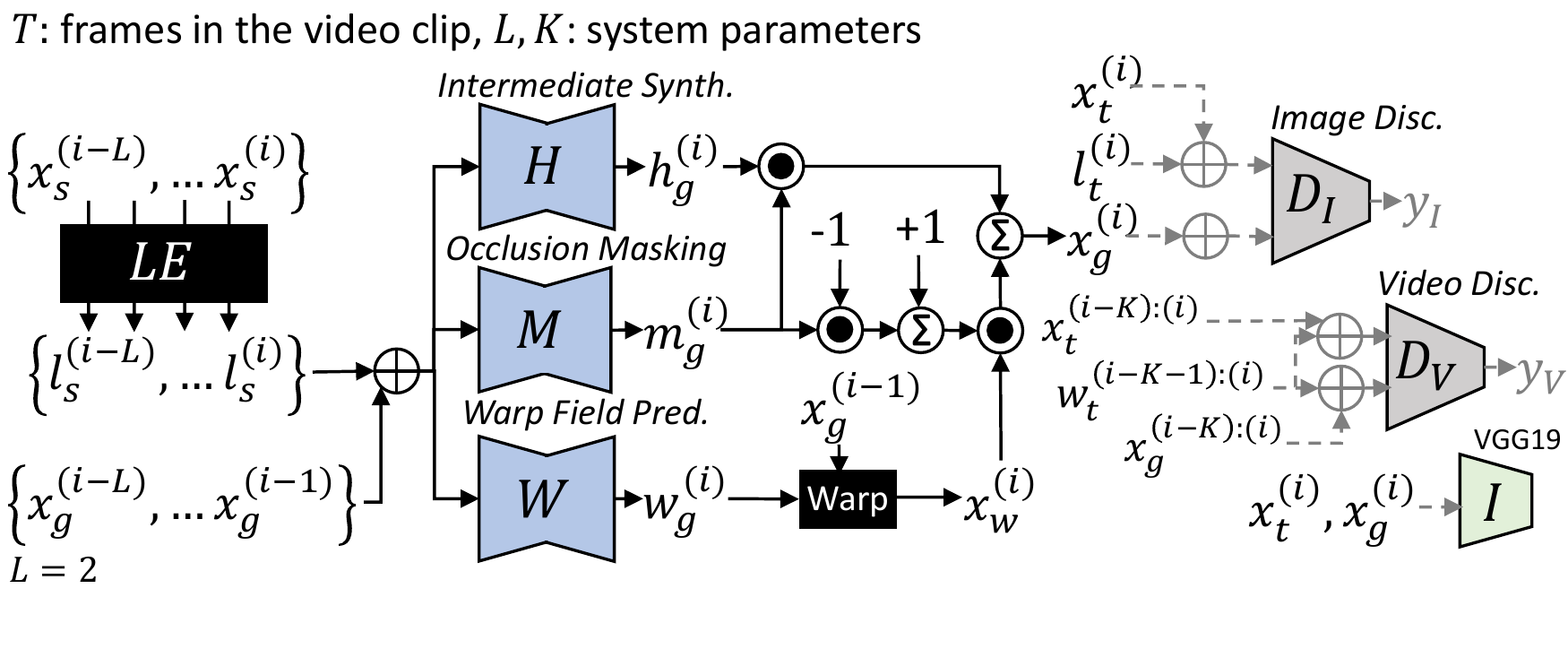}\vspace{-.5em}
		\includegraphics[width=\textwidth]{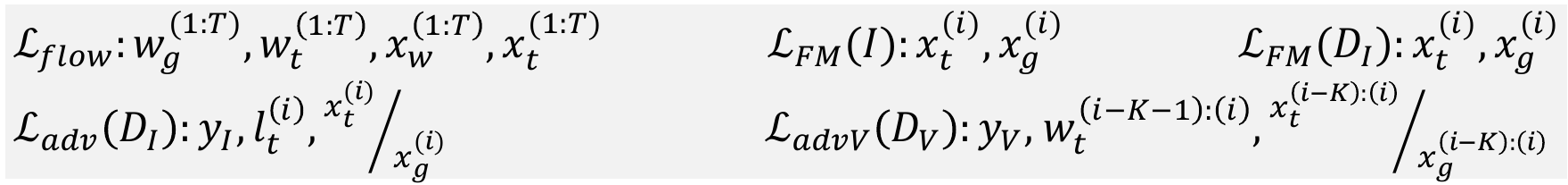}
\end{subfigure}
	\begin{subfigure}[t]{.49\textwidth}	
	\centering
	\caption{\textbf{\cite{kim2018deep} Deep Video Portrait:} }
	\includegraphics[width=\textwidth]{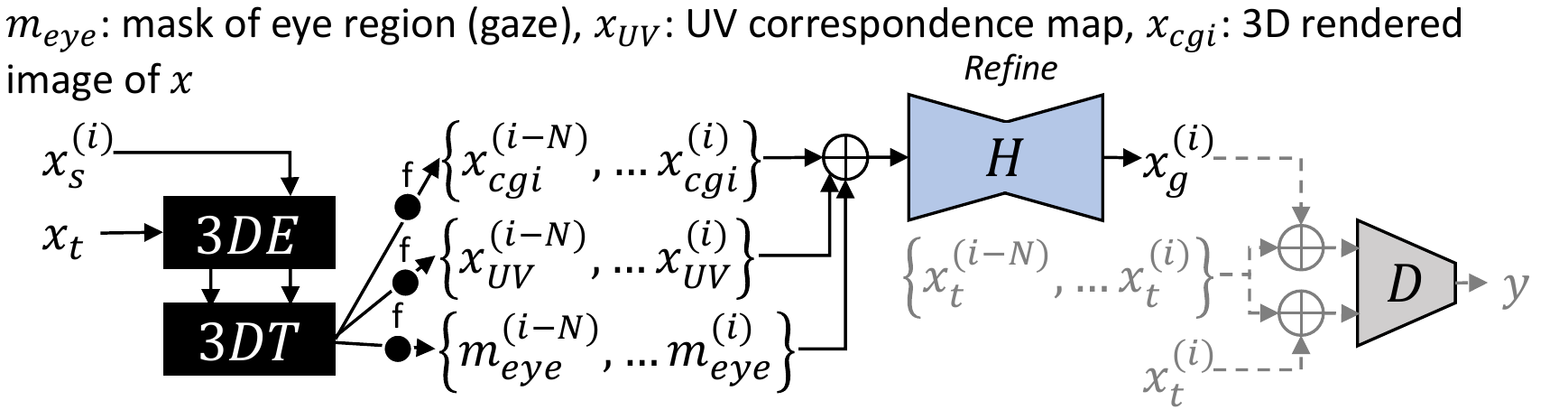}\vspace{.3em}
		\includegraphics[width=\textwidth]{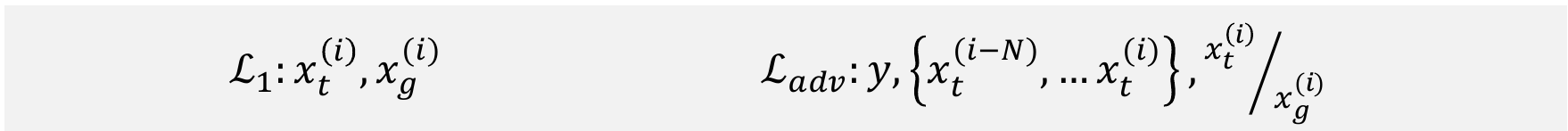}
\end{subfigure}
	\begin{subfigure}[t]{.49\textwidth}	
	\centering
	\caption{\textbf{\cite{pham2018generative} GATH:} }
	\includegraphics[width=\textwidth]{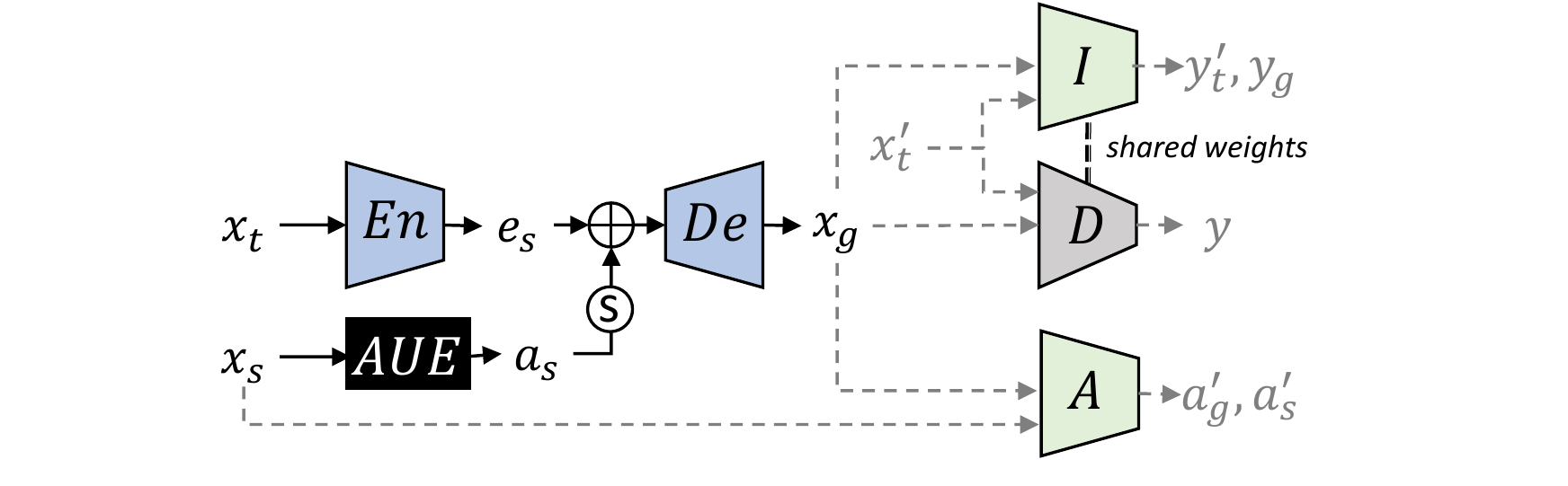}\vspace{-.3em}
		\includegraphics[width=\textwidth]{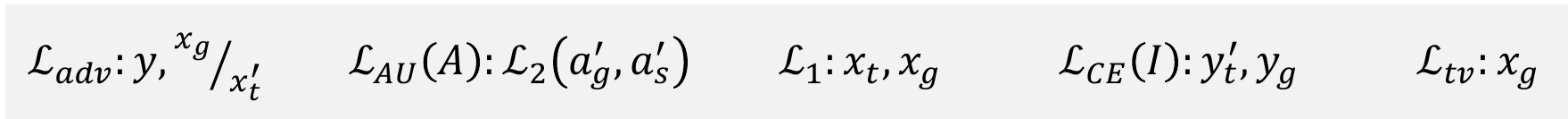}
\end{subfigure}
	\begin{subfigure}[t]{.49\textwidth}	
	\centering
	\caption{\textbf{\cite{pumarola2019ganimation} GANimation:} }
	\includegraphics[width=\textwidth]{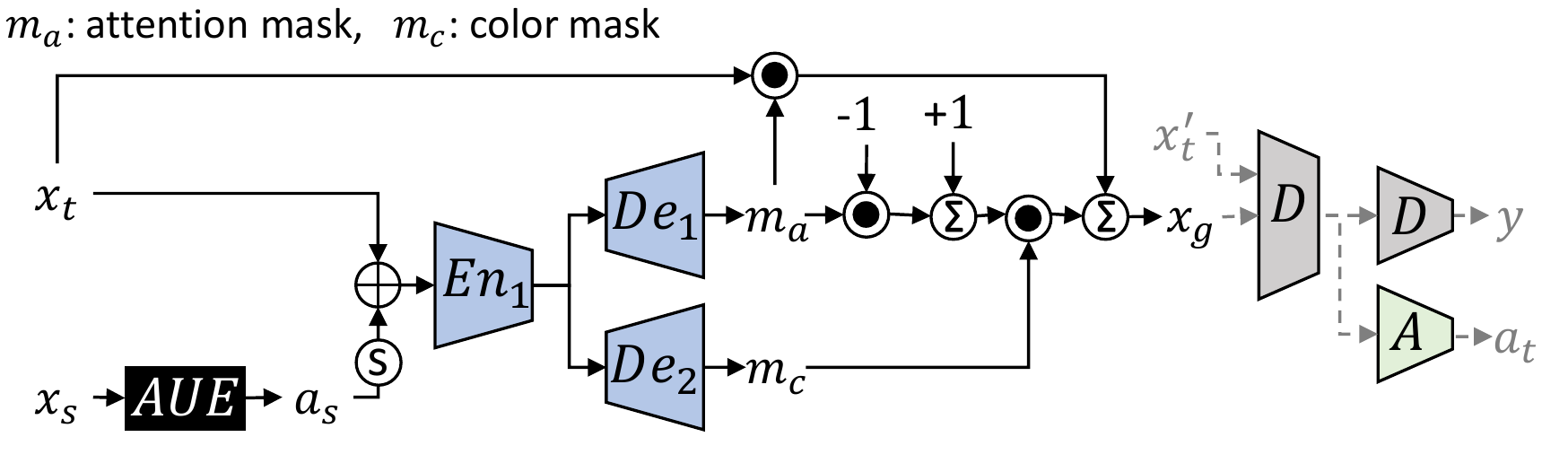}\vspace{-.5em}
		\includegraphics[width=\textwidth]{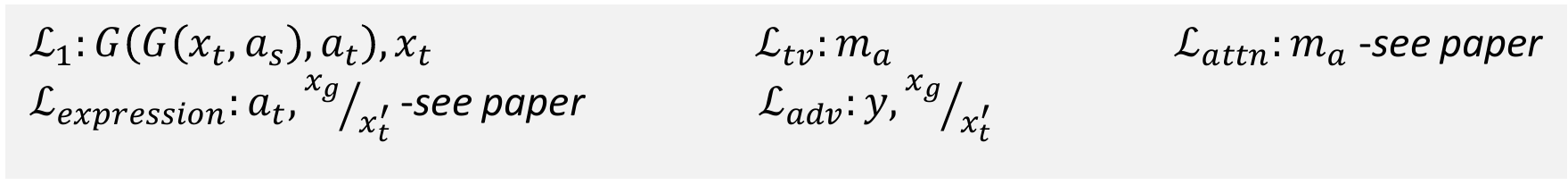}
\end{subfigure}
\begin{subfigure}[t]{.49\textwidth}	
	\centering
	\caption{\textbf{\cite{sanchez2018triple} GANotation:}}
	\includegraphics[width=\textwidth]{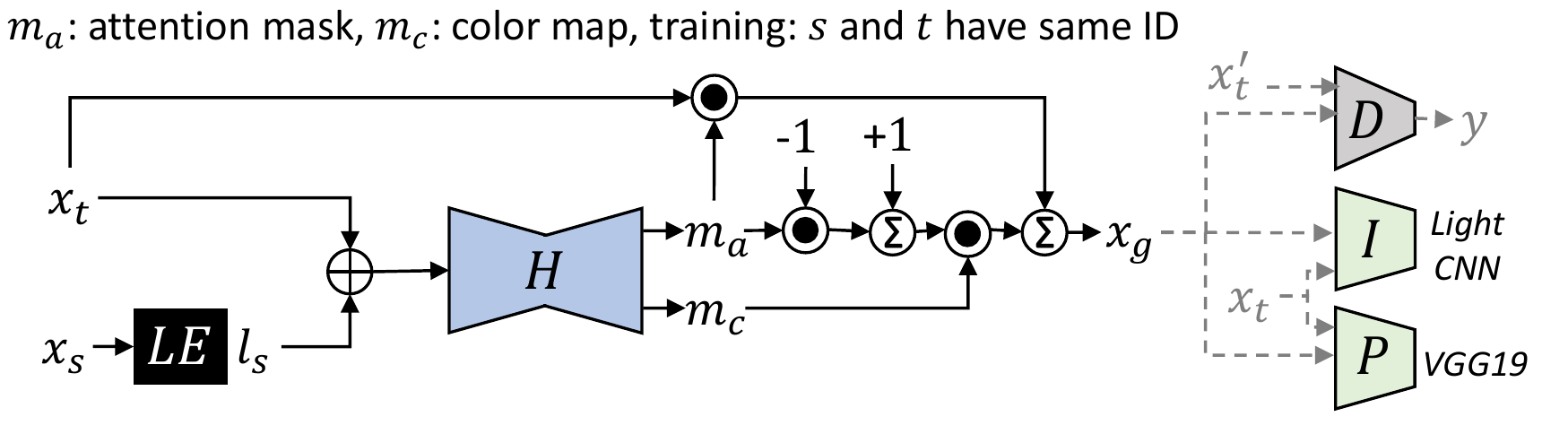}
		\includegraphics[width=\textwidth]{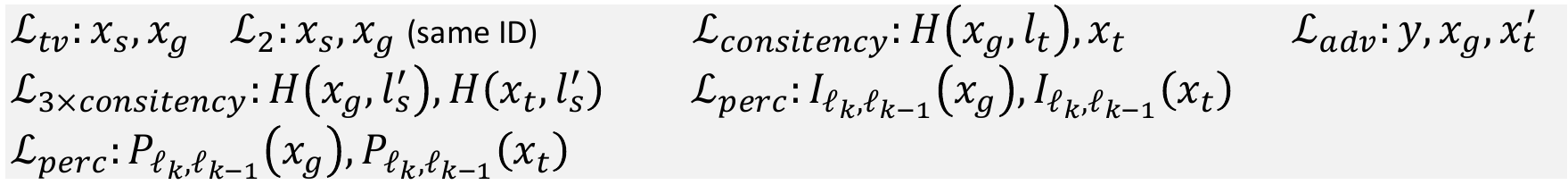}
	\vspace{-3em}
\end{subfigure}
	\begin{subfigure}[t]{.49\textwidth}	
	\centering
	\caption{\textbf{\cite{shen2018faceid} FaceID-GAN:} }
	\includegraphics[width=\textwidth]{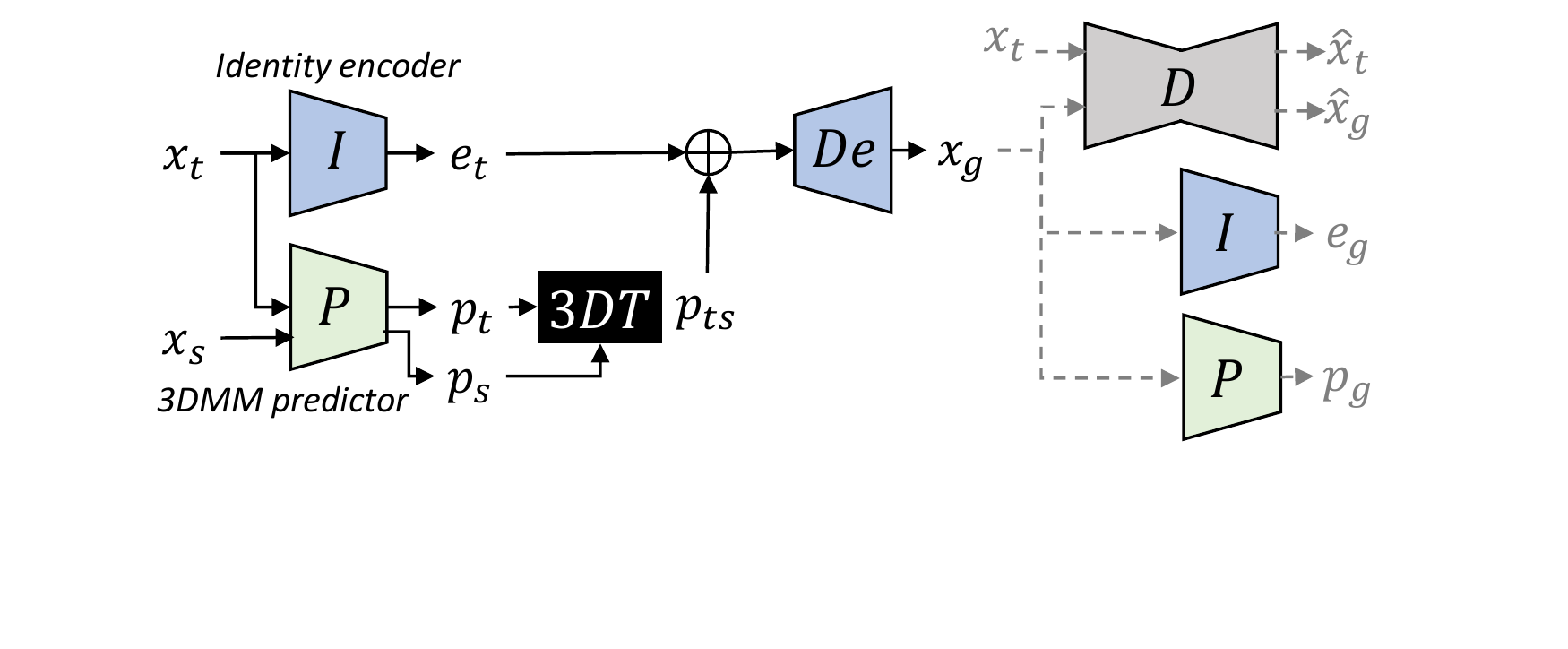}\vspace{-.1em}
		\includegraphics[width=\textwidth]{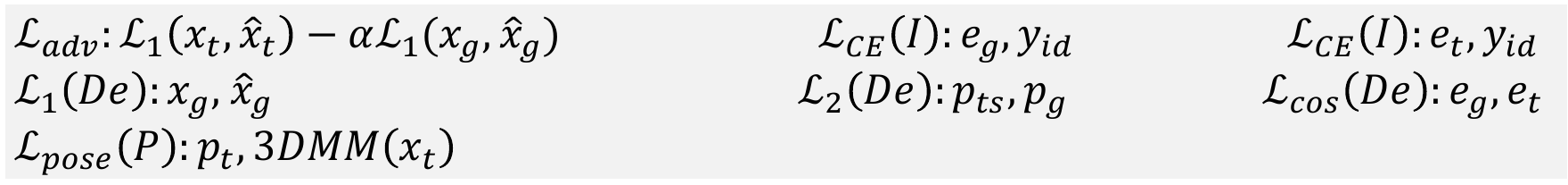}
\end{subfigure}
	\begin{subfigure}[t]{.49\textwidth}	
	\centering
	\caption{\textbf{\cite{shen2018facefeat} FaceFeat-GAN:} }
	\includegraphics[width=\textwidth]{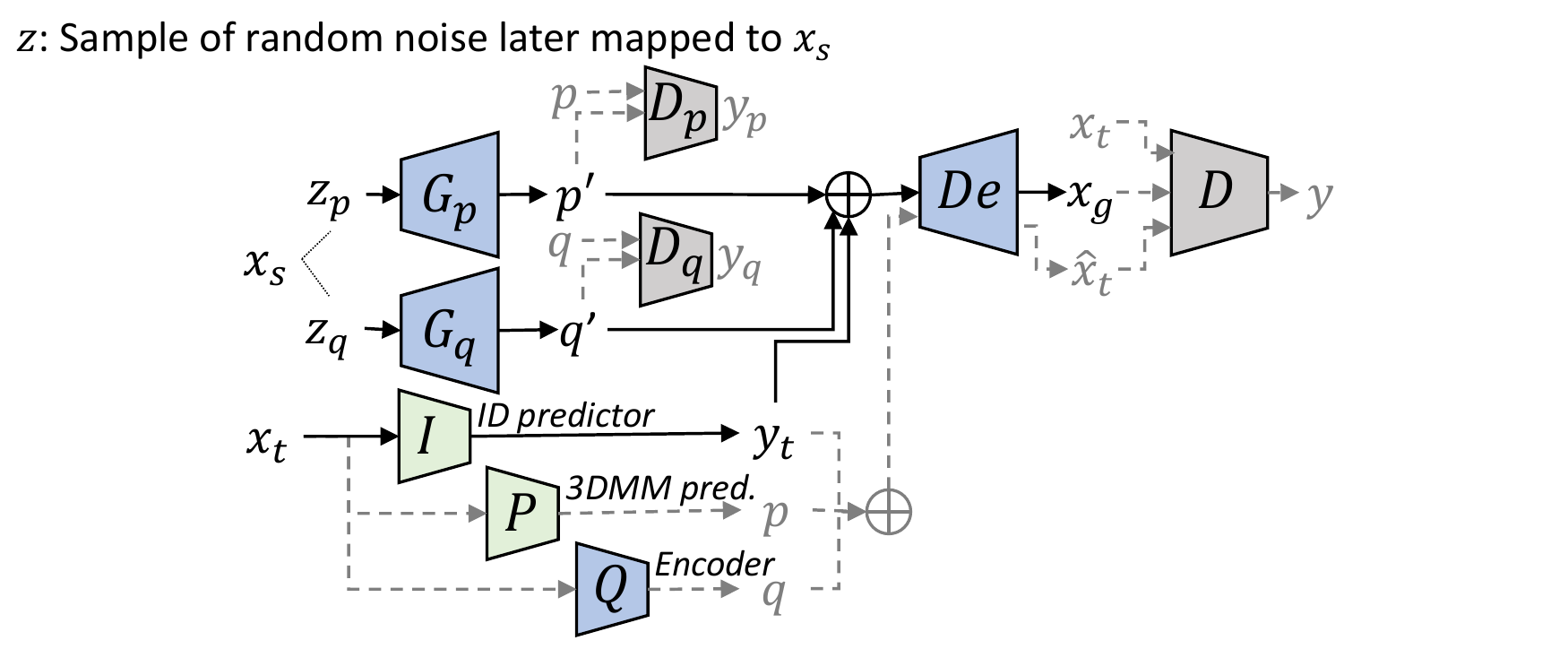}\vspace{-.1em}
		\includegraphics[width=\textwidth]{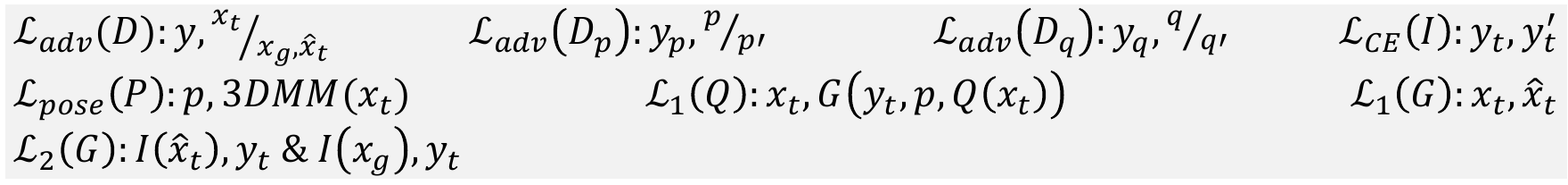}
\end{subfigure}
\vspace{-.5em}
	\caption{Architectural schematics of \textbf{reenactment networks}. Black lines indicate prediction flows used during deployment, dashed gray lines indicate dataflows performed during training. Zoom in for more detail.
}\label{fig:schem_reen1}
\end{figure}

\begin{figure}	
	\begin{subfigure}[t]{.49\textwidth}	
	\centering
	\caption{\textbf{\cite{nagano2018pagan} paGAN:} }
	\includegraphics[width=\textwidth]{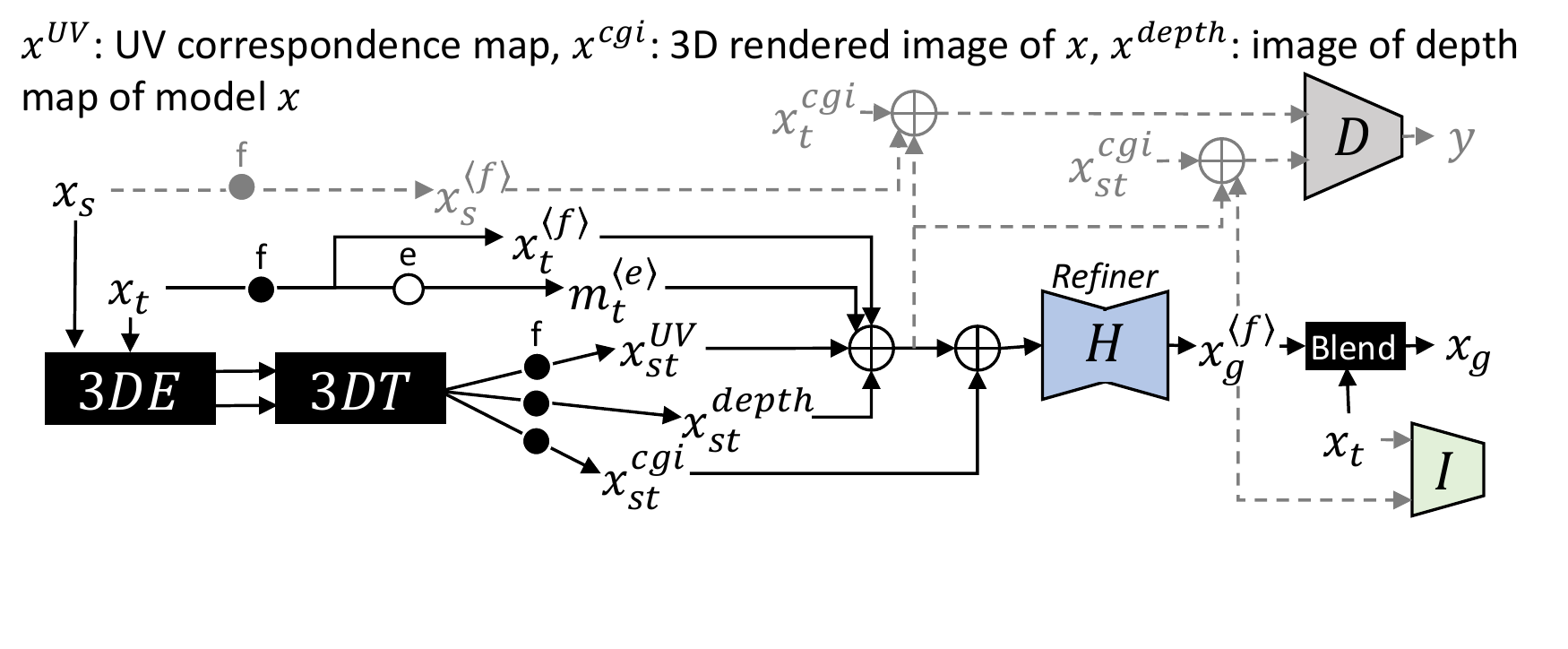}\vspace{-.6em}
		\includegraphics[width=\textwidth]{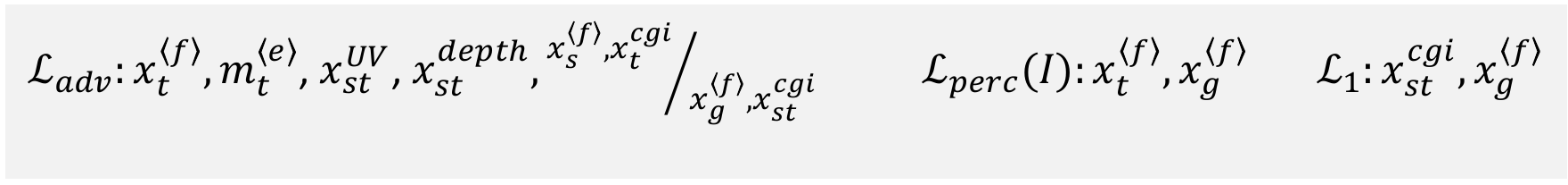}
\end{subfigure}
	\begin{subfigure}[t]{.49\textwidth}	
	\centering
	\caption{\textbf{\cite{wiles2018x2face} X2Face:} }
	\includegraphics[width=\textwidth]{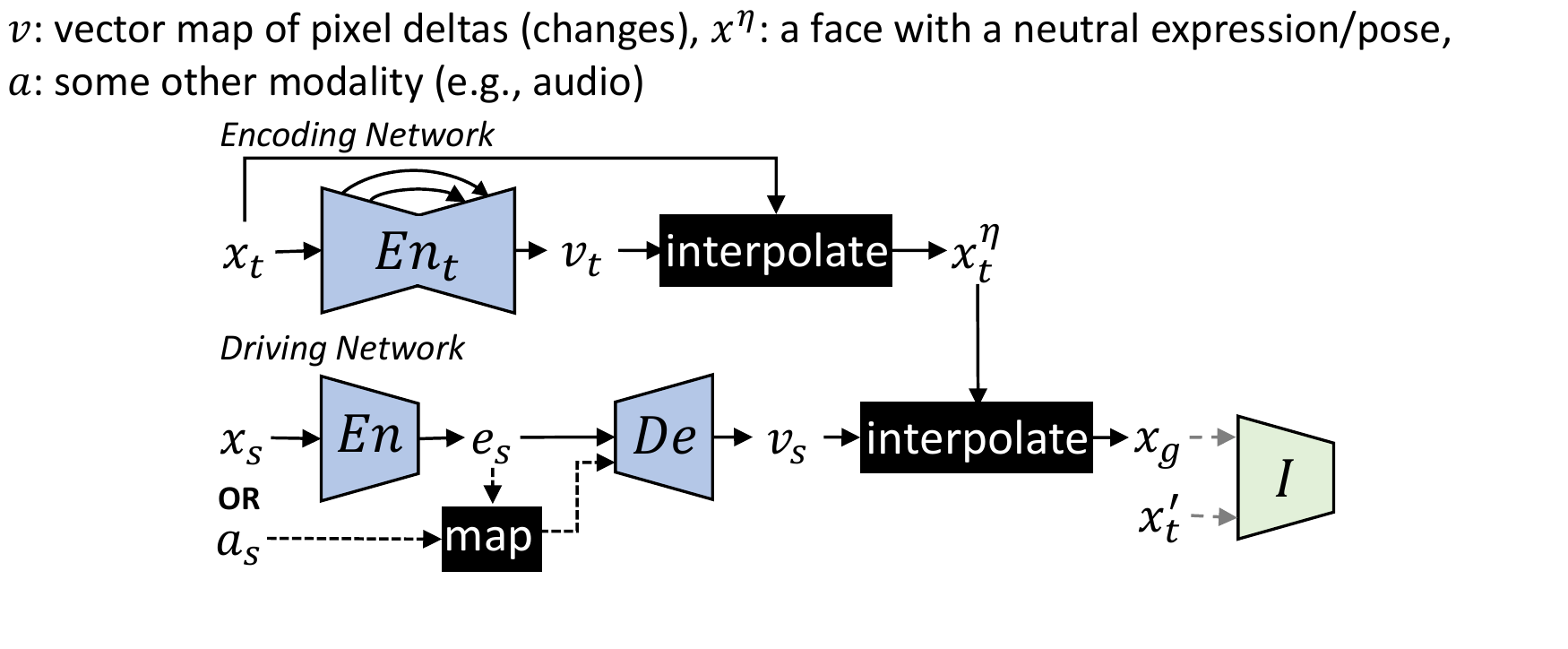}\vspace{-.6em}
		\includegraphics[width=\textwidth]{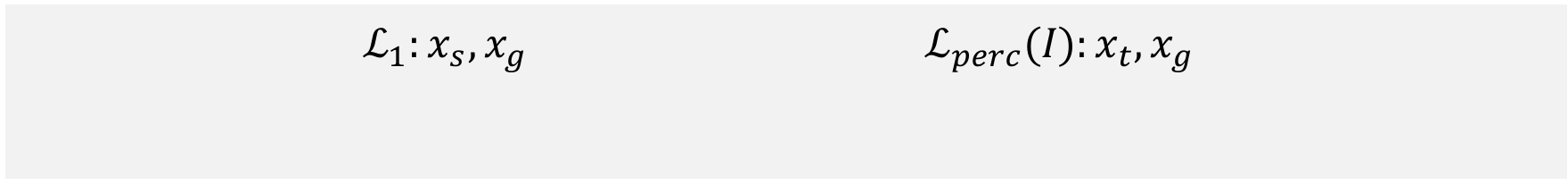}
\end{subfigure}
	\begin{subfigure}[t]{.49\textwidth}	
	\centering
	\caption{\textbf{\cite{zhang2019faceswapnet} FaceSwapNet:} }
	\includegraphics[width=\textwidth]{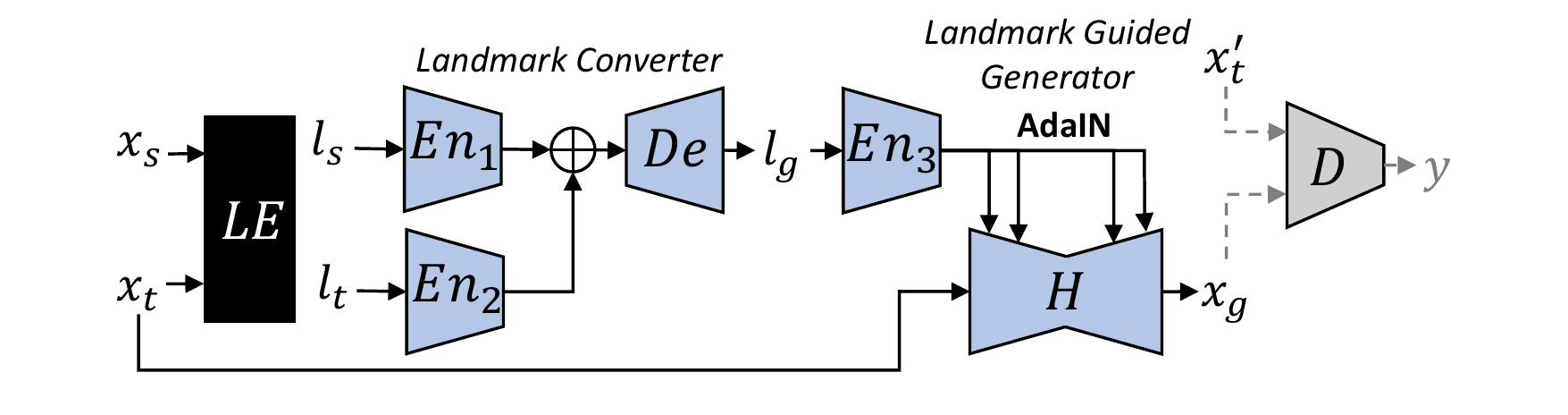}\vspace{-.3em}
		\includegraphics[width=\textwidth]{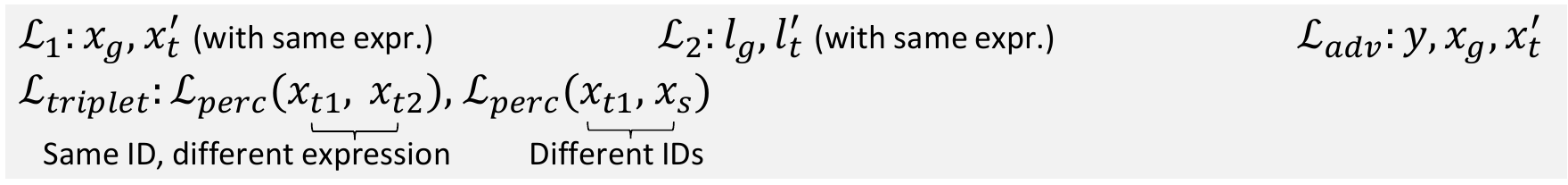}
\end{subfigure}
	\begin{subfigure}[t]{.49\textwidth}	
	\centering
	\caption{\textbf{\cite{nirkin2019fsgan} FSGAN:} }
	\includegraphics[width=\textwidth]{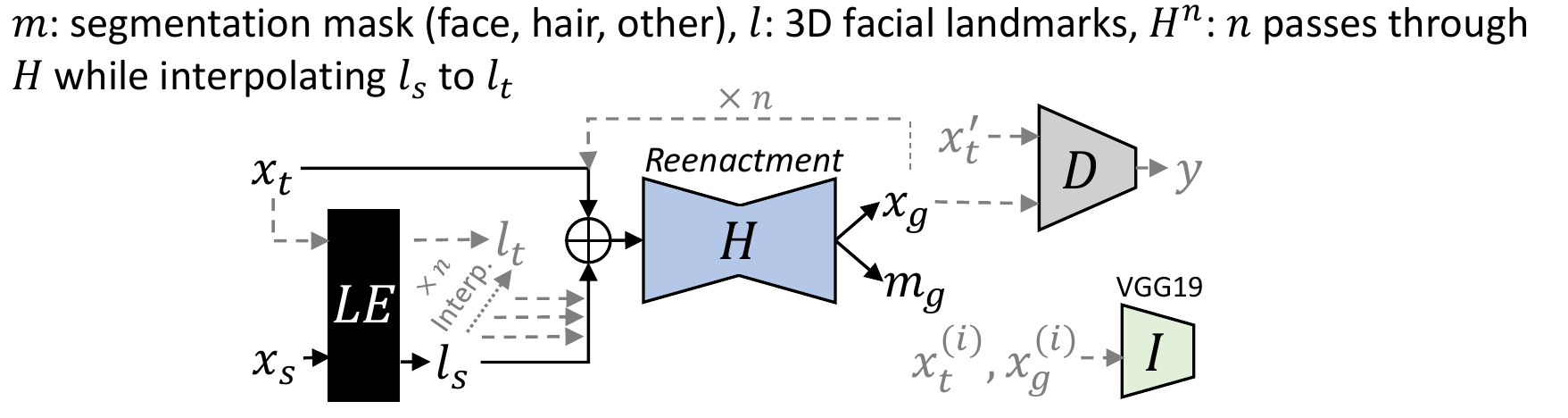}\vspace{-.3em}
		\includegraphics[width=\textwidth]{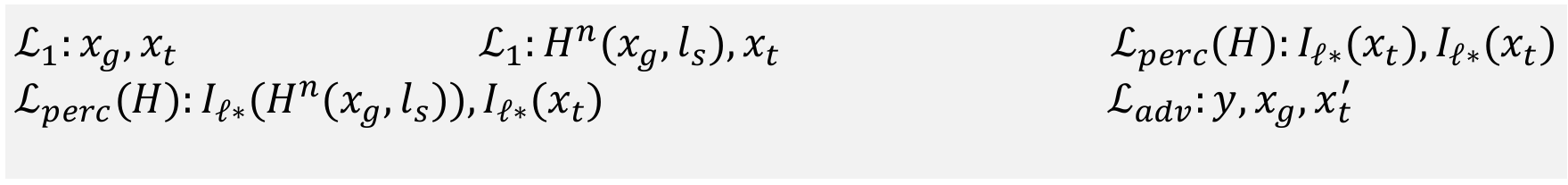}
\end{subfigure}
	\begin{subfigure}[t]{.49\textwidth}	
	\centering
	\caption{\textbf{\cite{fu2019high} Fu et al. 2019:} }
	\includegraphics[width=\textwidth]{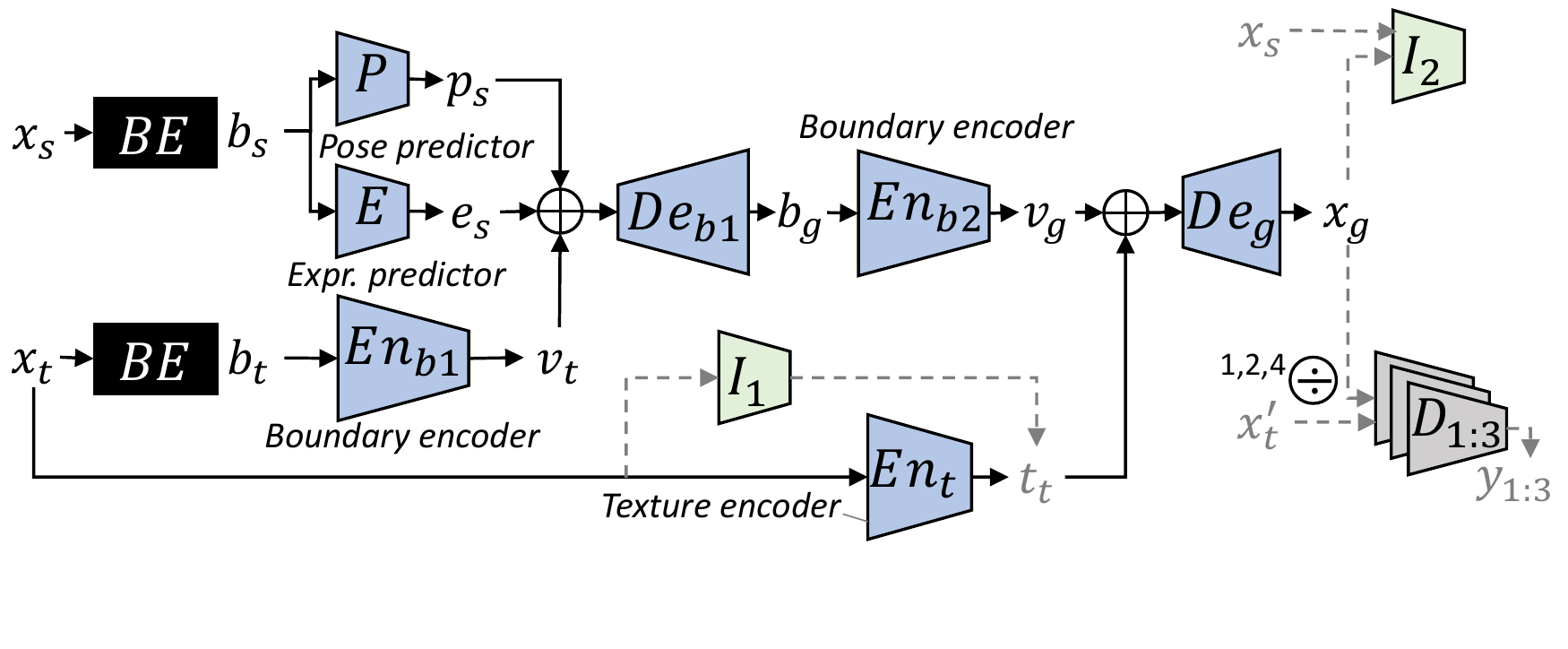}\vspace{-.8em}
		\includegraphics[width=\textwidth]{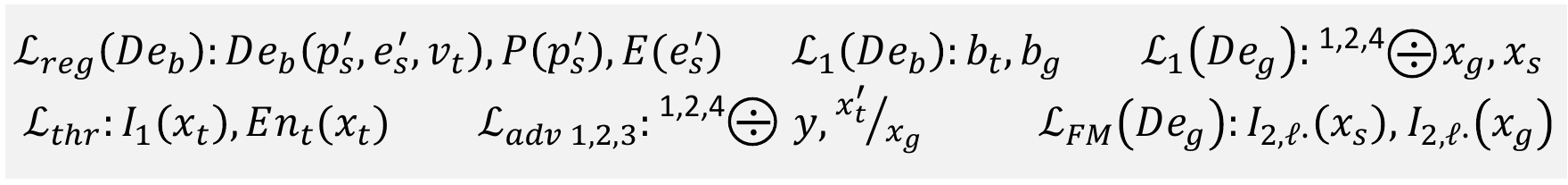}
\end{subfigure}
	\begin{subfigure}[t]{.49\textwidth}	
	\centering
	\caption{\textbf{\cite{tripathy2019icface} ICFace:} }
	\includegraphics[width=\textwidth]{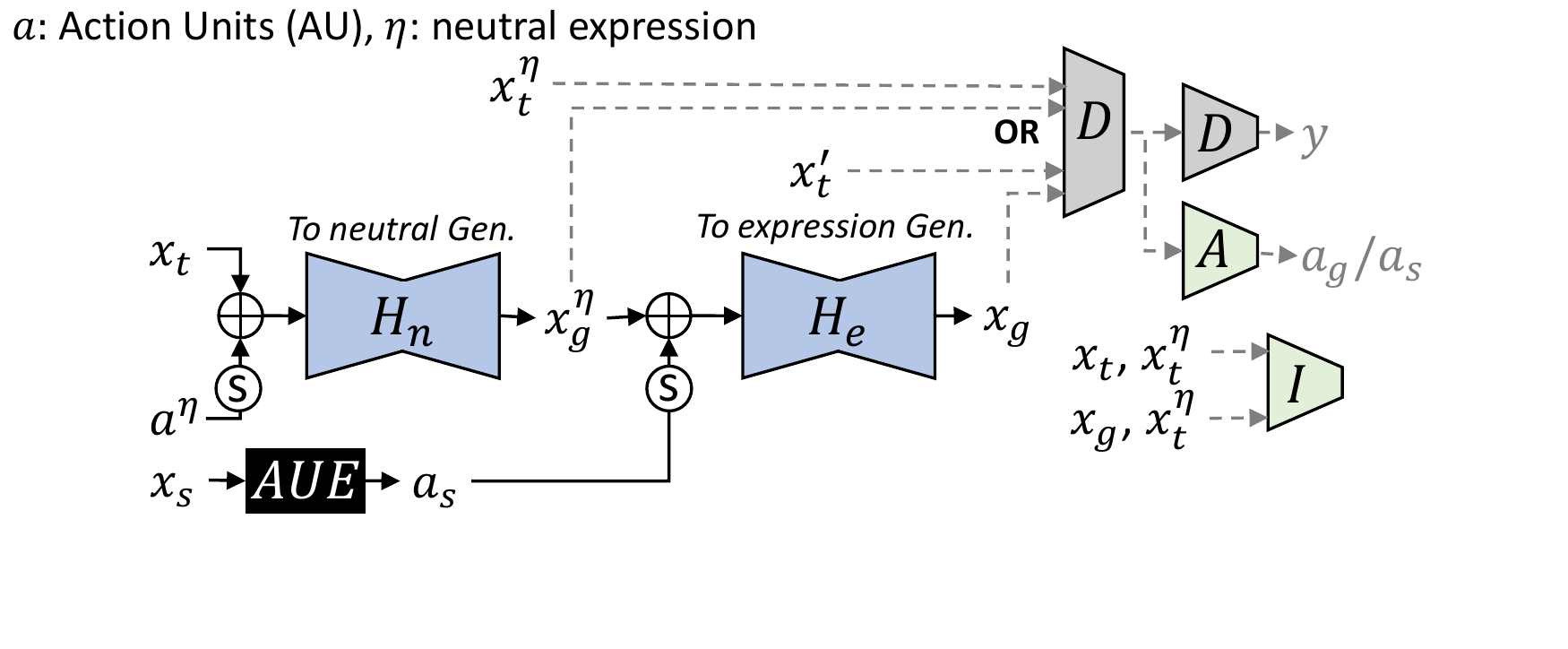}\vspace{-.8em}
		\includegraphics[width=\textwidth]{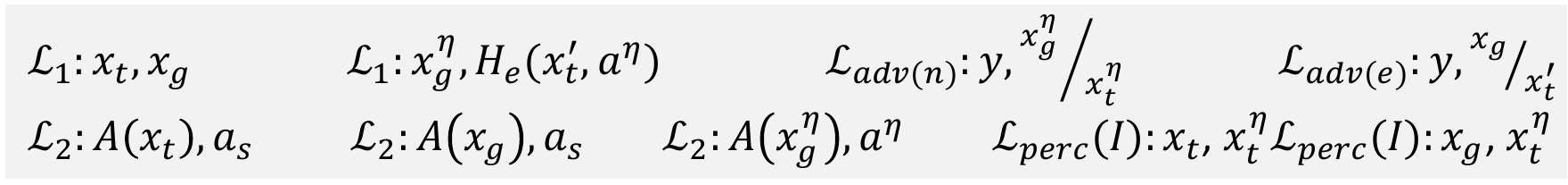}
\end{subfigure}
	\begin{subfigure}[t]{.49\textwidth}	
	\centering
	\caption{\textbf{\cite{qian2019make} AF-VAE:} }
	\includegraphics[width=\textwidth]{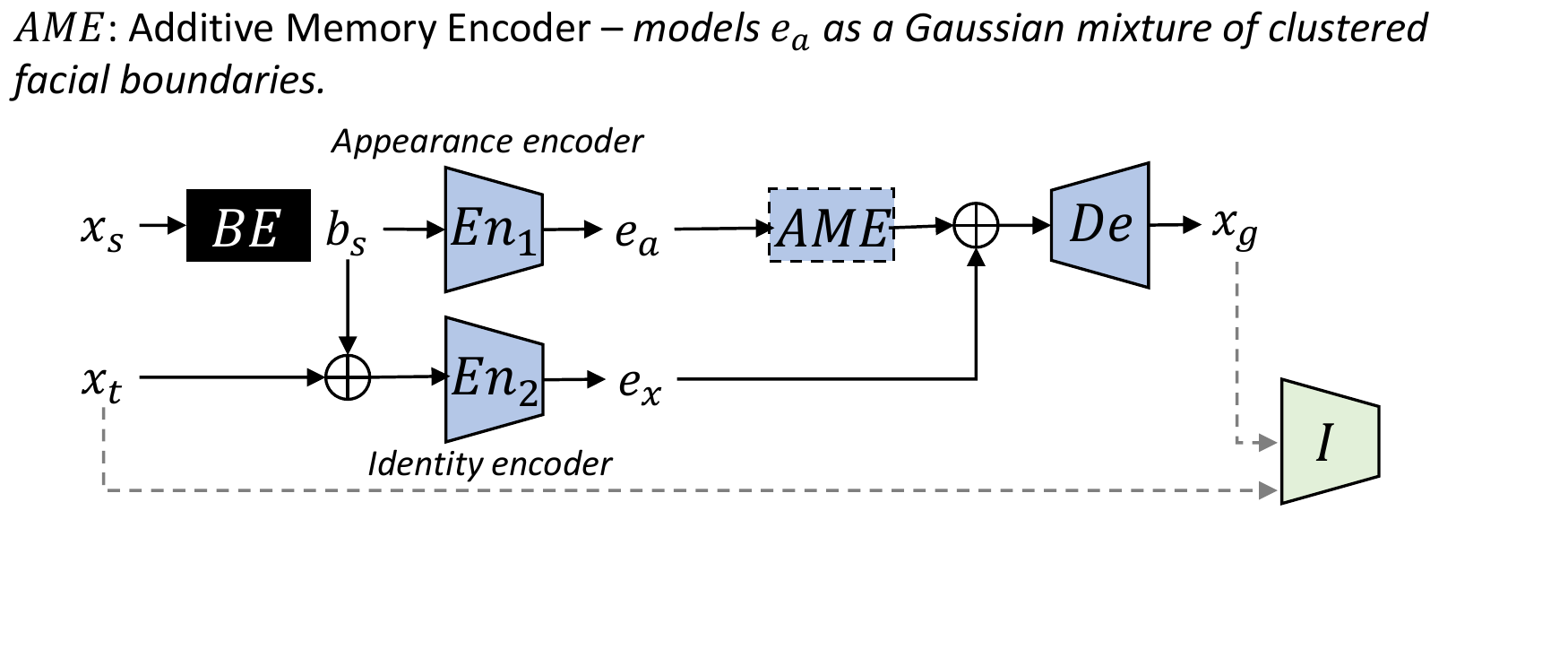}\vspace{-.8em}
		\includegraphics[width=\textwidth]{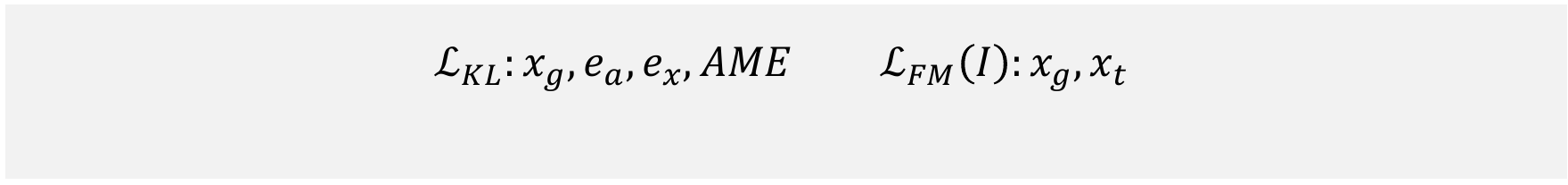}
\end{subfigure}
	\begin{subfigure}[t]{.49\textwidth}	
	\centering
	\caption{\textbf{\cite{geng2019warp} wg-GAN:}}
	\includegraphics[width=\textwidth]{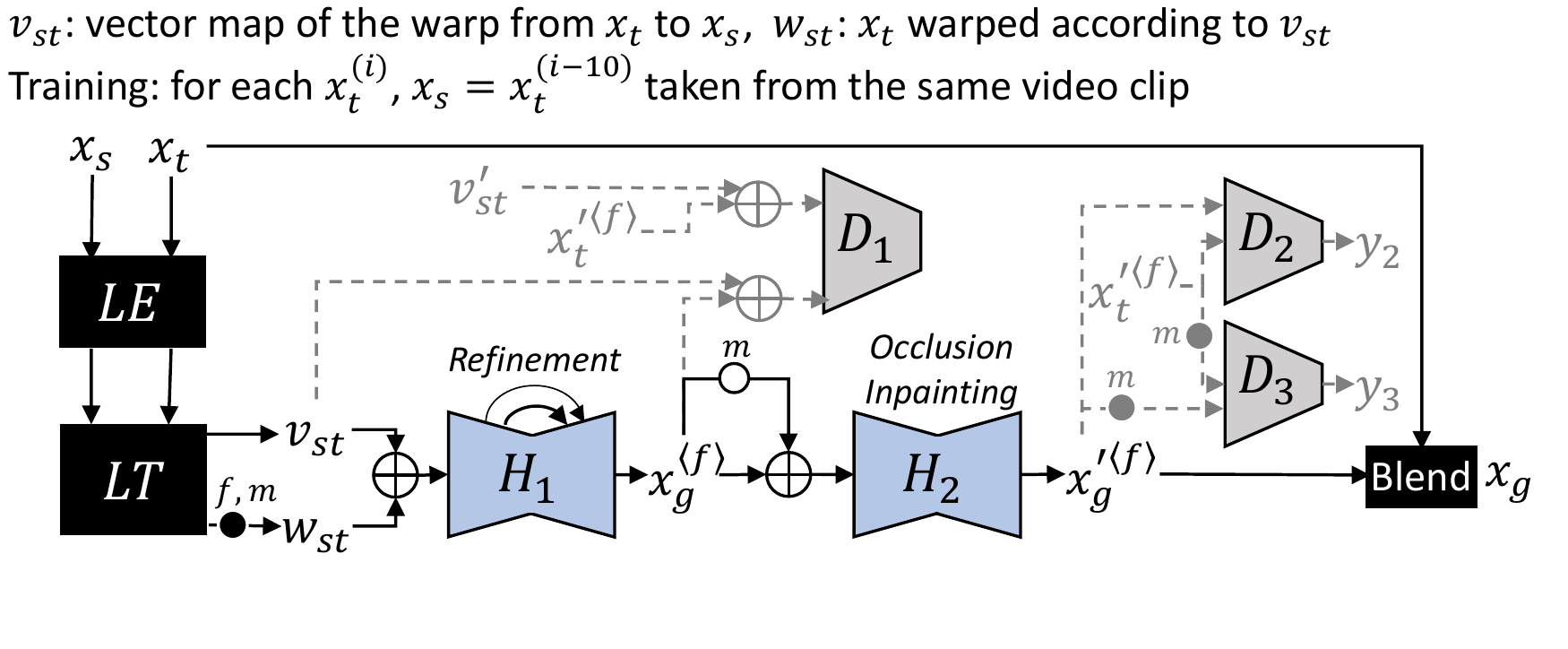}\vspace{-.8em}
		\includegraphics[width=\textwidth]{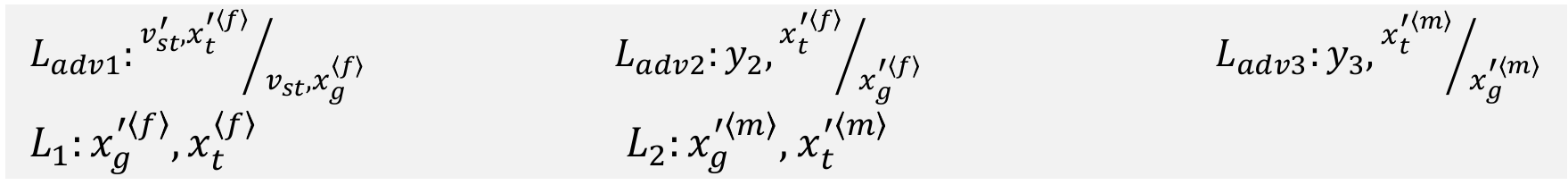}
\end{subfigure}
	\begin{subfigure}[t]{.49\textwidth}	
	\centering
	\caption{\textbf{\cite{otberdout2019dynamic} Motion\&Texture-GAN:} }
	\includegraphics[width=\textwidth]{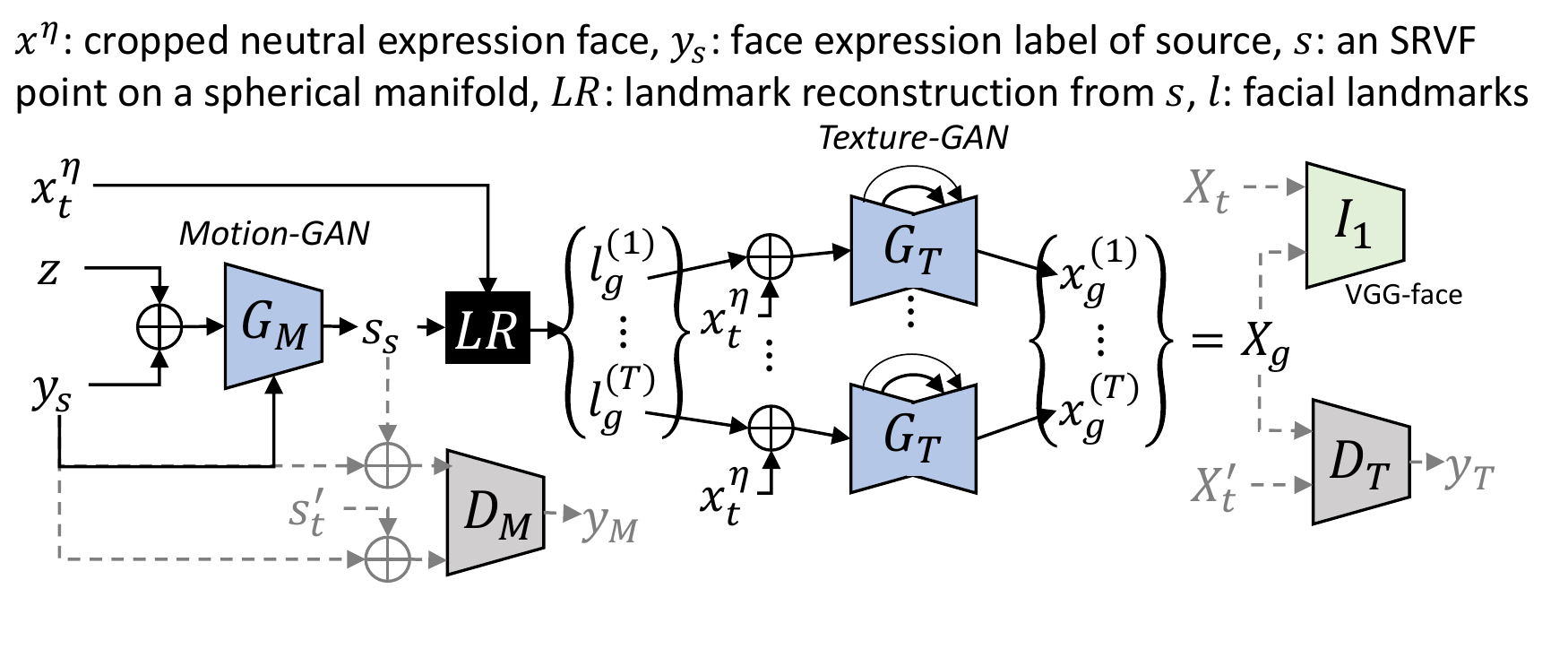}\vspace{-.5em}
		\includegraphics[width=\textwidth]{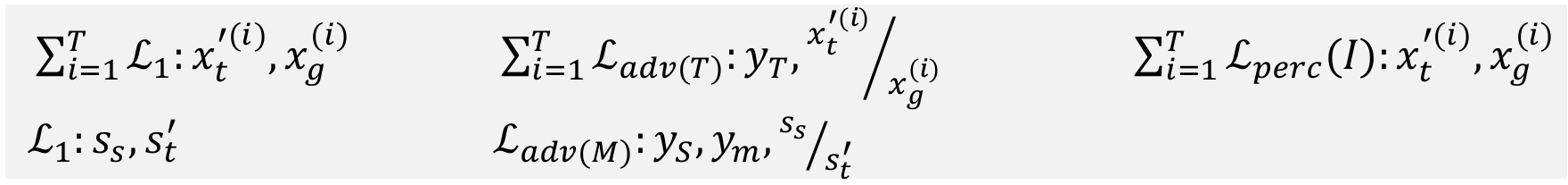}
\end{subfigure}
	\begin{subfigure}[t]{.49\textwidth}	
	\centering
	\caption{\textbf{\cite{wang2020imaginator} ImaGINator:} }
	\includegraphics[width=\textwidth]{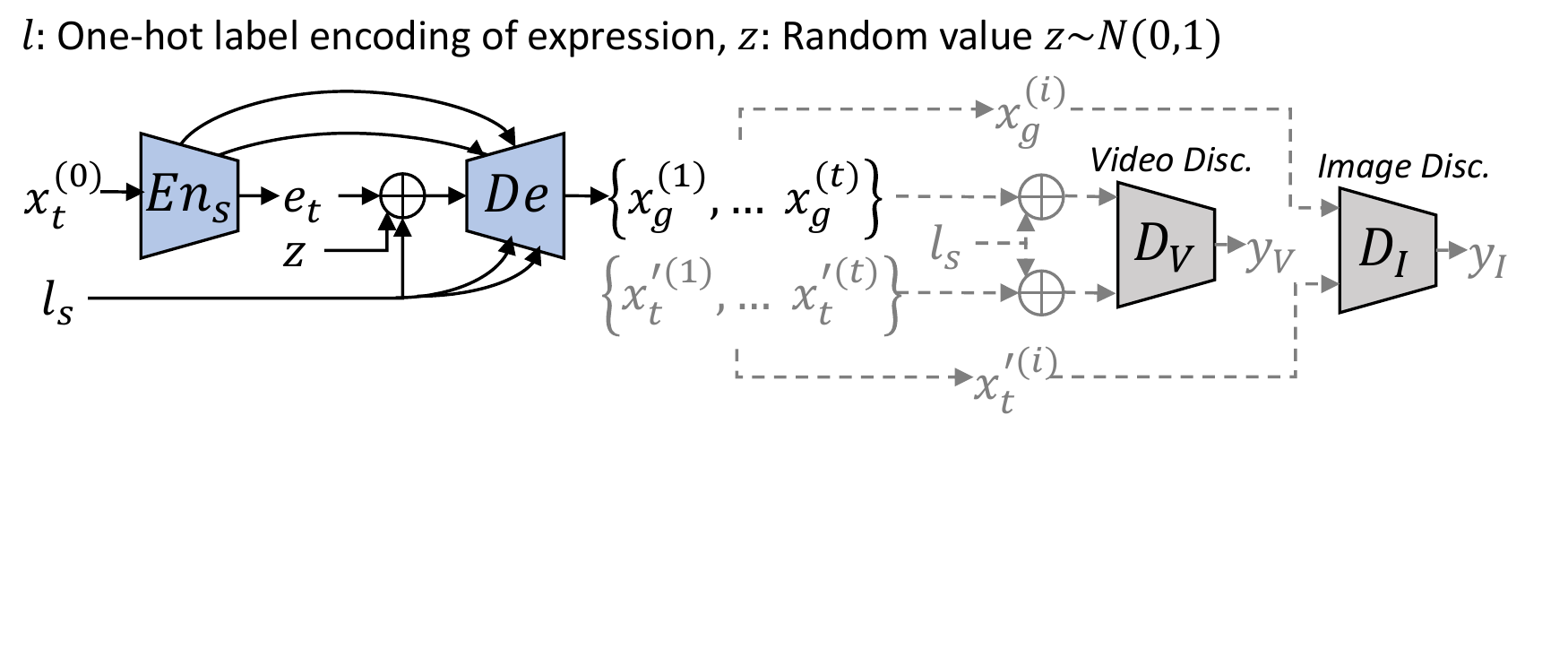}\vspace{-.5em}
		\includegraphics[width=\textwidth]{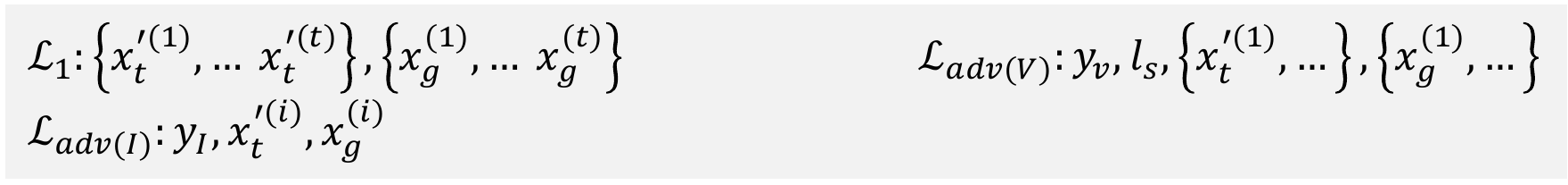}
\end{subfigure}
\vspace{-1em}
	\caption{Architectural schematics of \textbf{reenactment networks}. Black lines indicate prediction flows used during deployment, dashed gray lines indicate dataflows performed during training. Zoom in for more detail.
	}\label{fig:schem_reen1b}
\end{figure}

\begin{figure}	
	\begin{subfigure}[t]{.49\textwidth}	
	\centering
	\caption{\textbf{\cite{siarohin2019animating} Monkey-NET:}}
	\includegraphics[width=\textwidth]{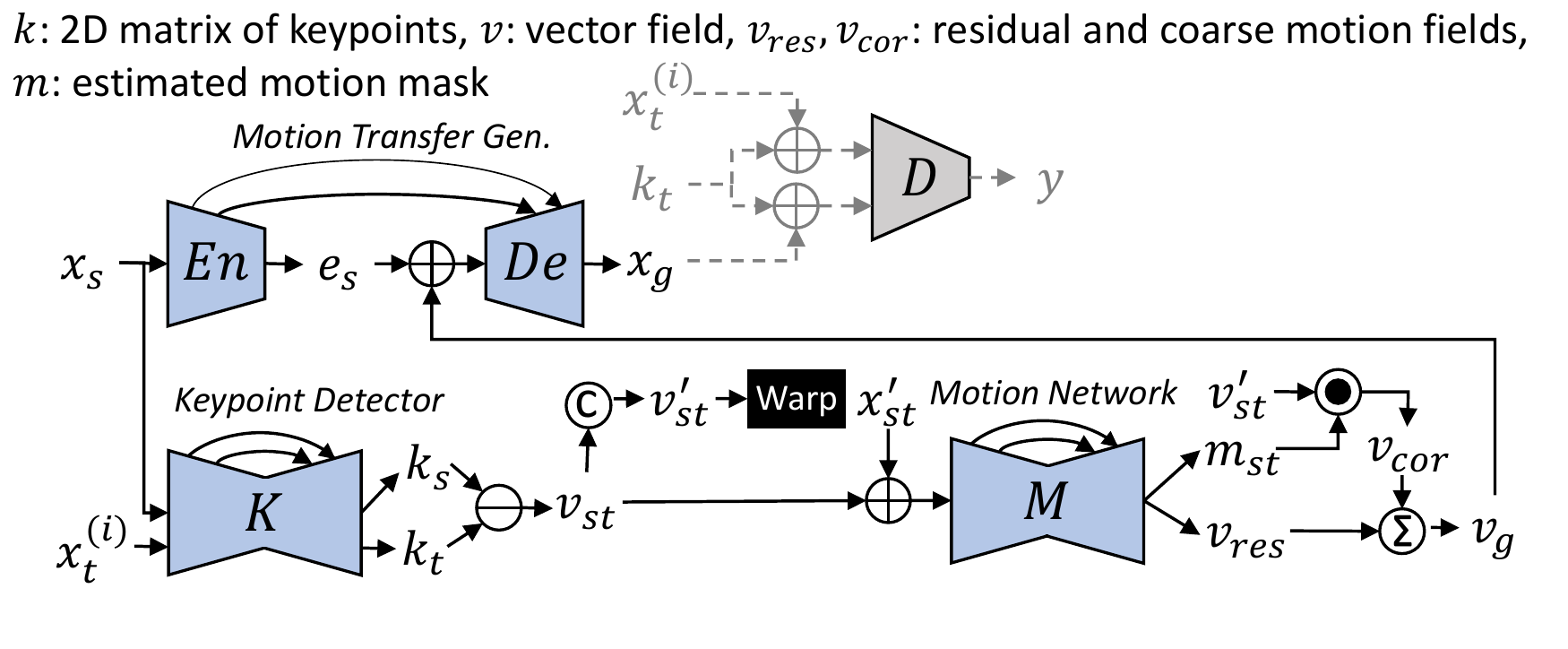}\vspace{-.5em}
		\includegraphics[width=\textwidth]{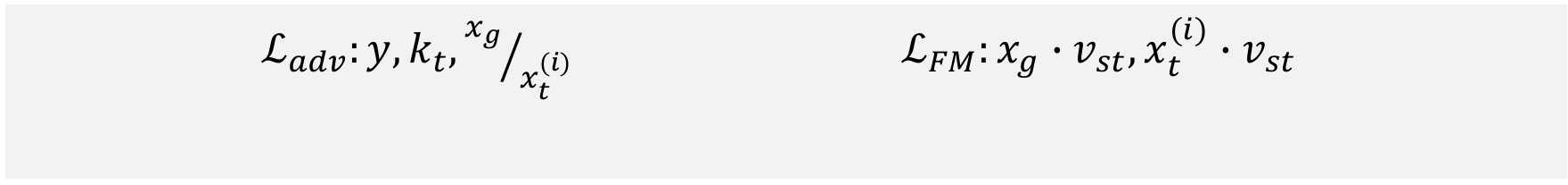}
\end{subfigure}
	\begin{subfigure}[t]{.49\textwidth}	
	\centering
	\caption{\textbf{\cite{zakharov2019few} Neural Talking Heads:} }
	\includegraphics[width=\textwidth]{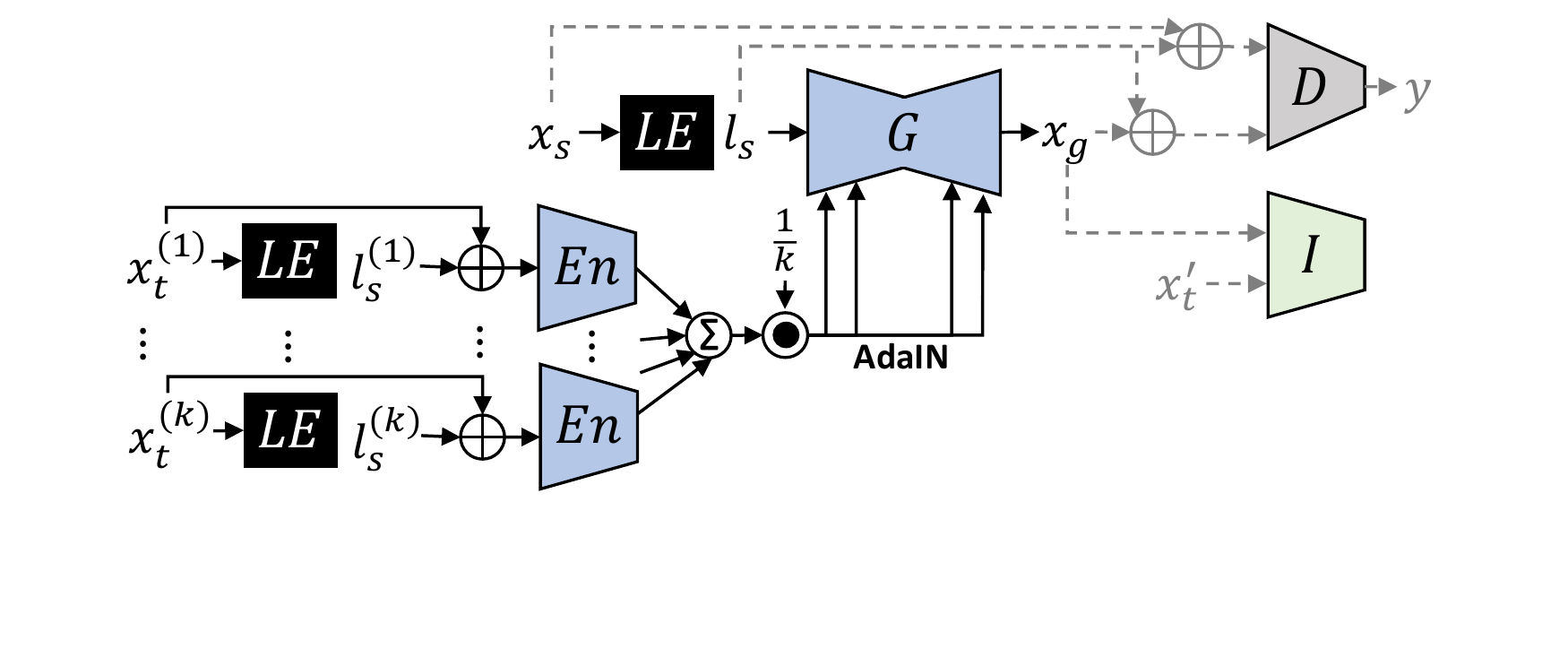}\vspace{-.4em}
		\includegraphics[width=\textwidth]{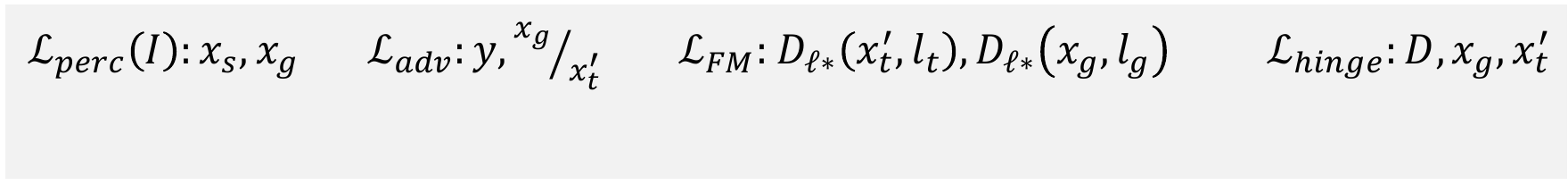}
\end{subfigure}
	\begin{subfigure}[t]{.49\textwidth}	
	\centering
	\caption{\textbf{\cite{MarioNETte:AAAI2020} MarioNETte:}}
	\includegraphics[width=\textwidth]{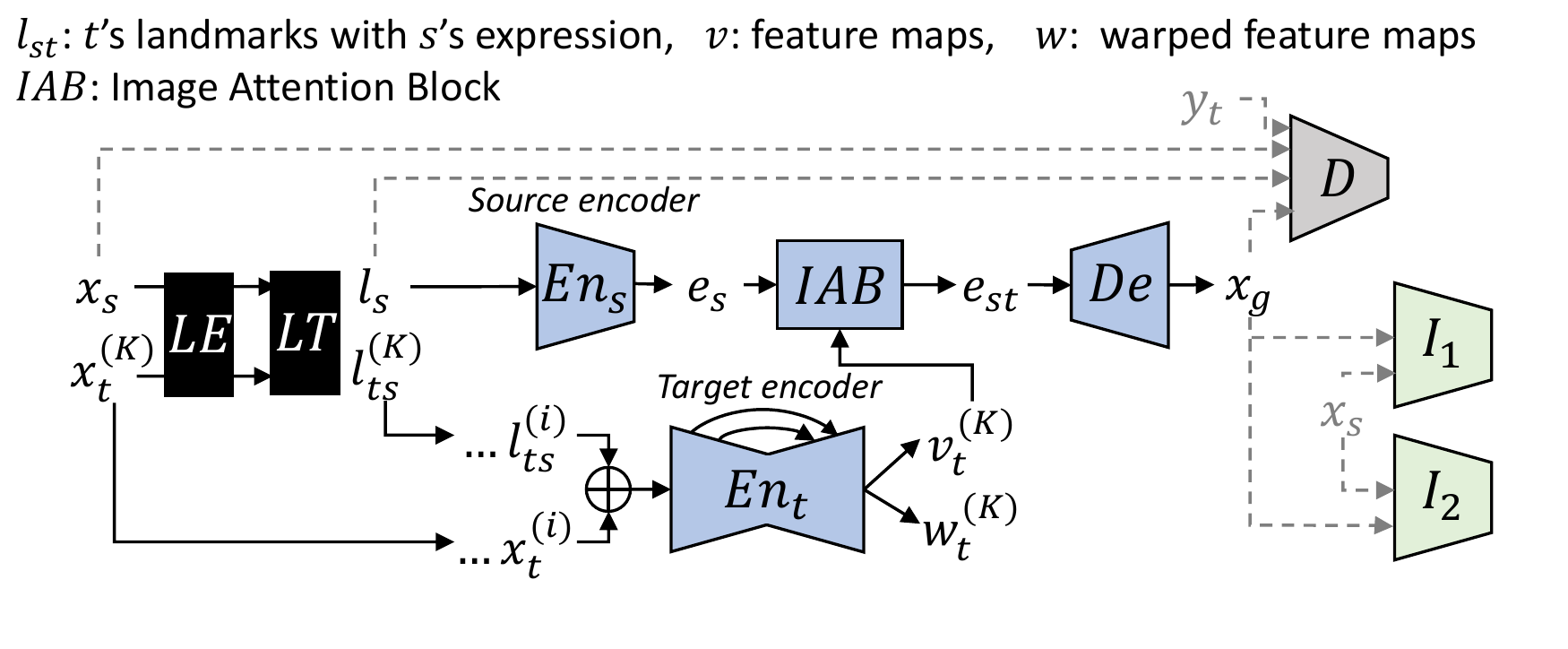}\vspace{-.5em}
		\includegraphics[width=\textwidth]{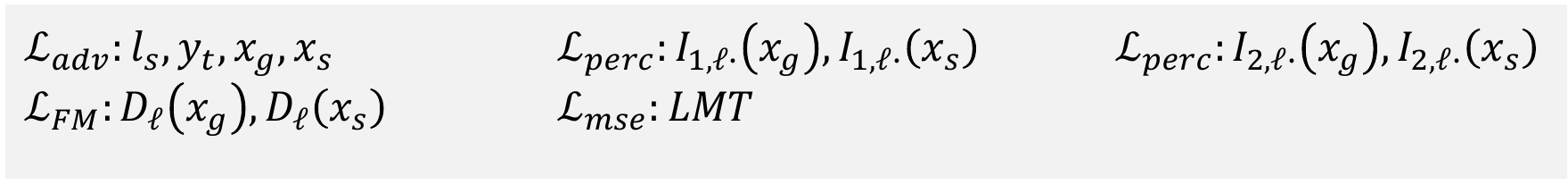}
\end{subfigure}
	\begin{subfigure}[t]{.49\textwidth}	
	\centering
	\caption{\textbf{\cite{liu2019video} Liu et al. 2019:} }
	\includegraphics[width=\textwidth]{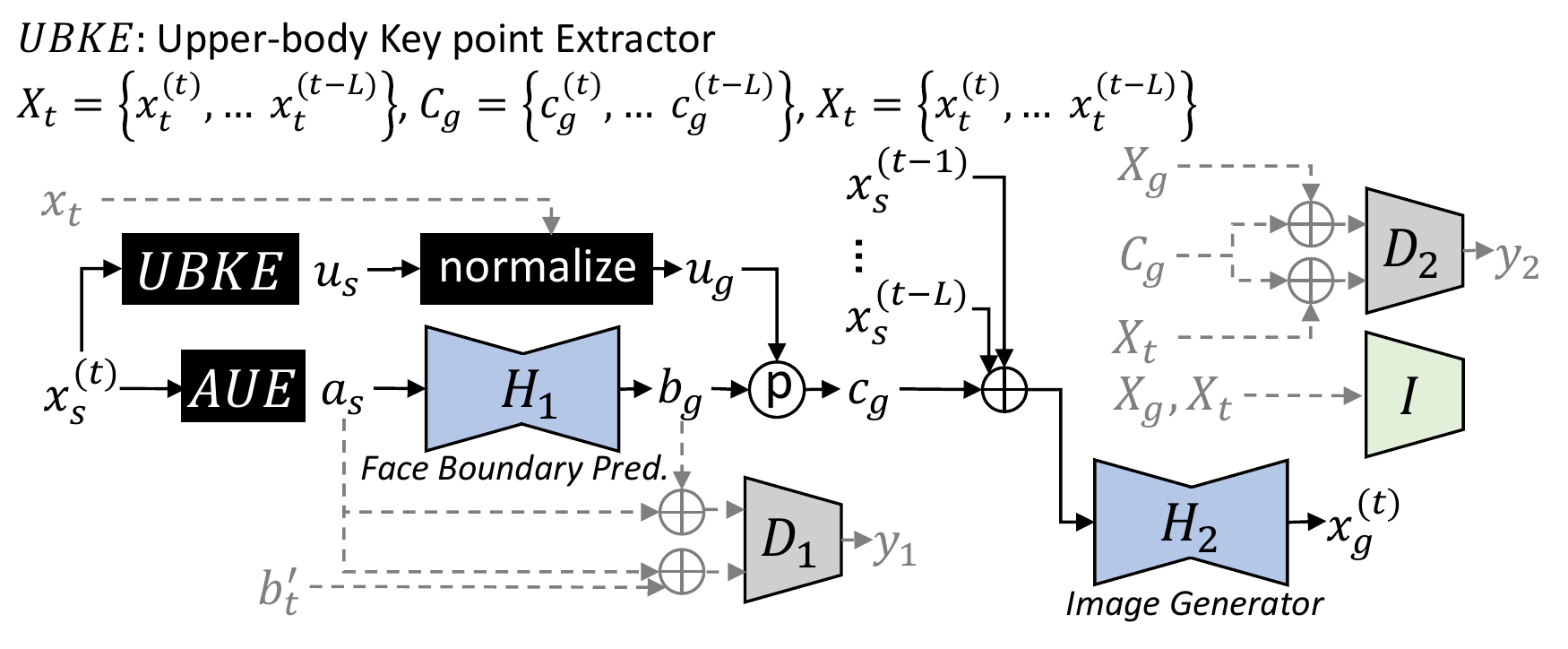}\vspace{-.5em}
		\includegraphics[width=\textwidth]{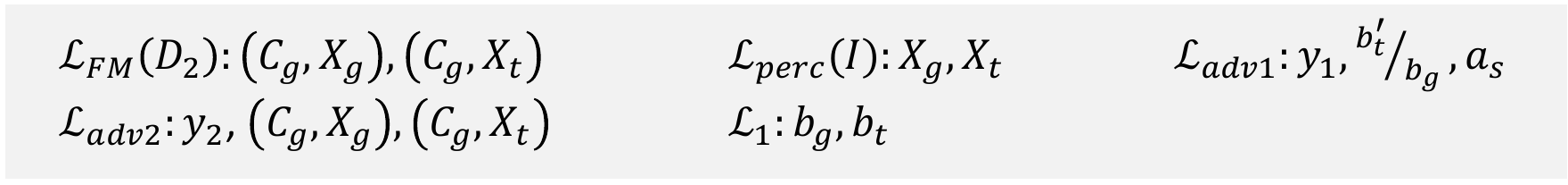}
\end{subfigure}

	\caption{Architectural schematics of the \textbf{reenactment networks}. Black lines indicate prediction flows used during deployment, dashed gray lines indicate dataflows performed during training. Zoom in for more detail.}\label{fig:schem_reen2}
\vspace{-1em}
\end{figure}

	Regardless, today deep learning approaches are recognized as the simplest way to generate believable content. 
	To help the reader understand the networks and follow the text, we provide the model's network schematics and loss functions in figures \ref{fig:schem_reen1}-\ref{fig:schem_reen2}.

\subsubsection{\textbf{One-to-One} (Identity to Identity)}

	In 2017, the authors of \cite{xu2017face} proposed using a CycleGAN for facial reenactment, without the need for data pairing. The two domains where video frames of $s$ and $t$. \blue{However, to avoid artifacts in $x_g$, the authors note that both domains must share a similar distributions (e.g., poses and expressions).}
	
	In 2018, Bansal et al. proposed a generic translation network based on CycleGAN called Recycle-GAN \cite{bansal2018recycle}. Their framework improves temporal coherence and mitigates artifacts by including next-frame predictor networks for each domain. For facial reenactment, the authors train their network to translate the facial landmarks of $x_s$ into portraits of $x_t$.
	\looseness=-1

\subsubsection{\textbf{Many-to-One} (Multiple Identities to a Single Identity)}\label{subsubsec:expression:many2one}
	In 2017, the authors of \cite{bao2017cvae} proposed CVAE-GAN, a conditional VAE-GAN where the generator is conditioned on an attribute vector or class label. \blue{However, reenactment with CVAE-GAN requires manual attribute morphing by interpolating the latent variables (e.g., between target poses).}
	
	Later, in 2018, a large number of source-identity agnostic models were published, each proposing a different method to decoupling $s$ from $t$:\footnote{Although works such as \cite{olszewski2017realistic} and \cite{zhou2017photorealistic} achieved fully agnostic models (many-to-many) in 2017, their works were on low resolution or partial faces.}
	\looseness=-1
	
	\vspace{.5em}\noindent\textbf{Facial Boundary Conversion.}
	One approach was to first convert the structure of source's facial boundaries to that of the target's before passing them through the generator \cite{wu2018reenactgan}. In their framework `ReenactGAN', the authors use a CycleGAN to transform the boundary $b_s$ to the target's face shape as $b_t$ before generating $x_g$ with a pix2pix-like generator.	
	
	\vspace{.5em}\noindent\textbf{Temporal GANs.}
	To improve the temporal coherence of deepfake videos, the authors of \cite{tulyakov2018mocogan} proposed MoCoGAN: a temporal GAN which generates videos while disentangling the motion and content (objects) in the process. Each frame is generated using a target expression label $z_c$, and a motion embedding $z_{M}^{(i)}$ for the $i$-th frame, obtained from a noise seeded RNN. MoCoGAN uses two discriminators, one for realism  (per frame) and one for temporal coherence (on the last $T$ frames). 
	
	In \cite{wang2018vid2vid}, the authors proposed a framework called Vid2Vid, which is similar to pix2pix but for videos. Vid2Vid considers the temporal aspect by generating each frame based on the last $L$ source and generated frames. The model also considers optical flow to perform next-frame occlusion prediction (due to moving objects). Similar to pix2pixHD, a progressive training strategy is to generate high resolution imagery. In their evaluations, the authors demonstrate facial reenactment using the source's facial boundaries. \blue{In comparison to MoCoGAN, Vid2Vid is more practical since it the deepfake is driven by $x_s$ (e.g., an actor) instead of crafted labels. }
	
	\blue{The authors of \cite{kim2018deep} took temporal deepfakes one step further achieving complete facial reenactment (gaze, blinking, pose, mouth, etc.) with only one minute of training video.} Their approach was to extract the source and target's 3D facial models from 2D images using monocular reconstruction, and then for each frame, (1) transfer the facial pose and expression of the source's 3D model to the target's, and (2) produce $x_{g}$ with a modified pix2pix framework, using the last 11 frames of rendered heads, UV maps, and gaze masks as the input.

	
	\captionsetup[subfigure]{labelformat=empty,labelfont=bf,textfont=normalfont,singlelinecheck=off,justification=raggedright}

	\renewcommand{\thesubfigure}{\arabic{subfigure}}

	\subsubsection{\textbf{Many-to-Many} (Multiple IDs to Multiple IDs)}\label{subsubsec:expression:many2many}
	\hfill \\
	\noindent\textbf{Label Driven Reenactment.}
	The first attempts at identity agnostic models were made in 2017, where the authors of \cite{olszewski2017realistic} used a conditional GAN (CGAN) for the task. Their approach was to (1) extract the inner-face regions as $(x_t,x_s)$, and then (2) pass them to an ED to produce $x_g$ subjected to $\mathcal{L}_{1}$ and $\mathcal{L}_{adv}$ losses. \blue{The challenge of using a CGAN was that the training data had to be paired (images of different identities with the same expression).} 
	
	Going one step further, in \cite{zhou2017photorealistic} the authors reenacted full portraits at low resolutions. Their approach was to decoupling the identities was to use a conditional adversarial autoencoder to disentangle the identity from the expression in the latent space. \blue{However, their approach is limited to driving $x_t$ with discreet AU expression labels (fixed expressions) that capture $x_s$.} 
	A similar label based reenactment was presented in the evaluation of StarGAN \cite{choi2018stargan}; an architecture similar to CycleGAN but for $N$ domains (poses, expressions, etc).
	
	Later, in 2018, the authors of \cite{pham2018generative} proposed GATH which can drive $x_t$ using continuous action units (AU) as an input, extracted from $x_s$. \blue{Using continuous AUs enables smoother reenactments over previous approaches \cite{olszewski2017realistic,zhou2017photorealistic,choi2018stargan}.} Their generator is ED network trained on the loss signals from using three other networks: (1) a discriminator, (2) an identity classifier, and (3) a pretrained AU estimator. The classifier shares the same hidden weights as the discriminator to disentangle the identity from the expressions.

	\vspace{.5em}\noindent\textbf{Self-Attention Modeling.}
	Similar to \cite{pham2018generative}, another work called GANimation \cite{pumarola2019ganimation} reenacts faces through AU value inputs estimated from $x_s$. Their architecture uses an AU based generator that uses a self attention model to handle occlusions, and mitigate other artifacts. Furthermore, another network penalizes $G$ with an expression prediction loss, and shares its weights with the discriminator to encourage realistic expressions. Similar to CycleGAN, GANimation uses a cycle consistency loss which eliminates the need for image pairing.	
	
	\blue{Instead of relying on AU estimations, the authors of \cite{sanchez2018triple} propose GANnotation which uses facial landmark images. Doing so enables the network to learn facial structure directly from the input but is more susceptible to identity leakage compared to AUs which are normalized.} GANotation generates $x_g$ based on $(x_t,l_s)$, where $l_s$ is the facial landmarks of $x_s$. The model uses the same self attention model as GANimation, but proposes a novel ``triple consistency loss'' to minimize artifacts in $x_g$. The loss teaches the network how to deal with intermediate poses/expressions not found in the training set. Given $l_s,l_t$ and $l_z$ sampled randomly from the same video, the loss is computed as 
\begin{equation}
	\mathcal{L}_{trip}=\|G(x_t,l_s)-G(G(x_t,l_z),l_s)\|^2
\end{equation}	

	\vspace{.5em}\noindent\textbf{3D Parametric Approaches.}
	Concurrent to the work of \cite{kim2018deep}, other works also leveraged 3D parametric facial models to prevent identity leakage in the generation process.
	In \cite{shen2018faceid}, the authors propose FaceID-GAN which can reenacts $t$ at oblique poses and high resolution. Their ED generator is trained in tandem with a 3DMM face model predictor, where the model parameters of $x_t$ are used to transform $x_s$ before being joined with the encoder's embedding. Furthermore, to prevent identity leakage from $x_s$ to $x_g$, FaceID-GAN incorporates an identification classifier within the adversarial game. The classifier has $2N$ outputs where the first $N$ outputs (corresponding to training set identities) are activated if the input is real and the rest are activated if it's fake.
	
	Later, the authors of \cite{shen2018faceid} proposed FaceFeat-GAN which improves the diversty of the faces while preserving the identity \cite{shen2018facefeat}. The approach is to use a set of GANs to learn facial feature distributions as encodings, and then use these generators to create new content with a decoder. 
	Concretely, three encoder/predictor neural networks $P$, $Q$, and $I$, are trained on real images to extract feature vectors from portraits. $P$ predicts 3DMM parameters $p$, $Q$ encodes the image as $q$ capturing general facial features using feedback from $I$, and $I$ is an identity classifier trained to predict label $y_i$. Next two GANs, seeded with noise vectors, produce $p'$ and $q'$ while a third GAN is trained to reconstruct $x_t$ from $(p,q,y_i)$ and $x_g$ from $(p',q',y_i)$. To reenact $x_t$, (1) $y_t$ is predicted using $I$ (even if the identity was previously unseen), (2) $z_p$ and $z_q$ are selected empirically to fit $x_s$, and (3) the third GAN's generator uses $(p',q',y_t)$ to create $x_g$. 
	\blue{Although FaceFeat-GAN improves image diversty, it is less practical than FaceID-GAN since the GAN's input seed $z$ be selected empirically to fit $x_s$.} 

	In \cite{nagano2018pagan}, the authors present paGAN, a method for complete facial reenactment of a 3D avatar, using a single image of the target as input. An expression neutral image of $x_t$ is used to generate a 3D model which is then driven by $x_s$. The driven model is used to create inputs for a U-Net generator: the rendered head, its UV map, its depth map, a masked image of $x_t$ for texture, and a 2D mask indicating the gaze of $x_s$. \blue{Although paGAN is very efficient, the final deepfake is 3D rendered which detracts from the realism.}
	
	\vspace{.5em}\noindent\textbf{Using Multi-Modal Sources.}
	In \cite{wiles2018x2face} the authors propose X2Face which can reenact $x_t$ with $x_s$ or some other modality such as audio or a pose vector. X2Face uses two ED networks: an embedding network and a driving network. First the embedding network encodes 1-3 examples of the target's face to $v_t$: the optical flow field required to transform $x_t$ to a neutral pose and expression. Next, $x_t$ is interpolated according to $m_t$ producing $x_{t}^{'}$. Finally, the driving network maps $x_s$ to the vector map $v_s$, crafted to interpolate $x_{t}^{'}$ to $x_g$, having the pose and expression of $x_s$. During training, first $\mathcal{L}_{1}$ loss is used between $x_t$ and $x_g$, and then an identity loss is used between $x_s$ and $x_g$ using a pre-trained identity model trained on the VGG-Face Dataset. All interpolation is performed with a tensorflow interpolation layer to enable back propagation using $x_{t}^{'}$ and $x_g$. The authors also show how the embedding of driving network can be mapped to other modalities such as audio and pose.	
	\looseness=-1

	In 2019, nearly all works pursued identity agnostic models: 	\looseness=-1

	\vspace{.5em}\noindent\textbf{Facial Landmark \& Boundary Conversion.}
	In \cite{zhang2019faceswapnet}, the authors propose FaceSwapNet which tries to mitigate the issue of identity leakage from facial landmarks.  First two encoders and a decoder are used to transfer the expression in landmark $l_s$ to the face structure of $l_t$, denoted $l_g$. Then a generator network is used to convert $x_t$ to $x_g$ where $l_g$ is injected into the network with AdaIn layers like a Style-GAN. The authors found that it is crucial to use triplet perceptual loss with an external VGG network. 
	
	In \cite{fu2019high}, the authors propose a method for high resolution reenactment and at oblique angles. A set of networks encode the source's pose, expression, and the target's facial boundary for a decoder that generates the reenacted boundary $b_g$. Finally, an ED network generates $x_g$ using an encoding of $x_t$'s texture in its embedding. A multi-scale loss is used to improve quality and the authors utilize a small labeled dataset by training their model in a semi-supervised way.
\looseness=-1

	In \cite{nirkin2019fsgan}, the authors present FSGAN: a face swapping and facial reenactment model which can handle occlusions.	For reenactment a pix2pixHD generator receives $x_t$ and the source's 3D facial landmarks $l_s$, represented as a 256x256x70 image (one channel for each of the 70 landmarks). The output is $x_g$ and its segmentation map $m_g$ with three channels (background, face, and hair). The generator is trained recurrently where each output is passed back as input for several iterations while $l_s$ is interpolated incrementally from $l_s$ to $l_t$. To improve results further, delaunay Triangulation and barycentric coordinate interpolation are used to generate content similar to the target's pose. \blue{In contrast to other facial conversion methods \cite{zhang2019faceswapnet,fu2019high}, FSGAN uses fewer neural networks enabling real time reenactment at 30fps.}

	\vspace{.5em}\noindent\textbf{Latent Space Manipulation.}
	In \cite{tripathy2019icface}, the authors present a model called ICFace where the expression, pose, mouth, eye, and eyebrows of $x_t$ can be driven independently. Their architecture is similar to a CycleGAN in that one generator translates $x_t$ into a neutral expression domain as $x_{t}^{\eta}$ and another generator translates $x_{t}^{\eta}$ into an expression domain as $x_g$. Both generators ar conditioned on the target AU. 

	In \cite{qian2019make} the authors propose an Additive Focal Variational Auto-encoder (AF-VAE) for high quality reenactment. This is accomplished by separating a C-VAE's latent code into an appearance encoding $e_a$ and identity-agnostic expression coding $e_x$. To capture a wide variety of factors in $e_a$ (e.g., age, illumination, complexion, ...), the authors use an additive memory module during training which conditions the latent variables on a Gaussian mixture model, fitted to clustered set of facial boundaries. Subpixel convolutions were used in the decoder to mitigate artifacts and improve fidelity.	
	
	\vspace{.5em}\noindent\textbf{Warp-based Approaches.}
	In the past, facial reenactment was done by warping the image $x_t$ to the landmarks $l_s$ \cite{elor2017bringingPortraits}. In \cite{geng2019warp}, the authors propose wgGAN which uses the same approach but creates high-fidelity facial expressions by refining the image though a series of GANs: one for refining the warped face and another for in-painting the occlusions (eyes and mouth). \blue{A challenge with wgGAN is that the warping process is sensitive to head motion (change in pose).}

	In \cite{zhang2019one}, the authors propose a system \blue{which can also control the gaze}: a decoder generates $x_g$ with an encoding of $x_t$ as the input and a segmentation map of $x_s$ as reenactment guidance via SPADE residual blocks. The authors blend $x_g$ with a warped version, guided by the segmentation, to mitigate artifacts in the background.  
	
	\blue{To overcome issue of occlusions in the eyes and mouth, the authors of \cite{gu2020flnet} use multiple images of $t$ as a reference, in contrast to \cite{geng2019warp} and \cite{zhang2019one} which only use one.}  In their approach (FLNet), the model is provided with $N$ samples of $t$ ($X_t$) having various mouth expressions, along with the landmark deltas between $X_t$ and $x_s$ ($L_t$). Their model is an ED (configured like GANimation \cite{pumarola2019ganimation}) which produces (1) $N$ encodings for a warped $x_g$, (2) an appearance encoding, and (3) a selection (weight) encoding. The encodings are then coverted into images using seperate CNN layers and merged together through masked multiplication. The entire model is trained end-to-end in a self supervised manner using frames of $t$ taken from different videos. 
	\looseness=-1
	
	\vspace{.5em}\noindent\textbf{Motion-Content Disentanglement.}
	In \cite{otberdout2019dynamic} the authors propose a GAN to reenact neutral expression faces with smooth animations. The authors describe the animations as temporal curves in 2D space, summarized as points on a spherical manifold by calculating their square-root velocity function (SRVF). A WGAN is used to complete this distribution given target expression labels, and a pix2pix GAN is used to convert the sequences of reconstructed landmarks into a video frames of the target.
	
	\blue{In contrast to MoCoGAN \cite{tulyakov2018mocogan}, the authors of \cite{wang2020imaginator} propose ImaGINator: a conditional GAN which fuses both motion and content and uses with transposed 3D convolutions to capture the distinct spatio-temporal relationships.} The GAN also uses a temporal discriminator, and to increase diversity, the authors train the temporal discriminator with some videos using the wrong label. 
	
	\blue{A challenge with works such as \cite{otberdout2019dynamic}  and \cite{wang2020imaginator} is that they are label driven and produce videos with a set number of frames. This makes the deepfake creation process manual and less practical.}	In contrast, the authors of \cite{siarohin2019animating} propose Monkey-Net: a self supervised network for driving an image with an arbitrary video sequence. Similar to MoCoGAN \cite{tulyakov2018mocogan}, the authors decouple the source's content and motion. First a series of networks produce a motion heat map (optical flow) using the source and target's key-points, and then an ED generator produces $x_g$ using $x_s$ and the optical flow (in its embedding).
	
	Later in \cite{NIPS2019_8935}, the authors extend Monkey-Net by improving the object appearance when large pose transformations occur. They accomplish this by (1) modeling motion around the keypoints using affine transformations, (2) updating the key-point loss function accordingly, and (3) having the motion generator predict an occlusion mask on the preceding frame for in-painting inference. Their work has been implemented as a free real-time reenactment tool for video chats, called Avitarify.\footnote{\url{https://github.com/alievk/avatarify}}

	\subsubsection{\textbf{Few-Shot Learning}} 
	Towards the end of 2019 and into the beginning of 2020, researchers began looking into minimizing the amount of training data further via one-shot and few-shot learning. 
	
	In \cite{zakharov2019few}, the authors propose a few-shot model which works well at oblique angles. To accomplish this, the authors perform meta-transfer learning, where the network is first trained on many different identities and then fine-tuned on the target's identity. Then, an identity encoding of $x_t$ is obtained by averaging the encodings of $k$ sets of $(x_t,l_t)$. Then a pix2pix GAN is used to generate $x_g$ using $l_s$ as an input, and the identity encoding via AdaIN layers. Unfortunately, the authors note that their method is sensitive to identity leakage.
	
	In \cite{wang2019fewshotvid2vid} the authors of Vid2Vid (Section \ref{subsubsec:expression:many2one}) extend their work with few-shot learning. They use a network weight generation module which utilizes an attention mechanism. The module learns to extract appearance patterns from a few samples of $x_t$ which are injected into the video synthesis layers. \looseness=-1
	\blue{In contrast to FLNet \cite{gu2020flnet}, \cite{zakharov2019few}, and \cite{wang2019fewshotvid2vid} which merge the multiple representations of $t$ \textit{before} passing it through the generator. This approach is more efficient because it involves fewer passes through the model's networks.}
	
	In \cite{MarioNETte:AAAI2020}, the authors propose MarioNETte which alleviates identity leakage when the pose of $x_s$ is different than $x_t$. \blue{In contrast to other works which encode the identity separately or use of AdaIN layers, the authors use an image attention block and target feature alignment.} This enables the model to better handle the differences between face structures. 
	Finally, the identity is also preserved using a novel landmark transformer inspired by \cite{blanz1999morphable}.

\begin{figure*}	
	\centering
		\begin{subfigure}[t]{.49\textwidth}	
		\centering
		\caption{\textbf{\cite{suwajanakorn2017synthesizing} Synthesizing Obama:} }
		\includegraphics[width=\textwidth]{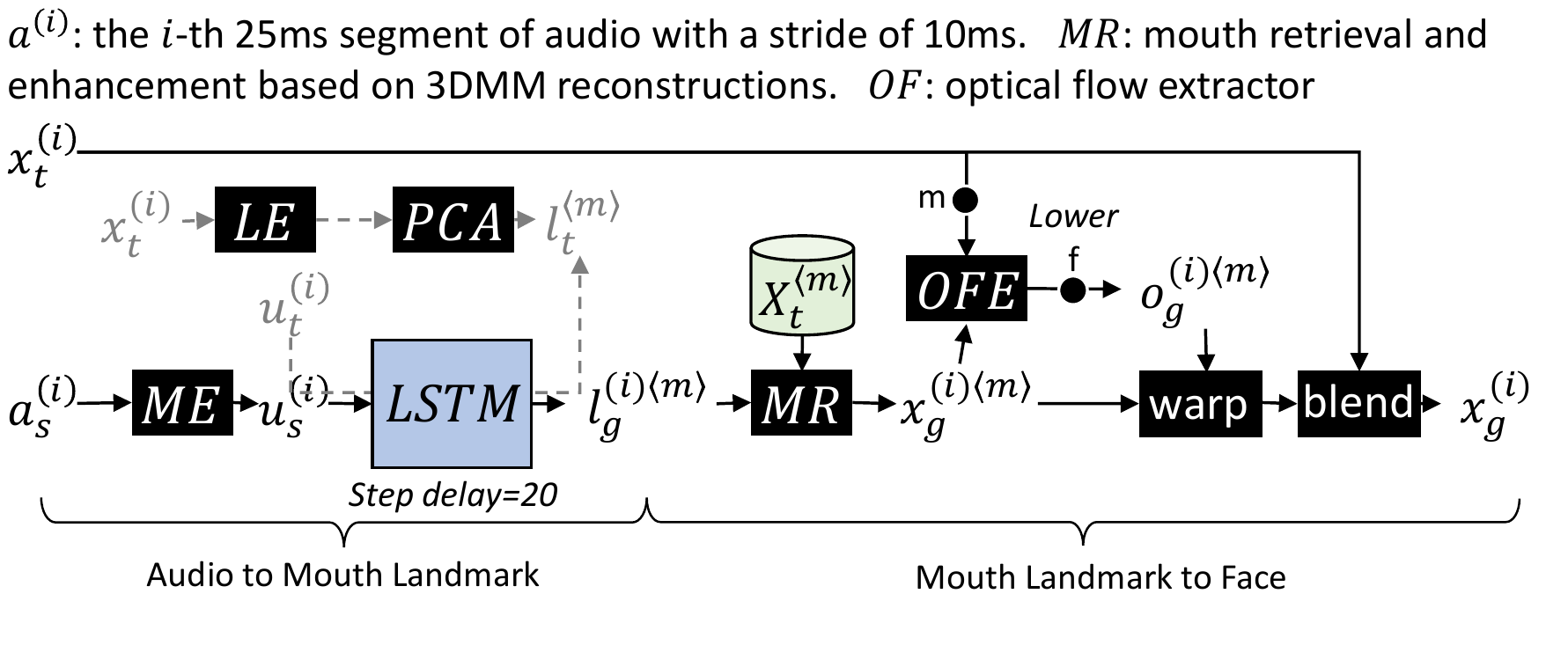}\vspace{-.2em}
		\includegraphics[width=\textwidth]{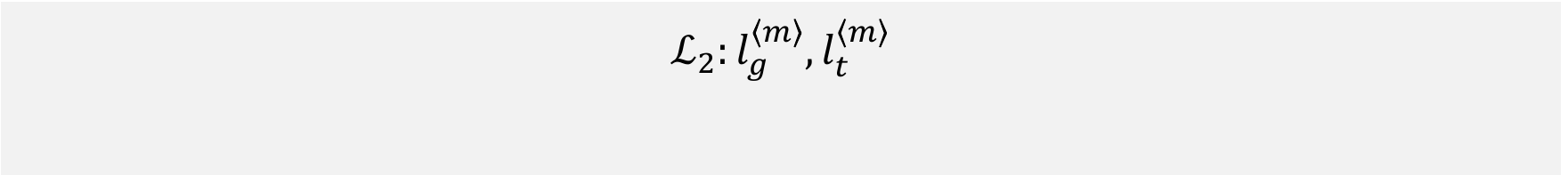}
	\end{subfigure}
	\begin{subfigure}[t]{.49\textwidth}	
	\centering
	\caption{\textbf{\cite{fried2019text} TETH:} }
	\includegraphics[width=\textwidth]{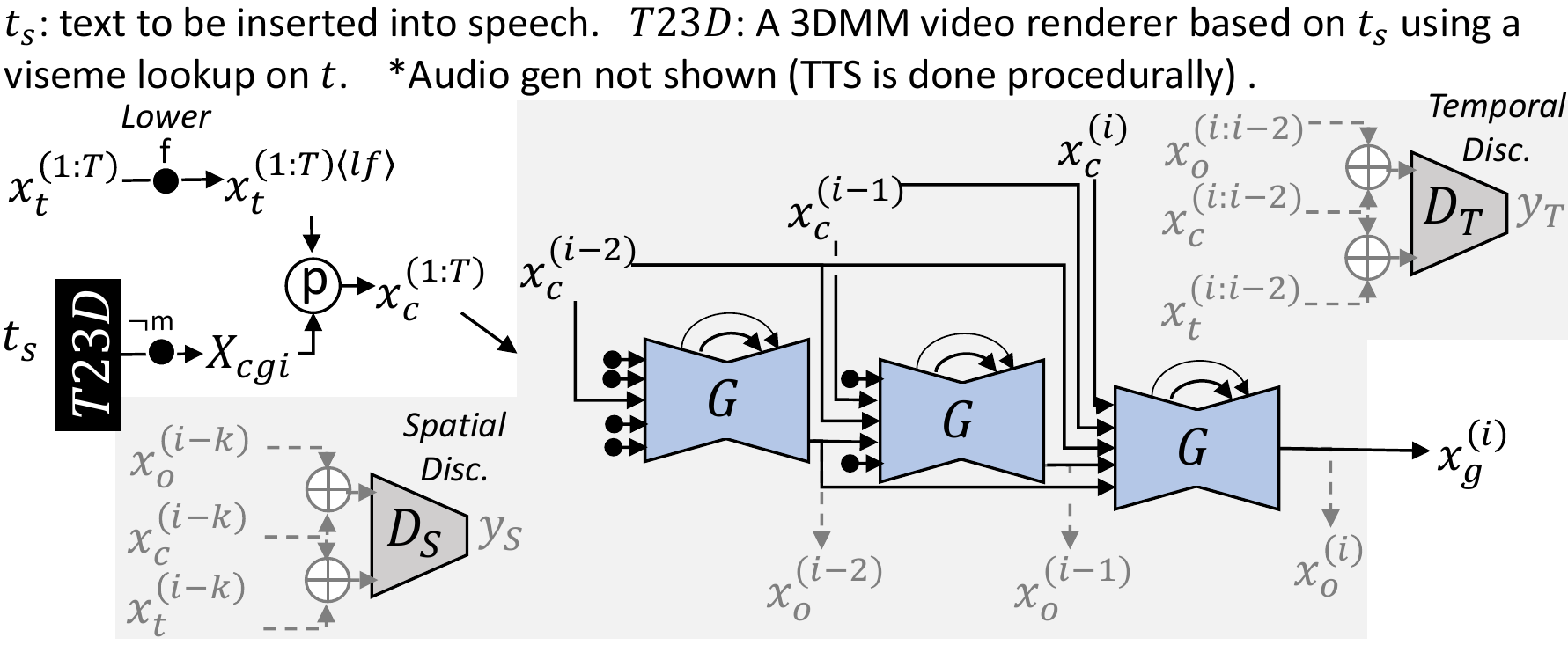}\vspace{0em}
		\includegraphics[width=\textwidth]{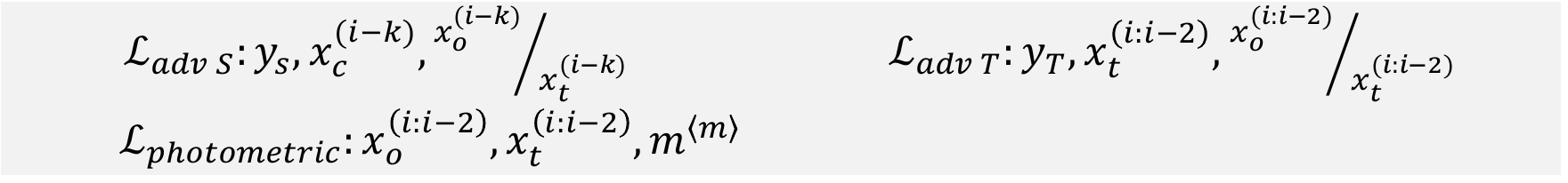}
\end{subfigure}
\begin{subfigure}[t]{.49\textwidth}	
	\centering
	\caption{\textbf{\cite{jalalifar2018speech} SD-CGAN:} }
	\includegraphics[width=\textwidth]{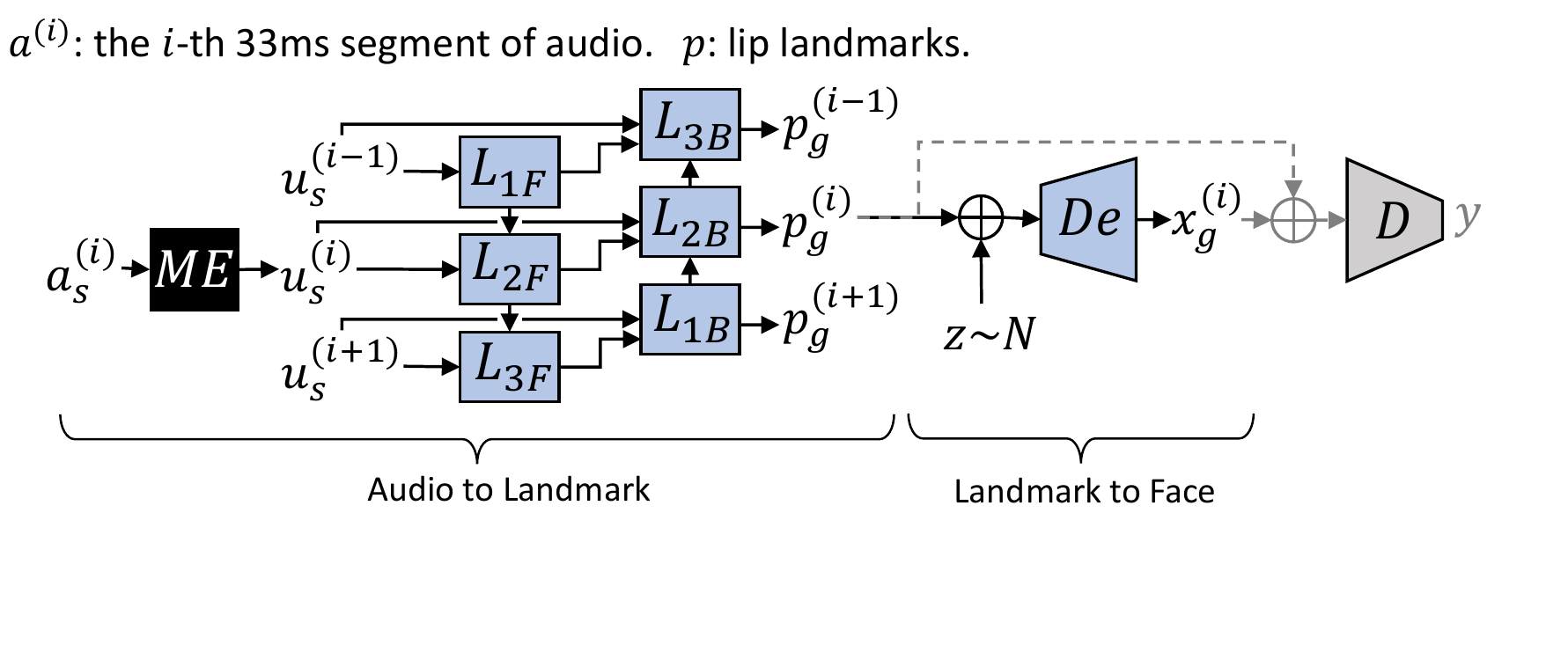}\vspace{-.1em}
		\includegraphics[width=\textwidth]{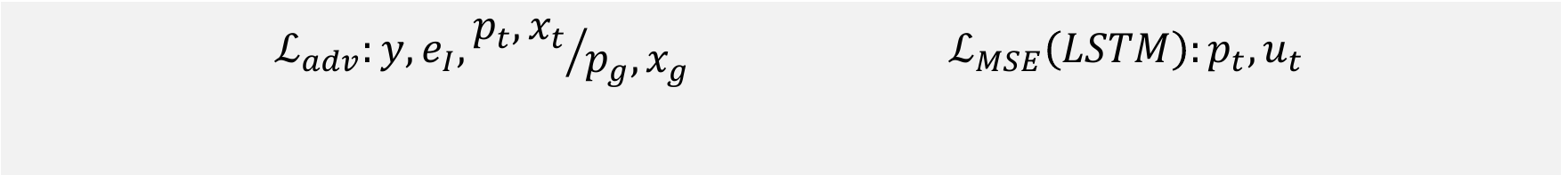}
\end{subfigure}
	\begin{subfigure}[t]{.49\textwidth}	
	\centering
	\caption{\textbf{\cite{thies2019neural} Neural Voice Puppetry:}}
	\includegraphics[width=\textwidth]{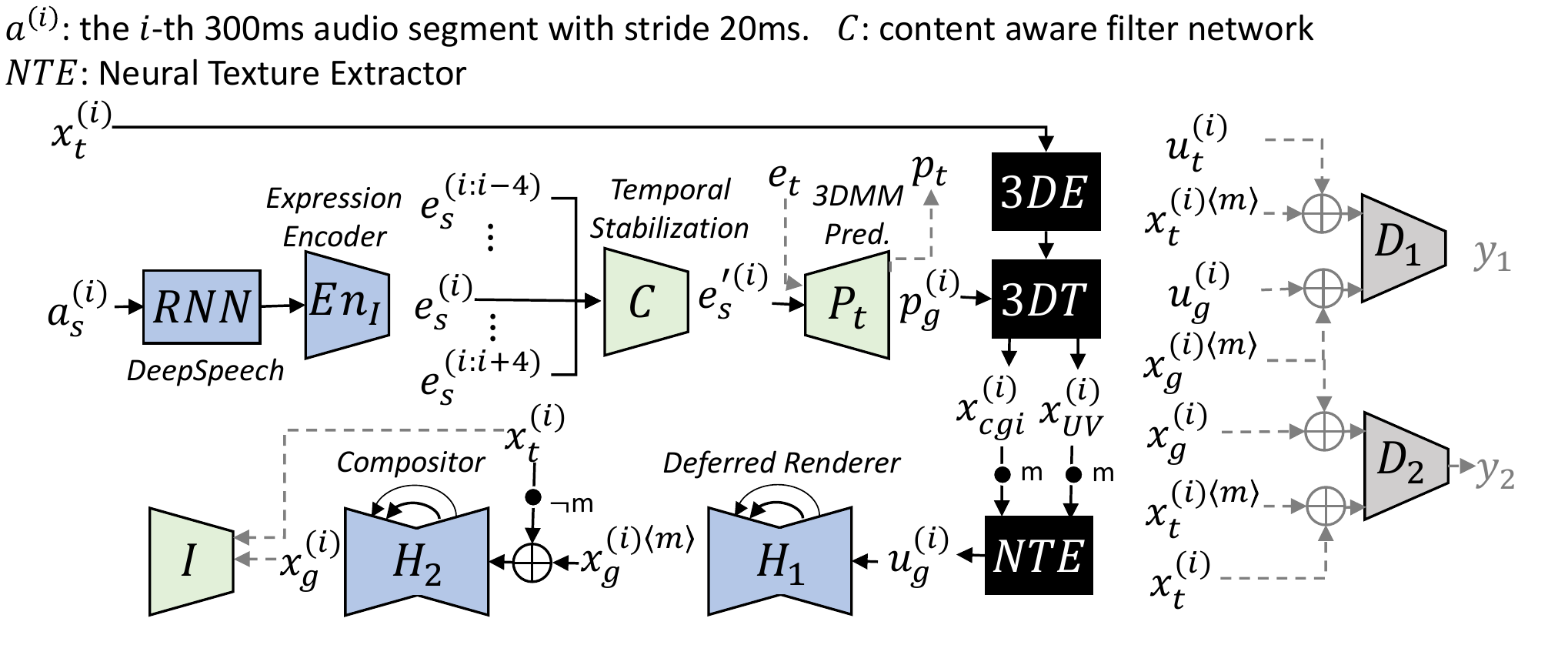}\vspace{-.1em}
		\includegraphics[width=\textwidth]{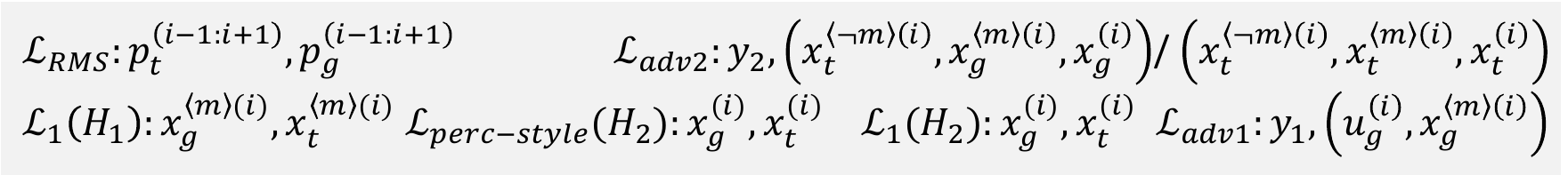}
\end{subfigure}
	\begin{subfigure}[t]{.49\textwidth}	
	\centering
	\caption{\textbf{\cite{chen2019hierarchical} ATVGnet:} }
	\includegraphics[width=\textwidth]{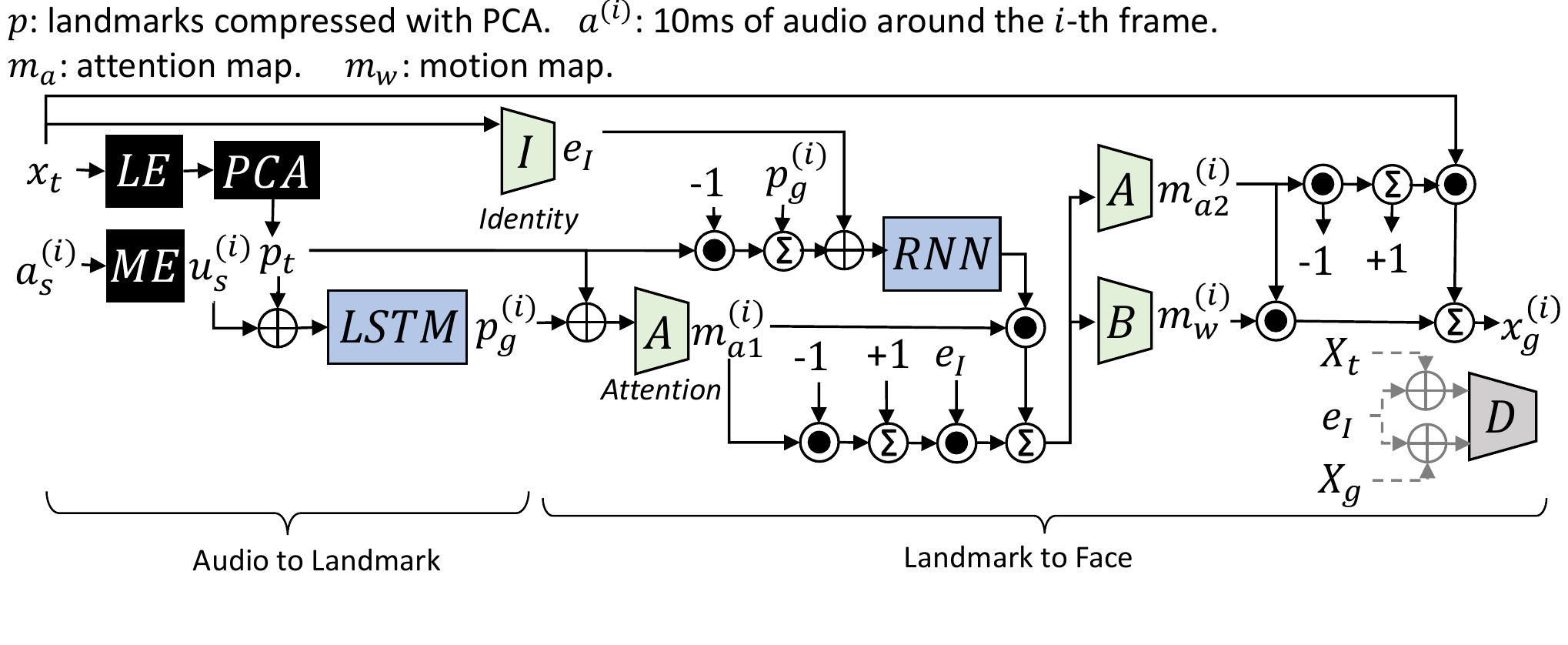}\vspace{-.5em}
		\includegraphics[width=\textwidth]{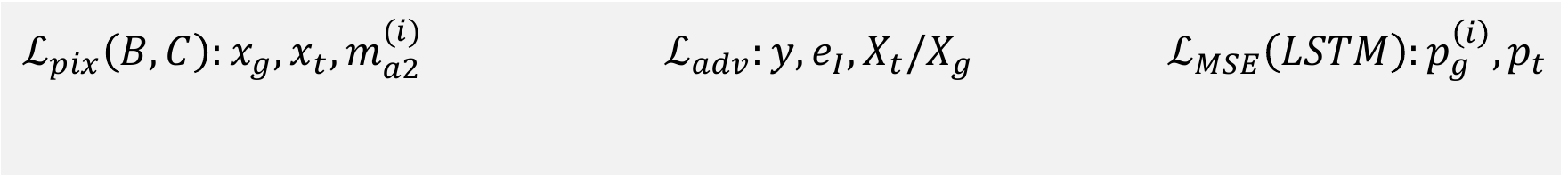}
\end{subfigure}
	\begin{subfigure}[t]{.49\textwidth}	
	\centering
	\caption{\textbf{\cite{zhou2019talking} DAVS:} }
	\includegraphics[width=\textwidth]{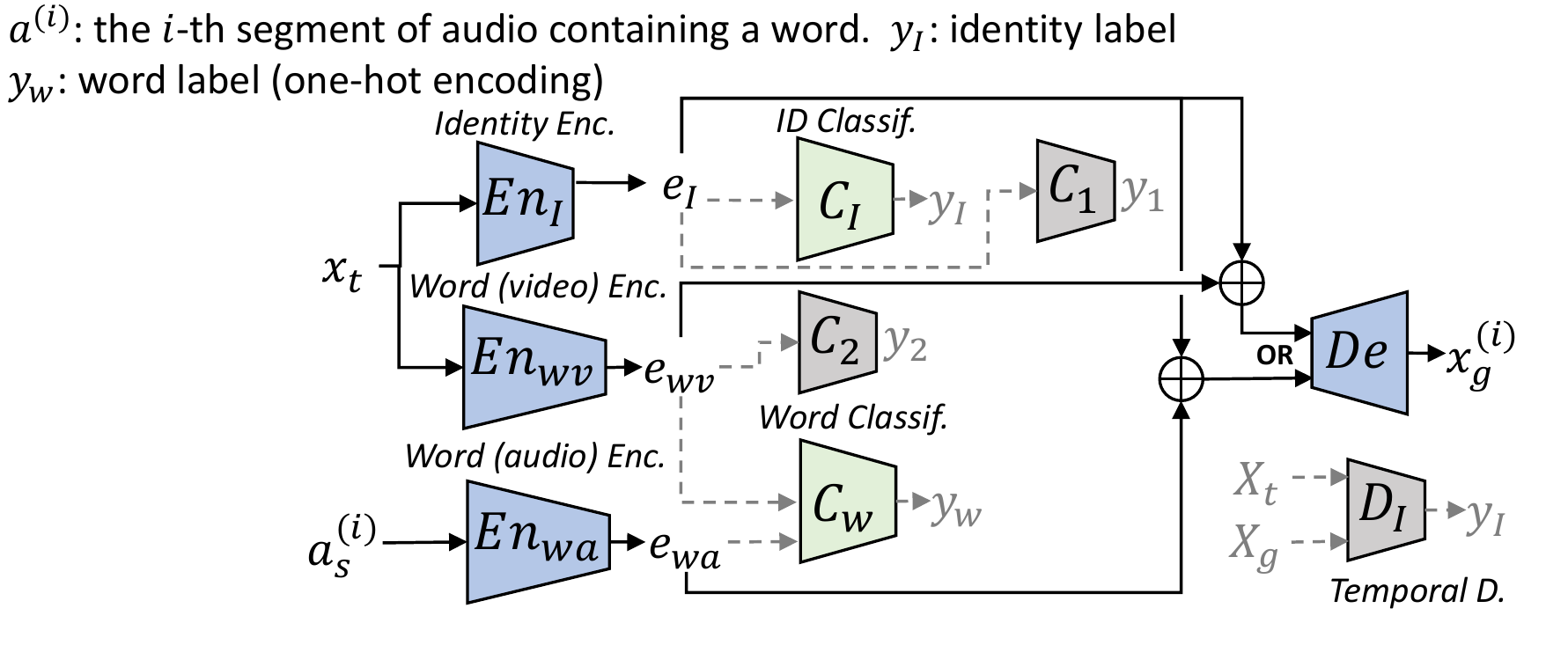}\vspace{-.5em}
		\includegraphics[width=\textwidth]{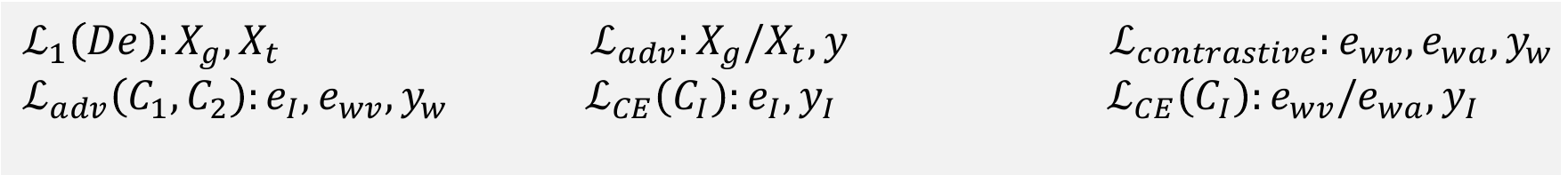}
\end{subfigure}

\begin{subfigure}[t]{.49\textwidth}	
	\centering
	\caption{\textbf{\cite{jamaludin2019you} Speech2Vid:} }
	\includegraphics[width=\textwidth]{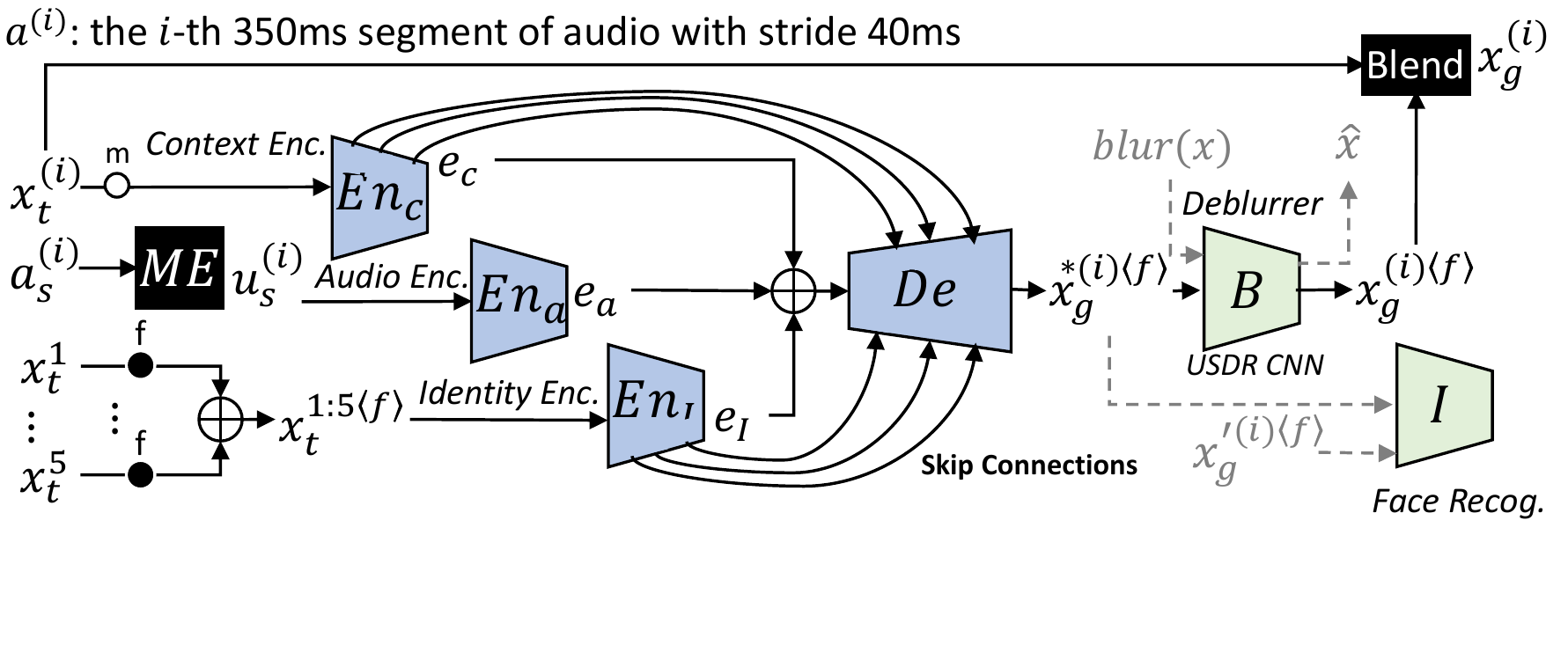}\vspace{-.2em}
		\includegraphics[width=\textwidth]{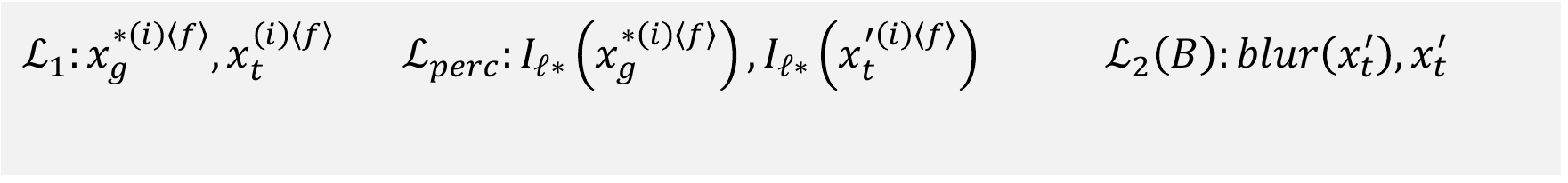}
\end{subfigure}
	\begin{subfigure}[t]{.49\textwidth}	
	\centering
	\caption{\textbf{\cite{vougioukas2019realistic} Speech Driven Animation:}}
	\includegraphics[width=\textwidth]{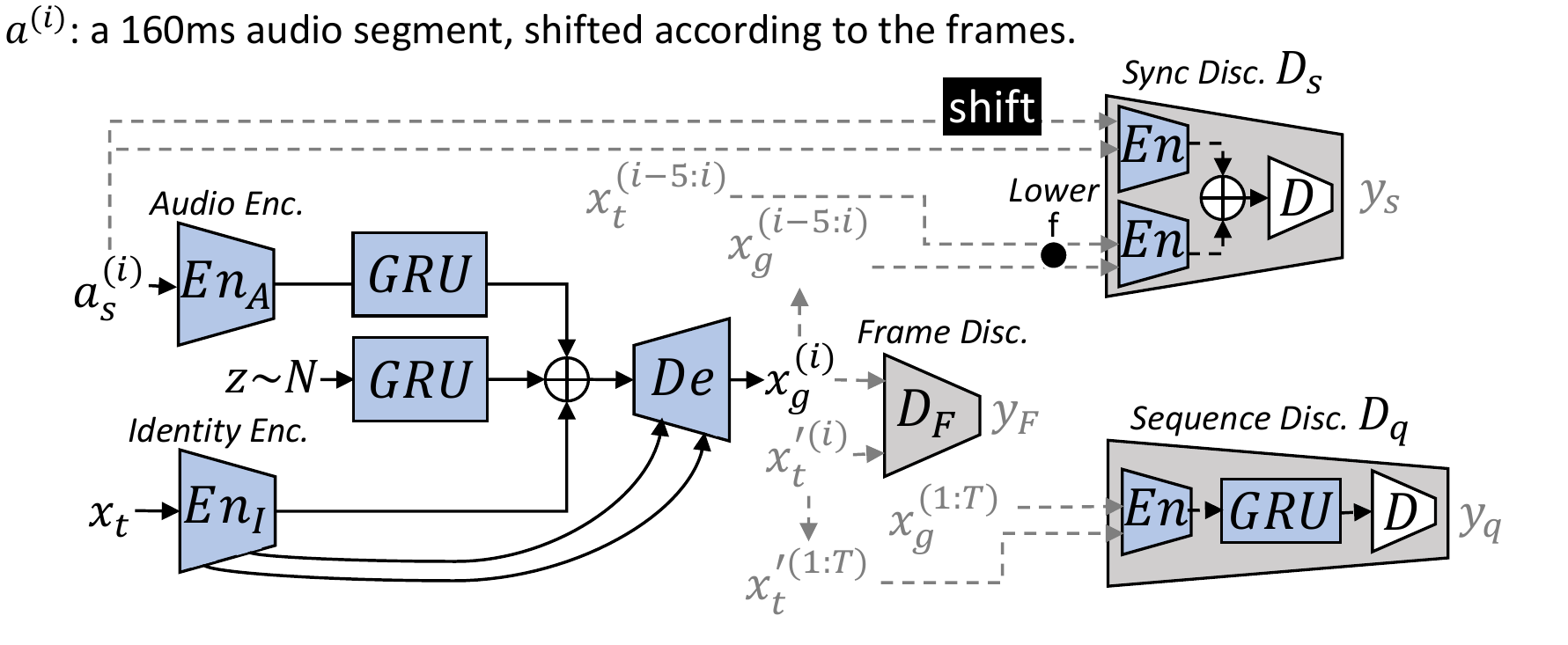}\vspace{-.2em}
		\includegraphics[width=\textwidth]{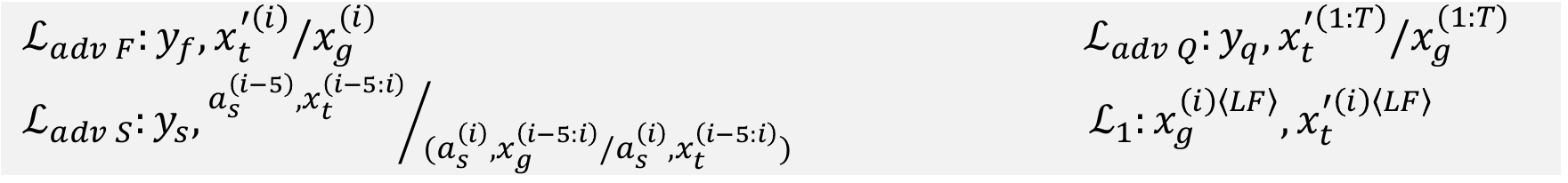}
\end{subfigure}

		\vspace{-1em}
	\caption{Architectural schematics for some \textbf{mouth reenactment networks}. Black lines indicate prediction flows used during deployment, dashed gray lines indicate dataflows performed during training.}\label{fig:schem_reen_mouth}
	\vspace{-2em}
\end{figure*}

	\subsection{Mouth Reenactment (Dubbing)}
	In contrast to expression reenactment, mouth reenactment (a.k.a., video or image dubbing) is concerned with driving a target's mouth with a segment of audio. Fig. \ref{fig:schem_reen_mouth} presents the relevant schematics for this section.
	
	\subsubsection{\textbf{Many-to-One} (Multiple Identities to a Single Identity)} \hfill \\
	\textbf{Obama Puppetry.}
	In 2017, the authors of \cite{suwajanakorn2017synthesizing} created a realistic reenactment of former president Obama. This was accomplished by (1) using a time delayed RNN over MFCC audio segments to generate a sequence of mouth landmarks (shapes), (2) generating the mouth textures (nose and mouth) by applying a weighted median to images with similar mouth shapes via PCA-space similarity, (3) refining the teeth by transferring the high frequency details other frames in the target video, and (4) by using dynamic programming to re-time the target video to match the source audio and blend the texture in.
	
	Later that year, the authors of \cite{kumar2017obamanet} presented ObamaNet: a network that reenacts an individual's mouth and voice \blue{using text as input instead of audio like \cite{suwajanakorn2017synthesizing}}. The process is to (1) convert the source text to audio using Char2Wav \cite{sotelo2017char2wav}, (2) generate a sequence of mouth-keypoints using a time-delayed LSTM on the audio, and (3) use a U-Net CNN to perform in-painting on a composite of the target video frame with a masked mouth and overlayed keypoints.
	
	Later in 2018, Jalalifar et al. \cite{jalalifar2018speech} proposed a network that synthesizes the entire head portrait of Obama, \blue{and therefore does not require pose re-timing and can trained end-to-end, unlike \cite{suwajanakorn2017synthesizing} and \cite{kumar2017obamanet}.} First, a bidirectional LSTM coverts MFCC audio segments into sequence of mouth landmarks, and then a pix2pix like network generates frames using the landmarks and a noise signal. After training, the pix2pix network is fine-tuned using a single video of the target to ensure consistent textures.			\looseness=-1
	
	\vspace{.5em}\noindent\textbf{3D Parametric Approaches.}
	Later on in 2019, the authors of \cite{fried2019text} proposed a method for editing a transcript of a talking heads which, in turn, modifies the target's mouth and speech accordingly.
	The approach is to (1) align phenomes to $a_s$, (2) fit a 3D parametric head model to each frame of $X_t$ like \cite{kim2018deep}, (3) blend matching phenomes to create any new audio content, (4) animate the head model with the respective frames used during the blending process, and (5) generate $X_g$ with a CGAN RNN using composites as inputs (rendered mouths placed over the original frame). 
	
	The authors of \cite{thies2019neural} had a different approach: (1) animate a the reconstructed 3D head with the predicted blend shape parameters from $a_s$ using a DeepSpeech model for feature extraction, (2) use Deferred Neural Rendering \cite{thies2019deferred} to generate the mouth region, and then (3) use a network to blend the mouth into the original frame. \blue{Compared to previous works, the authors found that their approach only requires 2-3 minutes of video while producing very realistic results. This is because neural rendering can summarize textures with a high fidelity and operate on UV maps --mitigating artifacts in how the textures are mapped to the face. }

	\subsubsection{\textbf{Many-to-Many} (Multiple IDs to Multiple IDs)}
	
	One of the first works to perform identity agnostic video dubbing was \cite{shimba2015talking}. There the authors used an LSTM to map MFCC audio segments to the face shape. The face shapes were represented as the coefficients of an active appearance model (AAM), which were then used to retrieve the correct face shape of the target. 
	
	\vspace{.5em}\noindent\textbf{Improvements in Lip-sync.}
	Noting a human's sensitivity to temporal coherence, the authors of \cite{song2018talking} use a GAN with three discriminators: on the frames, video, and lip-sync. Frames are generated by (1) encoding each MFCC audio segment $a_{s}^{(i)}$ and $x_t$ with separate encoders, (2) passing the encodings through an RNN, and (3) decoding the outputs as $x_{g}^{(i)}$ using a decoder.
	
	\blue{In \cite{yu2019mining} the authors try to improve the lipsyncing with a textual context.} A time-delayed LSTM is used to predict mouth landmarks given MFCC segments and the spoken text using a text-to-speech model. The target frames are then converted into sketches using an edge filter and the predicted mouth shapes are composited into them. Finally, a pix2pix like GAN with self-attention is used to generate the frames with both video and image conditional discriminators.
	
	\blue{Compared to direct models such as direct models \cite{song2018talking,yu2019mining}, the authors of \cite{chen2019hierarchical} improve the lip-syncing by preventing the model from learning irrelevant correlations between the audiovisual signal and the speech content.} This was accomplished with LSTM audio-to-landmark network and a landmark-to-identity CNN-RNN used in sequence. There, the facial landmarks are compressed with PCA and the attention mechanism from \cite{pumarola2019ganimation} is used to help focus the model on the relevant patterns. To improve synchronization further, the authors proposed a regression based discriminator which considers both sequence and content information. \looseness=-1
	
	\vspace{.5em}\noindent\textbf{EDs for Preventing Identity Leakage.}
	The authors in \cite{zhou2019talking} mitigate identity leakage by disentangling the speech and identity latent spaces using adversarial classifiers. Since their speech encoder is trained to project audio and video into the same latent space, the authors show how $x_g$ can be driven using $x_s$ or $a_s$.
	
	In \cite{jamaludin2019you}, the authors propose Speech2Vid which also uses separate encoders for audio and identity. \blue{However, to capture the identity better, the identity encoder $En_I$ uses a concatenation of five images of the target, and there are skip connections from the $En_I$ to the decoder.} To blend the mouth in better, a third `context' encoder is used to encourage in-painting. Finally, a VDSR CNN is applied to $x_g$ to sharpen the image.
	
	\blue{A disadvantage with \cite{zhou2019talking} and \cite{jamaludin2019you} is that they cannot control facial expressions and blinking. To resolve this,} the authors in \cite{vougioukas2019end} generate frames with a stride transposed CNN decoder on GRU-generated noise, in addition to the audio and identity encodings. Their video discriminator uses two RNNs for both the audio and video. When applying the L1 loss, the authors focus on the lower half of the face to encourage better lip sync quality over facial expressions.
	
	Later in \cite{vougioukas2019realistic}, the same authors \blue{improve the temporal coherence by splitting the video discriminator into two:} (1) for temporal realism in mouth to audio synchronization, and (2) for temporal realism in overall facial expressions. 	
	Then in \cite{kefalas2019speech}, the authors \blue{tune their approach further by fusing the encodings (audio, identity, and noise) with a polynomial fusion layer as opposed to simply concatenating the encodings together. Doing so makes the network less sensitive to large facial motions compared to \cite{vougioukas2019realistic} and \cite{jamaludin2019you}.}	
		\looseness=-1

	


	\vspace{-.2em}
	\subsection{Pose Reenactment}
	Most deep learning works in this domain focus on the problem of face frontalization. 
	However, there are some works which focus on facial pose reenactment.
	
	In \cite{hu2018pose} the authors use a U-Net to convert $(x_t,l_t,l_s)$ into $x_g$ using a GAN with two discriminators: one conditioned with the neutral pose image, and the other conditioned with the landmarks.
	In \cite{tran2018representation}, the authors propose DR-GAN for pose-invariant face recognition. To adjust the pose of $x_t$, the authors use an ED GAN which encodes $x_t$ as $e_t$, and then decodes $(e_t,p_s,z)$ as $x_g$, where $p_s$ is the source's pose vector and $z$ is a noise vector. \blue{Compared to \cite{hu2018pose}, \cite{tran2018representation} has the flexibility of manipulating the encodings for different tasks and the authors improve the quality of $x_g$ by averaging multiple examples of the identity encoding before passing it through the decoder (similar to \cite{gu2020flnet,zakharov2019few,wang2019fewshotvid2vid}).}	
	In \cite{cao20193d}, the authors suggest using two GANs: The first frontalizes the face and produces a UV map, and second rotates the face, given the target angle as an injected embedding. \blue{The result is that each model performs a less complex operation and can therefore the models collectively can produce a higher quality image.}

	\subsection{Gaze Reenactment}
	There are only a few deep learning works which have focused on gaze reenactment. In \cite{ganin2016deepwarp} the authors convert a cropped eye $x_t$, its landmarks, and the source angle, to a flow (vector) field using a 2-scale CNN. $x_g$ is then generated by applying a flow field to $x_t$ to warping it to the source angle. The authors then correct the illumination of $x_g$ with a second CNN. 
	\blue{A challenge with \cite{ganin2016deepwarp} is that the head must be frontal to avoid inconsistencies due to pose and perspective. To mitigate this issue,} the authors of \cite{yu2019improving} proposed the Gaze Redirection Network (GRN). In GRN, the target's cropped eye, head pose, and source angle are encoded separately and then passed though an ED network to generate an optical flow field. The field is used to warp $x_t$ into $x_g$. To overcome the lack of training data and the challenge of data pairing, the authors (1) pre-train their network on 3D synthesized examples, (2) further tune their network on real images, and then (3) fine tune their network on 3-10 examples of the target.

	\subsection{Body Reenactment}
\begin{figure*}	
	\begin{subfigure}[t]{.49\textwidth}	
	\centering
	\caption{\textbf{\cite{chan2019everybody} Everybody Dance Now:}}
	\includegraphics[width=\textwidth]{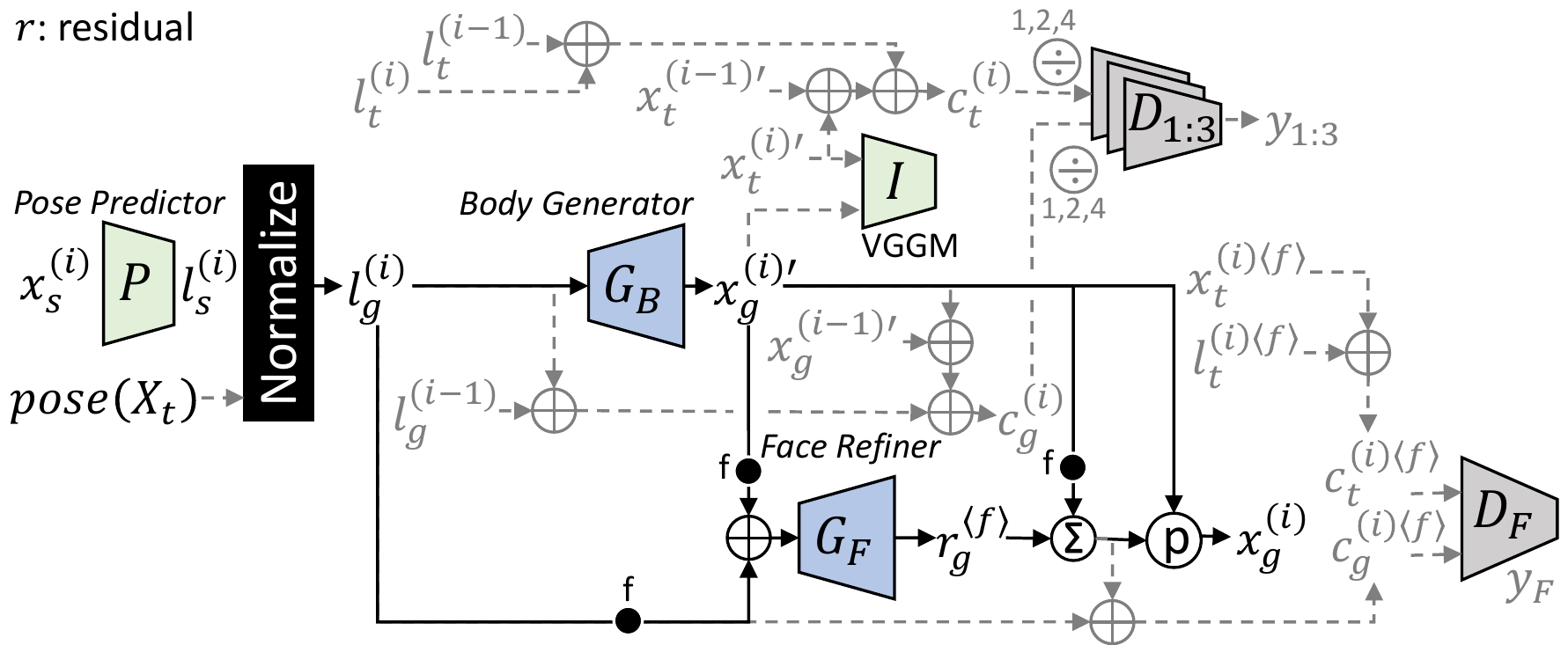}\vspace{0em}
		\includegraphics[width=\textwidth]{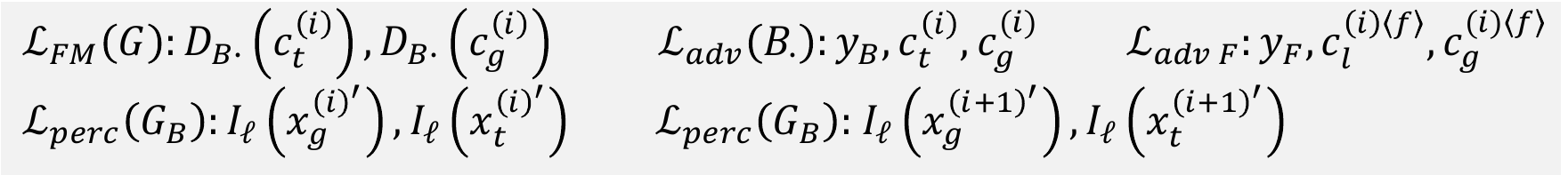}
\end{subfigure}
	\begin{subfigure}[t]{.49\textwidth}	
	\centering
	\caption{\textbf{\cite{liu2019neural} NRR-HAV:} }
	\includegraphics[width=\textwidth]{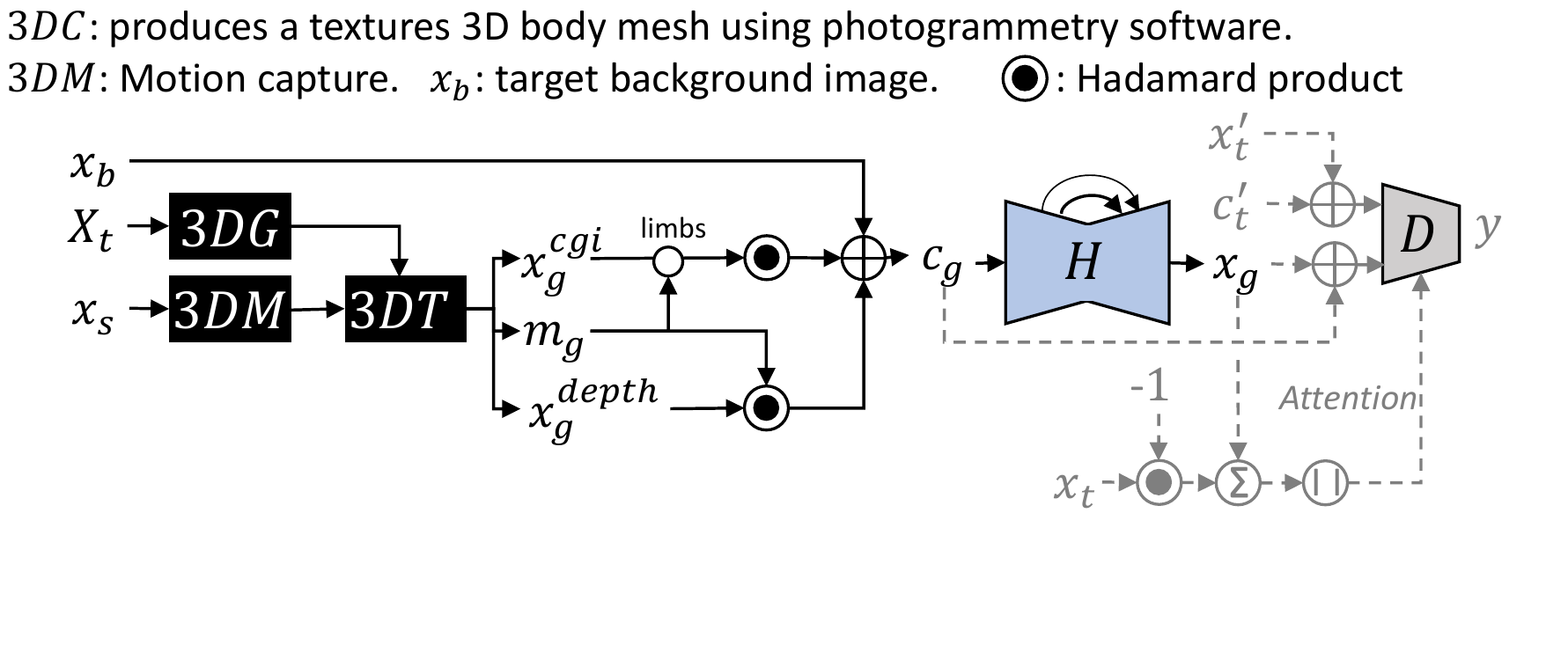}\vspace{0em}
		\includegraphics[width=\textwidth]{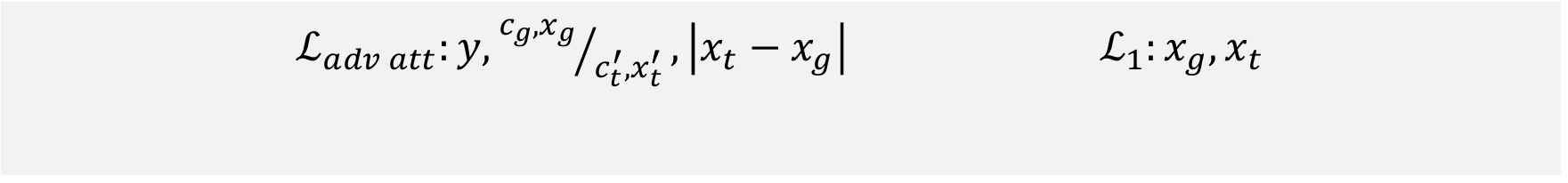}
\end{subfigure}
	\begin{subfigure}[t]{.49\textwidth}	
	\centering
	\caption{\textbf{\cite{aberman2019deep} Deep Vid. Perf. Cloning:}}
	\includegraphics[width=\textwidth]{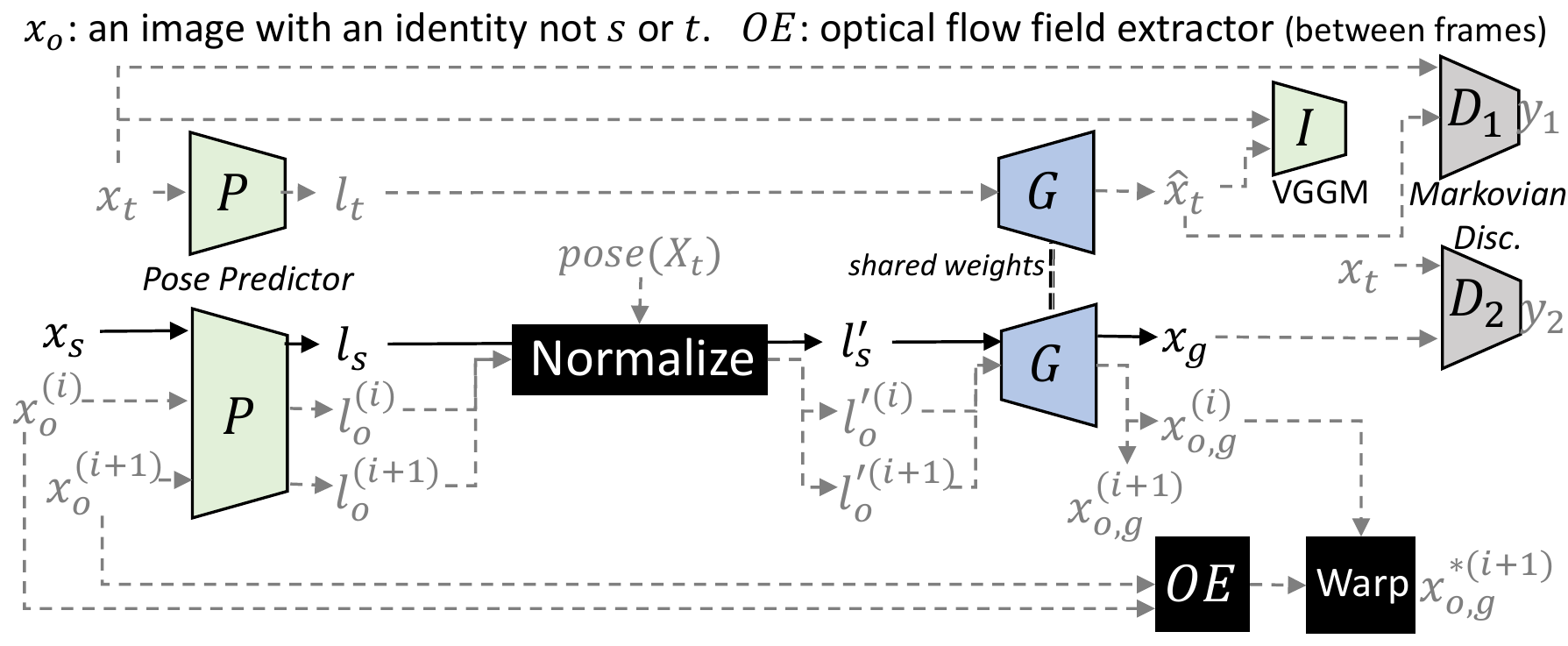}\vspace{0em}
		\includegraphics[width=\textwidth]{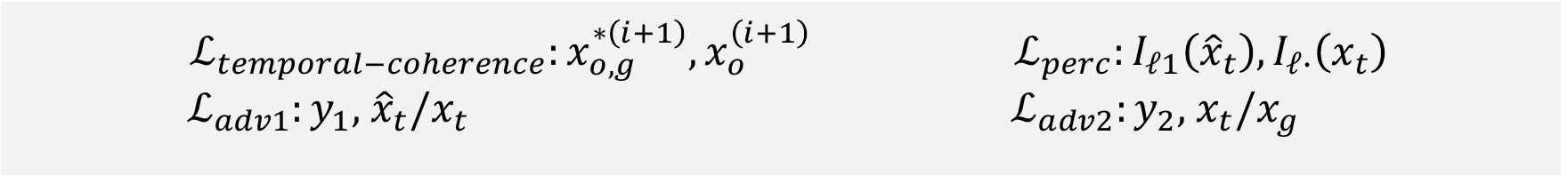}
\end{subfigure}
		\begin{subfigure}[t]{.49\textwidth}	
	\centering
	\caption{\textbf{\cite{zablotskaia2019dwnet} DwNet:} }
	\includegraphics[width=\textwidth]{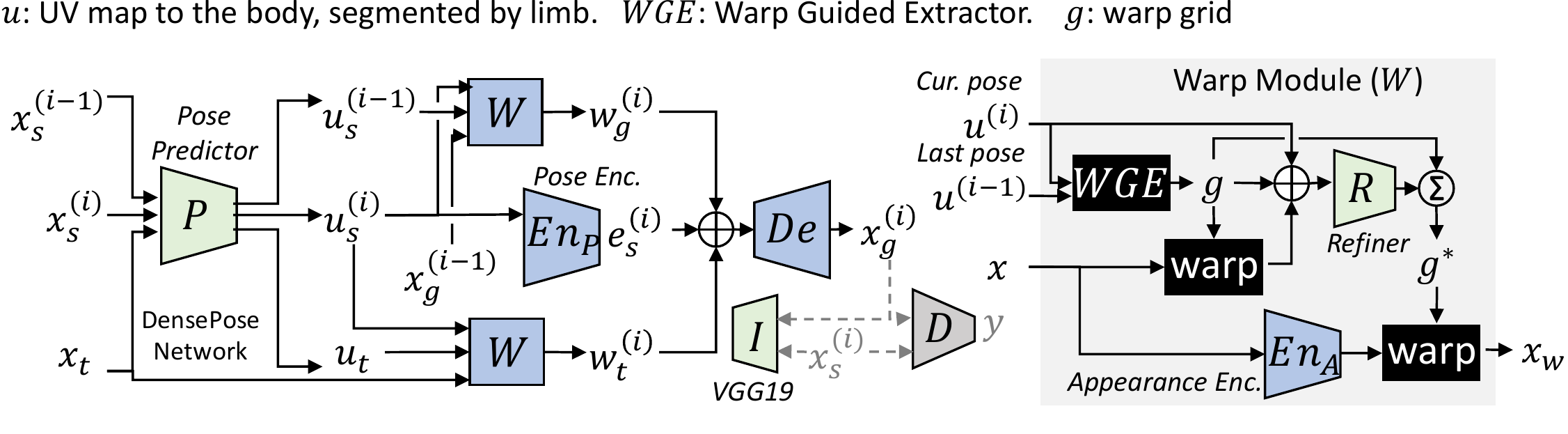}\vspace{2.5em}
		\includegraphics[width=\textwidth]{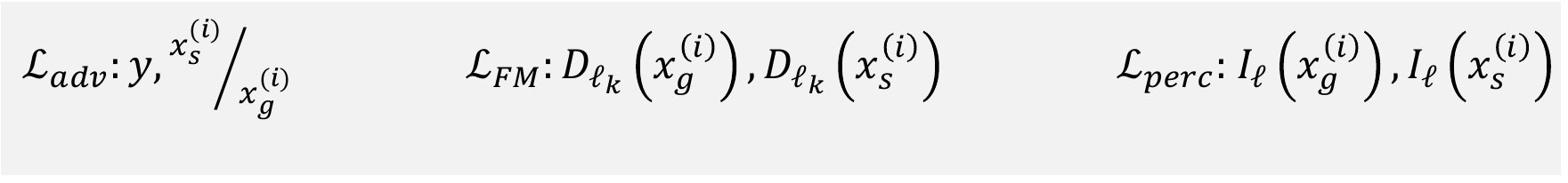}
\end{subfigure}

	\caption{Architectural schematics for some \textbf{body reenactment networks}. Black lines indicate prediction flows used during deployment, dashed gray lines indicate dataflows performed during training.}\label{fig:schem_reen_body}
	\vspace{-1em}
\end{figure*}

	Several facial reenactment papers from Section \ref{subsec:expression} discuss body reenactment too. For example, Vid2Vid \cite{wang2018vid2vid,wang2019fewshotvid2vid}, MocoGAN \cite{tulyakov2018mocogan}, and others \cite{siarohin2019animating,NIPS2019_8935}. In this section, we focus on methods which specifically target body reenactment. Schematics for some of these architectures can be found in Fig. \ref{fig:schem_reen_body}.

	\subsubsection{\textbf{One-to-One} (Identity to Identity)}
	In the work \cite{liu2019video}, the authors perform facial reenactment with the upper-body as well (arms and hands). The approach is to (1) use a pix2pixHD GAN to convert the source's facial boundaries to the targets, (2) and then paste them onto a captured pose skeleton of the source, and (3) use a pix2pixHD GAN to generate $x_g$ from the composite.

	\subsubsection{\textbf{Many-to-One} (Multiple Identities to a Single Identity)} \hfill \\
\textbf{Dance Reenactment.}
	In \cite{chan2019everybody} the authors make people dance using a target specific pix2pixHD GAN with a custom loss function. The generator receives an image of the captured pose skeleton and the discriminator receives the current and last image conditioned on their poses. The quality of face is then improved with a residual predicted by an additional pix2pixHD GAN, given the face region of the pose. A many-to-one relationship is achieved by normalizing the input pose to that of the target's.
	
	\blue{The authors of \cite{liu2019neural} then tried to overcome artifacts which occur in \cite{chan2019everybody} such stretched limbs due to incorrectly detected pose skeletons.} They used photogrammetry software on hundreds of images of the target, and then reenacted the 3D rendering of the target's body. The rendering, partitioned depth map, and background are then passed to a pix2pix model for image generation, using an attention loss.\looseness=-1
	
	\blue{Another artifact in \cite{chan2019everybody} was that the model could not generalize well to unseen poses. To improve the generalization, the authors of \cite{aberman2019deep} trained their network on many identities other than $s$ and $t$.} First they trained the GAN on paired data (the same identity doing different poses) and then later added another discriminator to evaluate the temporal coherence given (1) $x_{g}^{(i)}$ driven by another video, and (2) the optical flow predicted version.
	
	\blue{A challenge with the previous works was that they required a lots of training data. This was reduced from about an hour of video footage to only 3 minutes in \cite{zhou2019dance} by segmenting and orienting the limbs of $x_t$ according to $x_s$ before the generation step.} Then a pix2pixHD GAN uses this composition and the last $k$ frames' poses to generate the body. Finally, another pix2pixHD GAN is used to blend the body into the background.

	\subsubsection{\textbf{Many-to-Many} (Multiple IDs to Multiple IDs)} \hfill \\
\textbf{Pose Alignment.} 
	In \cite{siarohin2018deformable} the authors try to resolve the issue of misalignment when using pix2pix like architectures. They propose `deformable skip connections' which help orient the shuttled feature maps according to the source pose. The authors also propose a novel nearest neighbor loss instead of using L1 or L2 losses. To modify unseen identities at test time, an encoding of $x_t$ is passed to the decoder's inner layers.
	
	\blue{Although the work of \cite{siarohin2018deformable} helps align the general images, artifacts can still occur when $x_s$ and $x_t$ have very different poses. To resolve this, the authors of \cite{zhu2019progressive} use novel Pose-Attentional Transfer blocks (PATB) inside their GAN-based generator.} The architecture passes $x_t$ and the poses $p_s$ concatenated with $p_t$ through separate encoders which are passed though a series of PATBs before being decoded. The PATBs progressively transfer regional information of the poses to regions of the image to ultimately create a body that has better shape and appearance consistency.
	\looseness=-1
	
	\vspace{.5em}\noindent\textbf{Pose Warping.}
	In \cite{neverova2018dense} the authors use a pre-trained DensePose network \cite{alp2018densepose} to refine a predicted pose with a warped and in-painted DensePose UV spatial map of the target. \blue{Since the spatial map covers all surfaces of the body, the generated image has improved texture consistency.}
	\blue{In contrast to \cite{zhu2019progressive,siarohin2018deformable} which uses feature mappings to alleviate misalignment, the authors of \cite{zablotskaia2019dwnet} use warping which reduces the complexity of the network's task.} Their model, called DwNet, uses a `warp module' in an ED network to encode $x_{t}^{(i-1)}$ warped to $p_{s}^{(i)}$, where $p$ is a UV body map of a pose obtained a DensePose network. 
	
	\blue{A challenge with the alignment techniques of the previous works is that the body's 3D shape and limb scales are not considered by the network resulting in identity leakage from $x_s$. In \cite{liu2019liquid}, the authors counter this issue with their Liquid Warping GAN. This is accomplished by} predicting target and source's 3D bodies with the model in \cite{kanazawa2018end} and then by translating the two through a novel liquid warping block (LWB) in their generator. Specifically, the estimated UV maps of $x_s$ and $x_t$, along with their calculated transformation flow, are passed through a three stream generator which produces (1) the background via in-painting, (2) a reconstruction of the $x_s$ and its mask for feature mapping, and (3) the reenacted foreground and its mask. The latter two streams use a shared LWB to help the networks address multiple sources (appearance, pose, and identity). The final image is obtained through masked multiplication and the system is trained end-to-end. 
	
	\vspace{.5em}\noindent\textbf{Background Foreground Compositing.}
	In \cite{balakrishnan2018synthesizing}, the authors break the process down into three stages, trained end-to-end: (1) use a U-Net to segment $x_t$'s body parts and then orient them according to the source pose $p_s$, (2) use a second U-Net to generate the body $x_g$ from the composite, and (3) use a third U-Net to perform in-painting on the background and paste $x_g$ into it. 
	\blue{The authors of \cite{ma2018disentangled} then streamlined this process by using a single ED GAN network to disentangle the foreground appearance (body), background appearance, and pose. Furthermore, by using an ED network, the user gains control over each of these aspects.} This is accomplished by segmenting each of these aspects before passing them through encoders.
	\blue{To improve the control over the compositing, the authors of \cite{de2019conditional} used a CVAE-GAN. This enabled the authors to change the pose and appearance of bodies individually.} The approach was to condition the network on heatmaps of the predicted pose and skeleton.

	\subsubsection{\textbf{Few-Shot Learning}}
	In \cite{lee2019metapix}, the authors demonstrate the few-shot learning technique of \cite{finn2017model} on a pix2pixHD network and the network of \cite{balakrishnan2018synthesizing}. Using just a few sample images, they were able to transfer the resemblance of a target to new videos in the wild.

	\section{Replacement}\label{sec:repl}
\begin{figure*}[!htbp]	
	\centering
\begin{subfigure}[t]{.49\textwidth}	
	\centering
	\caption{\textbf{\cite{shaoanlu58:online} Face Swap GAN:} }
	\includegraphics[width=\textwidth]{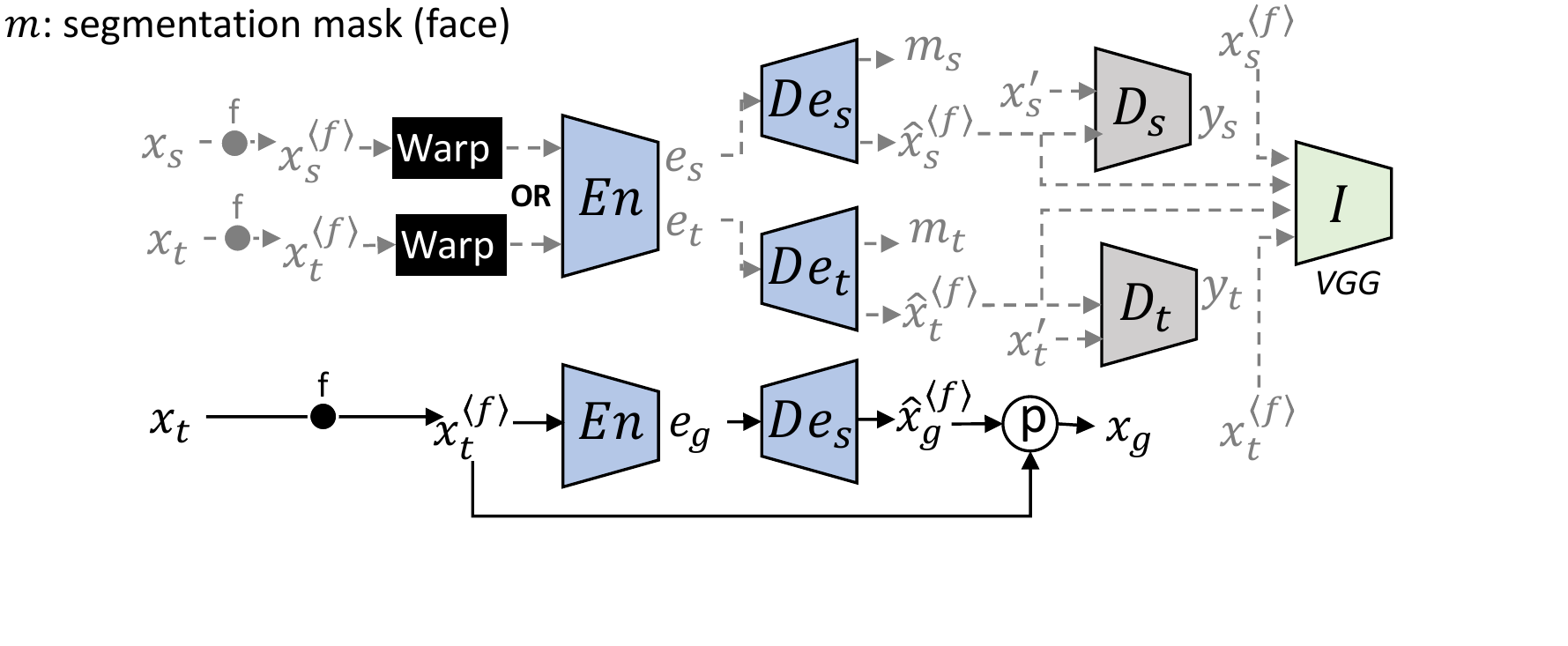}\vspace{-1.5em}
		\includegraphics[width=\textwidth]{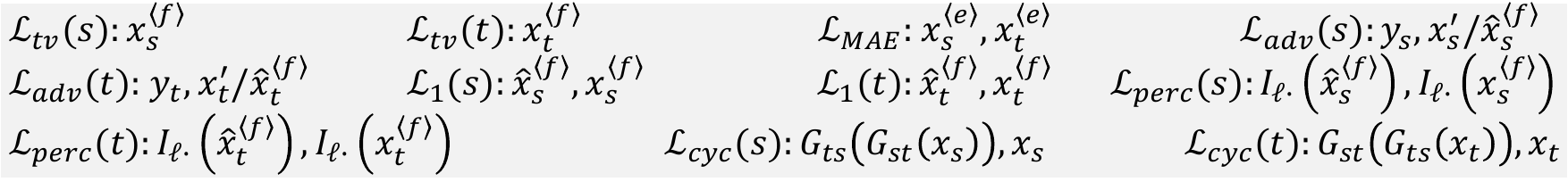}
\end{subfigure}
\begin{subfigure}[t]{.49\textwidth}	
	\centering
	\caption{\textbf{\cite{korshunova2017fast} Fast Face Swap:} }
	\includegraphics[width=\textwidth]{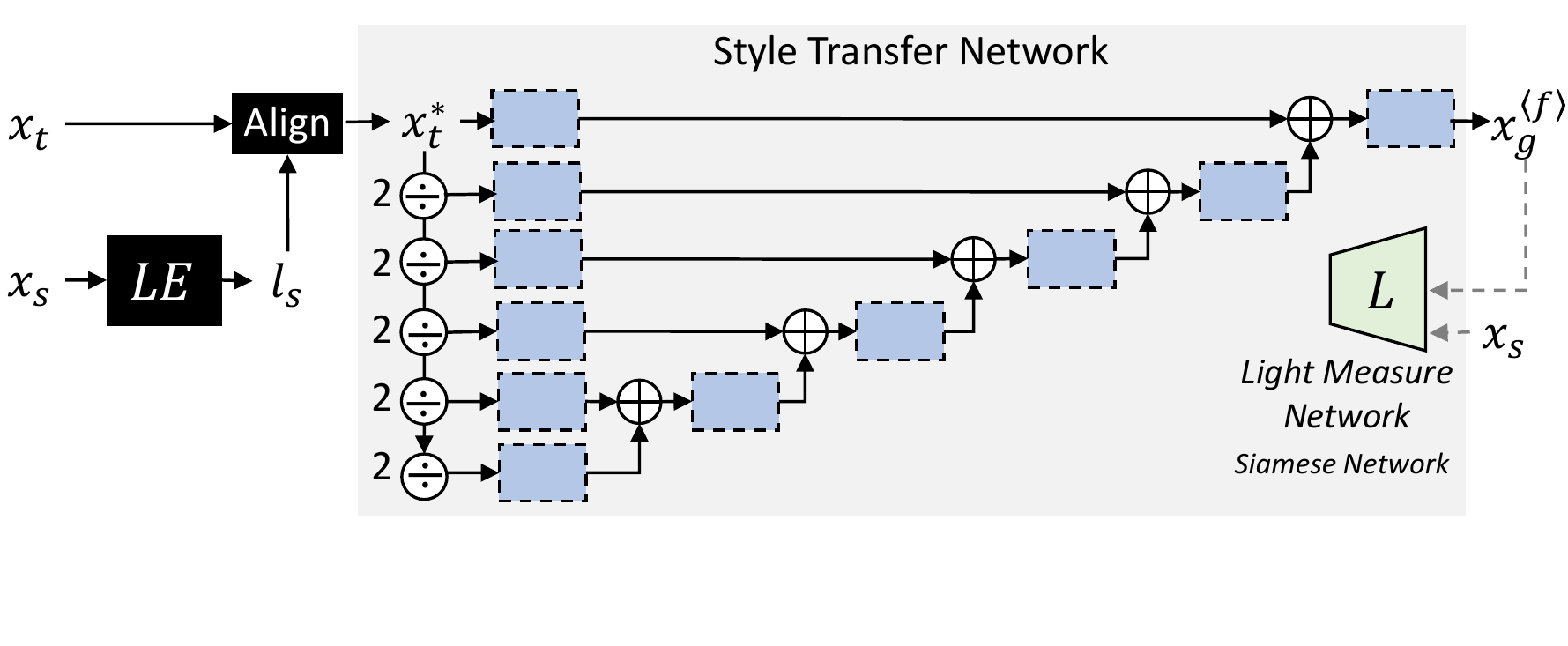}\vspace{-1.5em}
		\includegraphics[width=\textwidth]{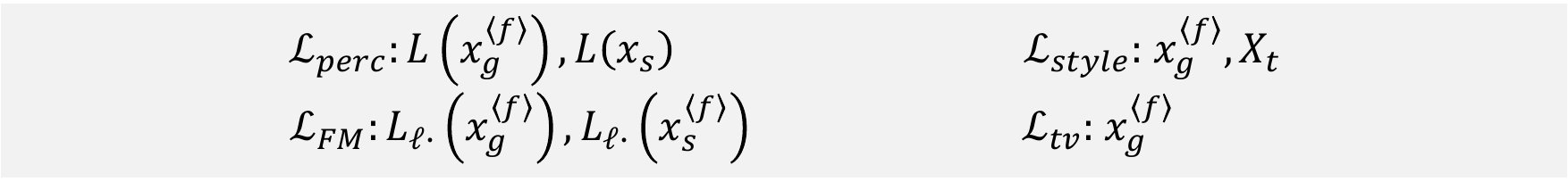}
\end{subfigure}
\begin{subfigure}[t]{.49\textwidth}	
	\centering
	\caption{\textbf{\cite{bao2018towards} OSIP-FS:} }
	\includegraphics[width=\textwidth]{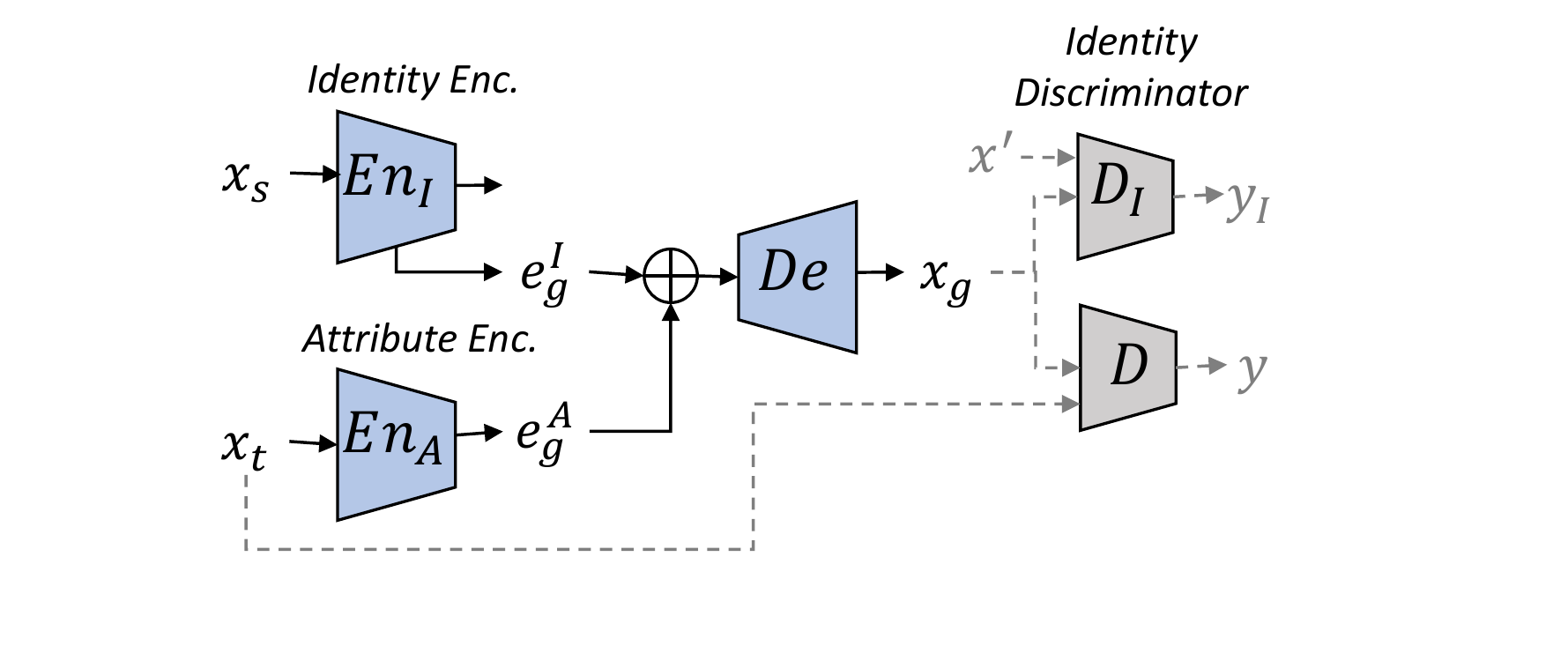}\vspace{-.1em}
		\includegraphics[width=\textwidth]{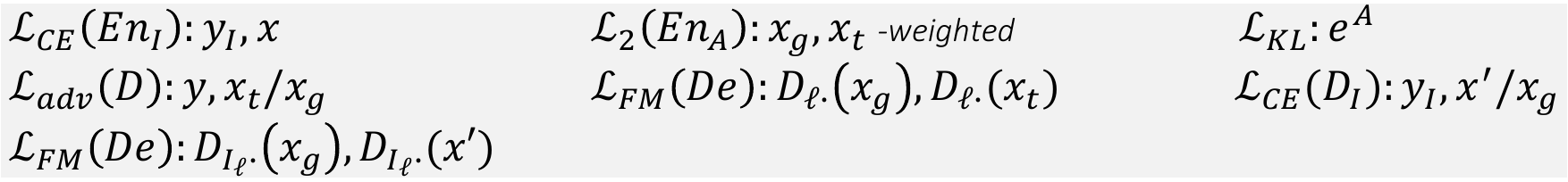}
\end{subfigure}
\begin{subfigure}[t]{.49\textwidth}	
	\centering
	\caption{\textbf{\cite{natsume2018rsgan} RSGAN:} }
	\includegraphics[width=\textwidth]{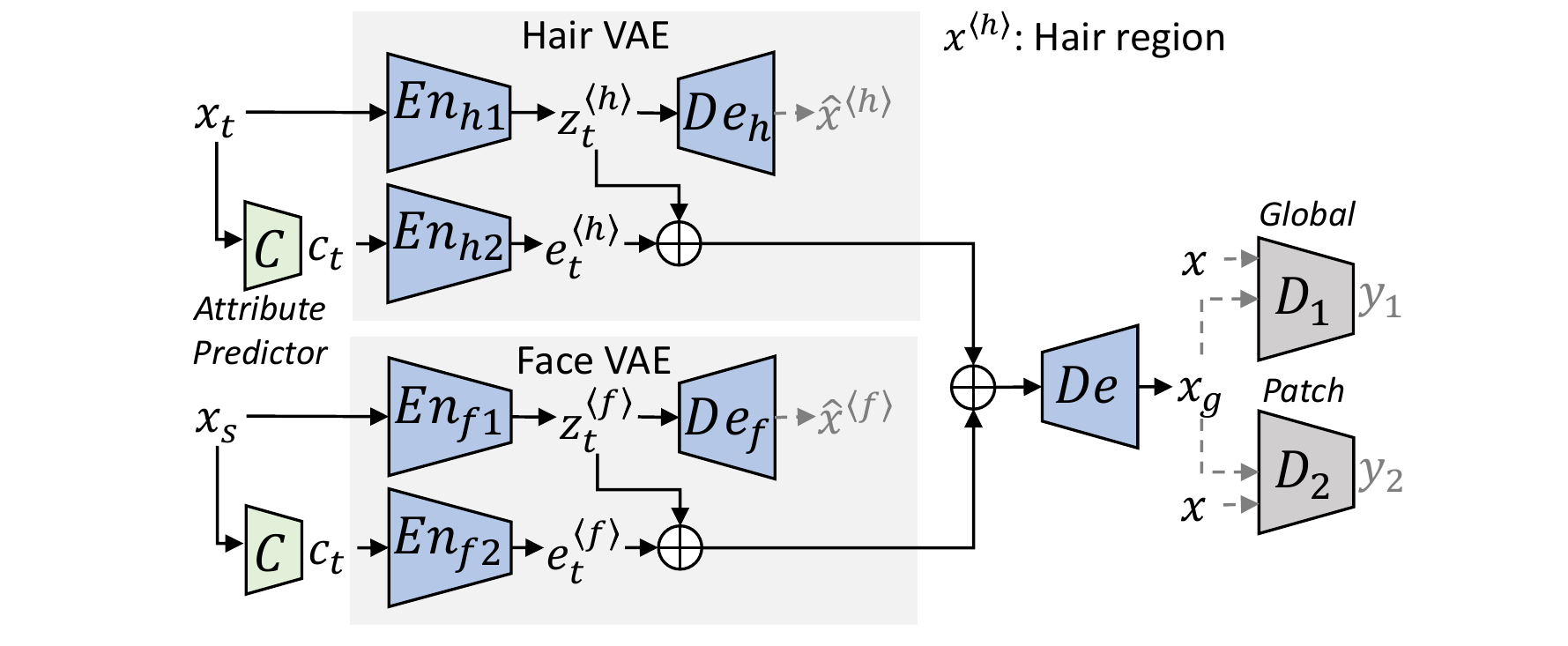}\vspace{-.1em}
		\includegraphics[width=\textwidth]{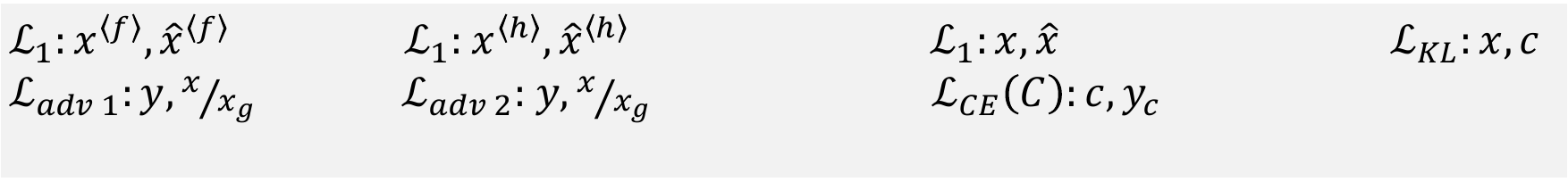}
\end{subfigure}
\begin{subfigure}[t]{.49\textwidth}	
	\centering
	\caption{\textbf{\cite{natsume2018fsnet} FSNet:}}
	\includegraphics[width=\textwidth]{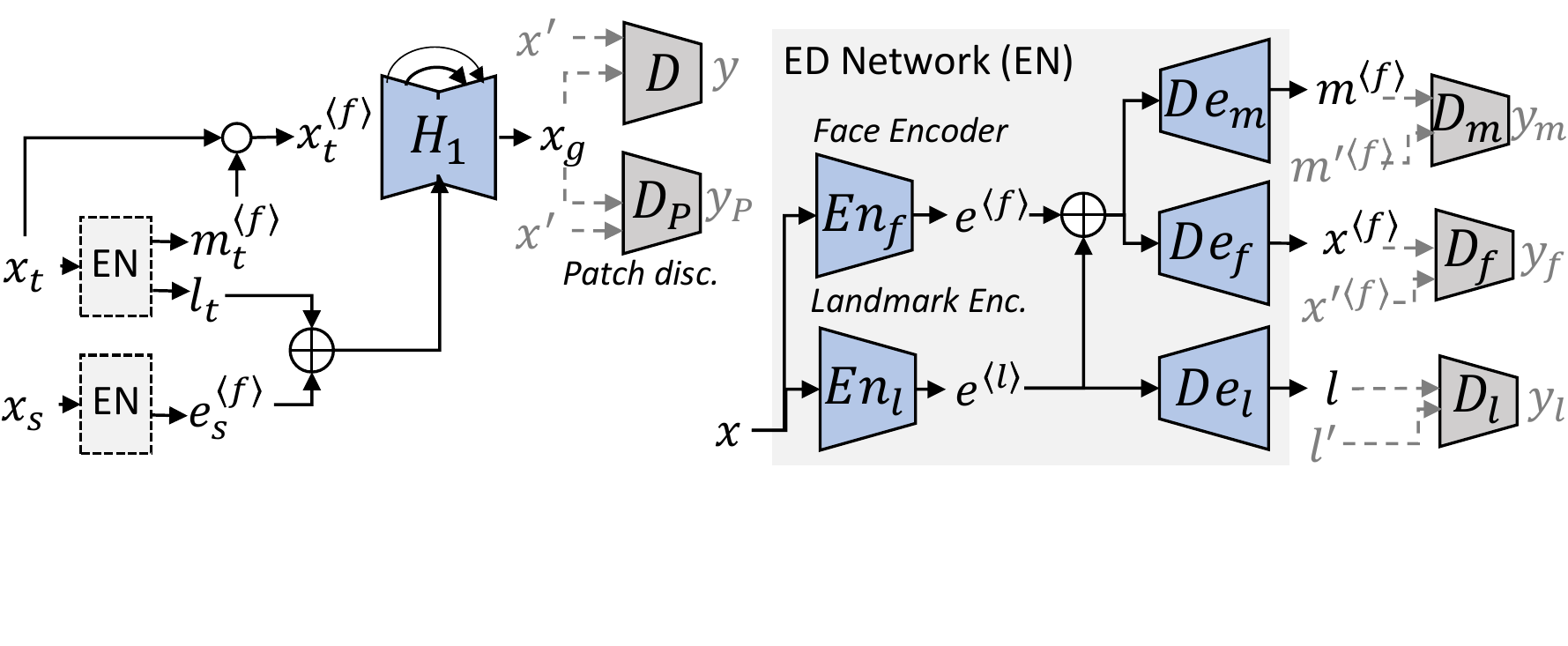}\vspace{-2em}
		\includegraphics[width=\textwidth]{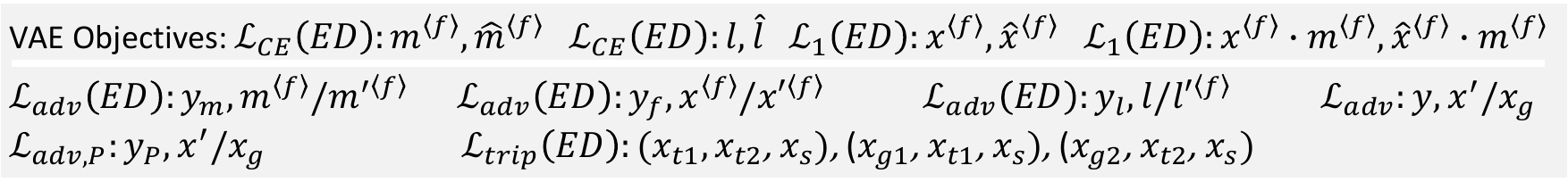}
\end{subfigure}
\begin{subfigure}[t]{.49\textwidth}	
	\centering
	\caption{\textbf{\cite{li2019faceshifter} FaceShifter:} }
	\includegraphics[width=\textwidth]{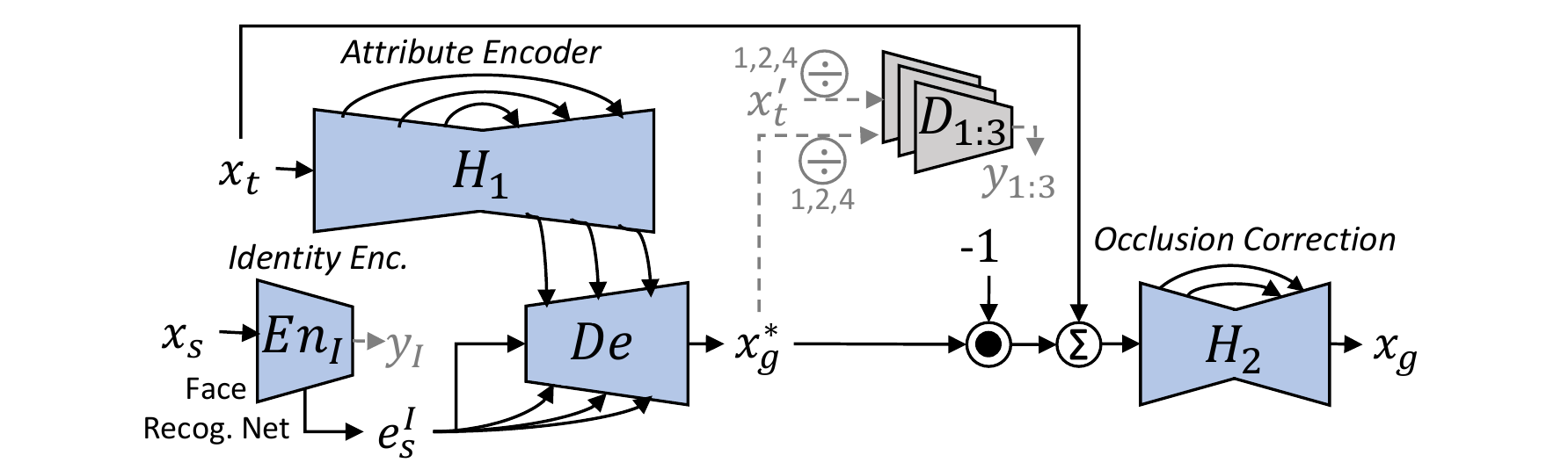}\vspace{0em}
		\includegraphics[width=\textwidth]{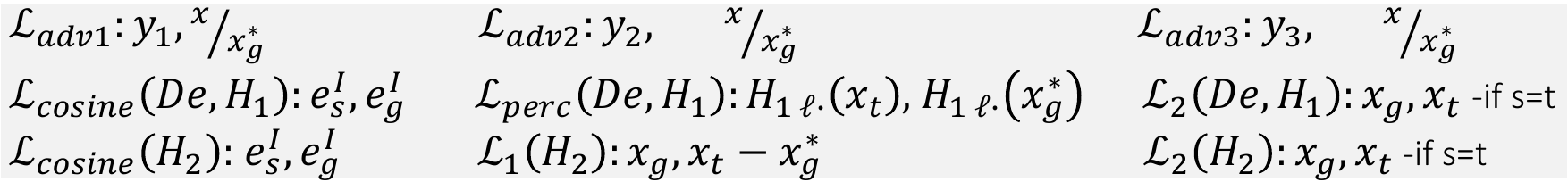}
\end{subfigure}
\begin{subfigure}[t]{.49\textwidth}	
	\centering
	\caption{\textbf{\cite{shaoanlu29:online} Few-Shot Face Translation:} }
	\includegraphics[width=\textwidth]{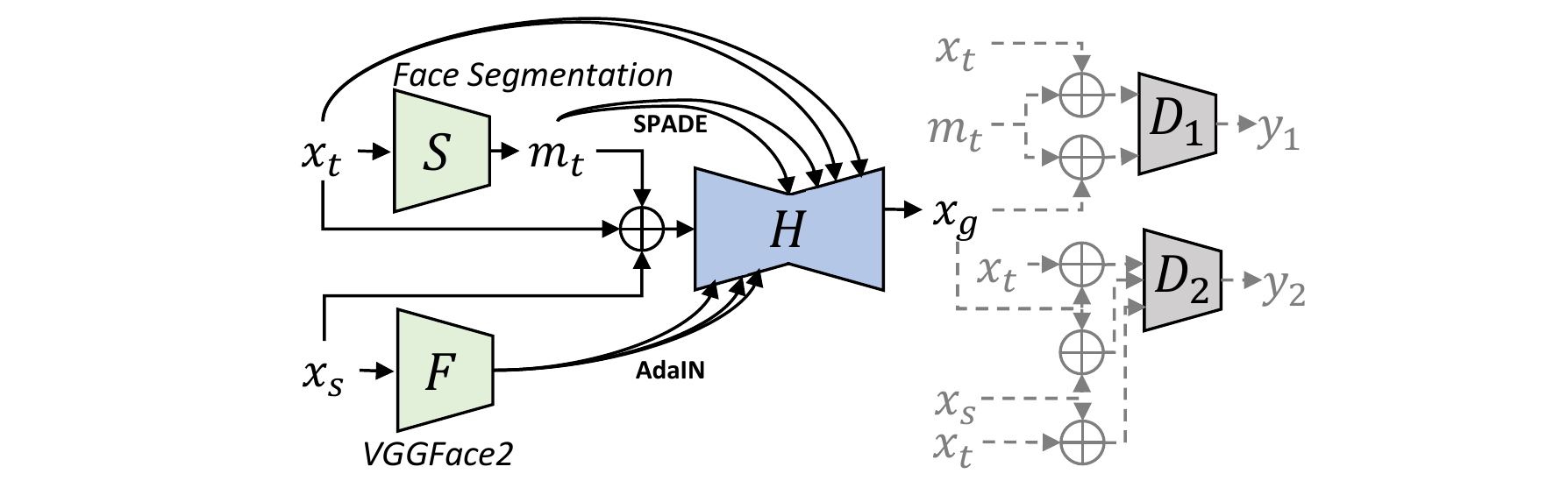}\vspace{1.2em}
		\includegraphics[width=\textwidth]{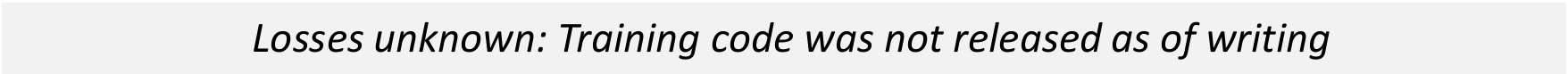}
\end{subfigure}
\begin{subfigure}[t]{.49\textwidth}	
	\centering
	\caption{\textbf{\cite{moniz2018unsupervised} Depth Nets:} }
	\includegraphics[width=\textwidth]{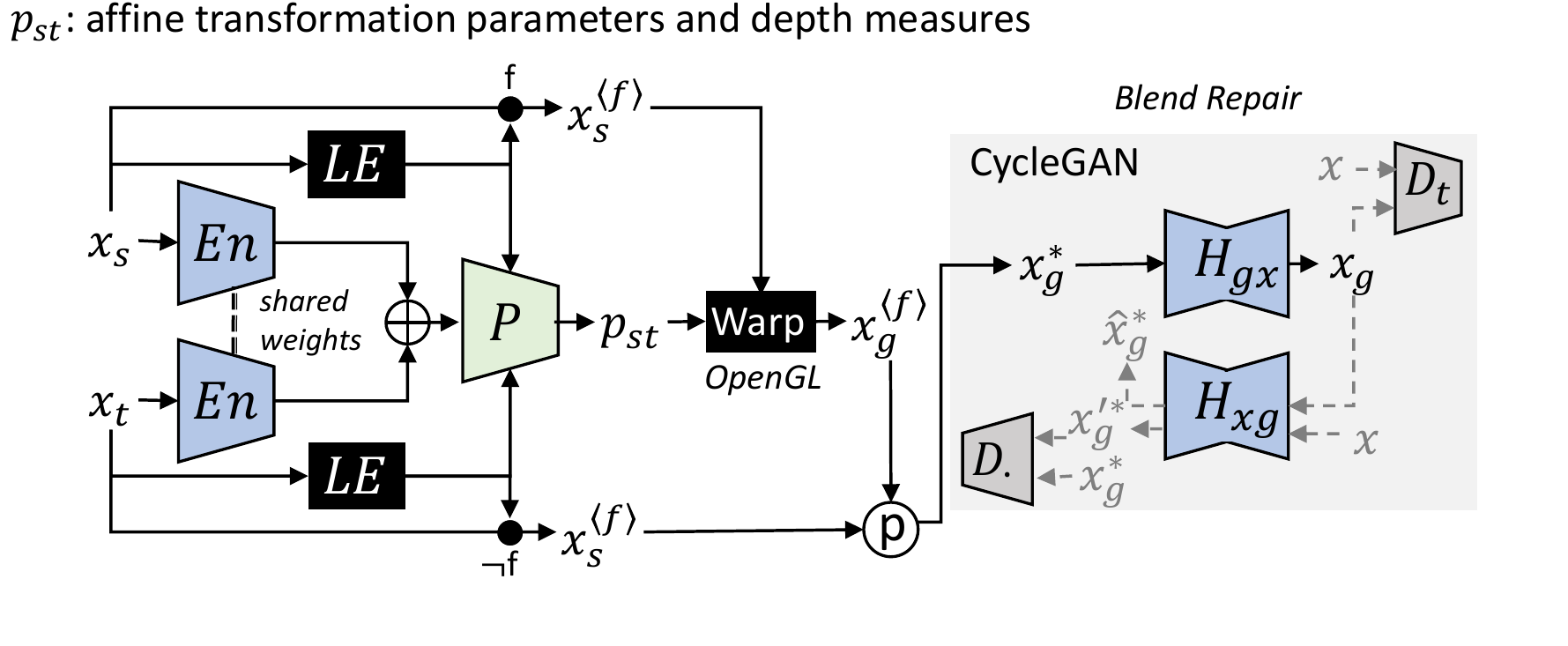}\vspace{-.5em}
		\includegraphics[width=\textwidth]{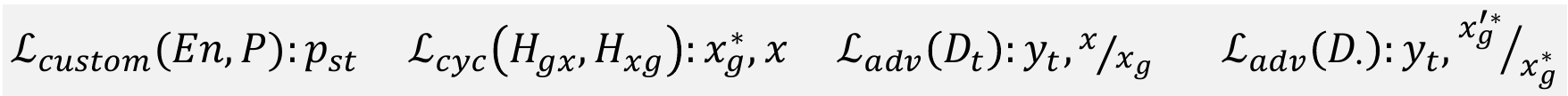}
\end{subfigure}
\begin{subfigure}[t]{.49\textwidth}	
	\centering
	\vspace{-0em}
	\caption{\textbf{\cite{xiao2019identity} IHPT:} }
	\includegraphics[width=\textwidth]{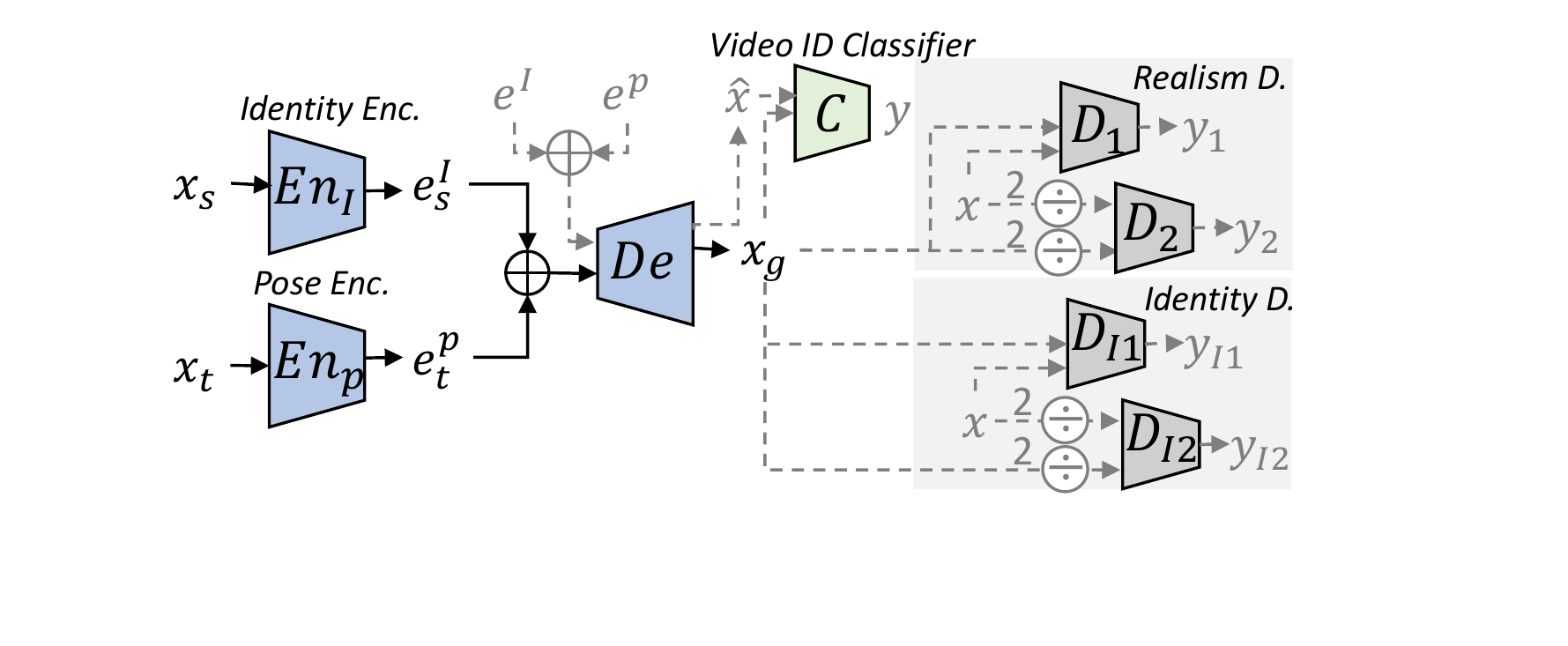}\vspace{-.5em}
		\includegraphics[width=\textwidth]{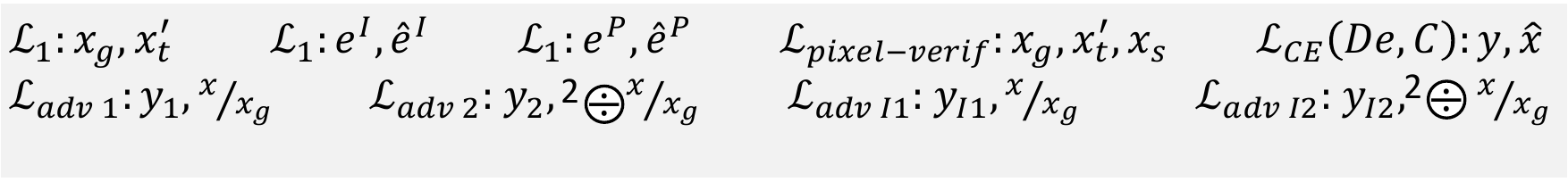}
\end{subfigure}
\begin{subfigure}[t]{.49\textwidth}	
	\centering
	\vspace{-.3em}
	\caption{\textbf{\cite{nirkin2019fsgan} FSGAN:} }
	\vspace{-0em}\includegraphics[width=\textwidth]{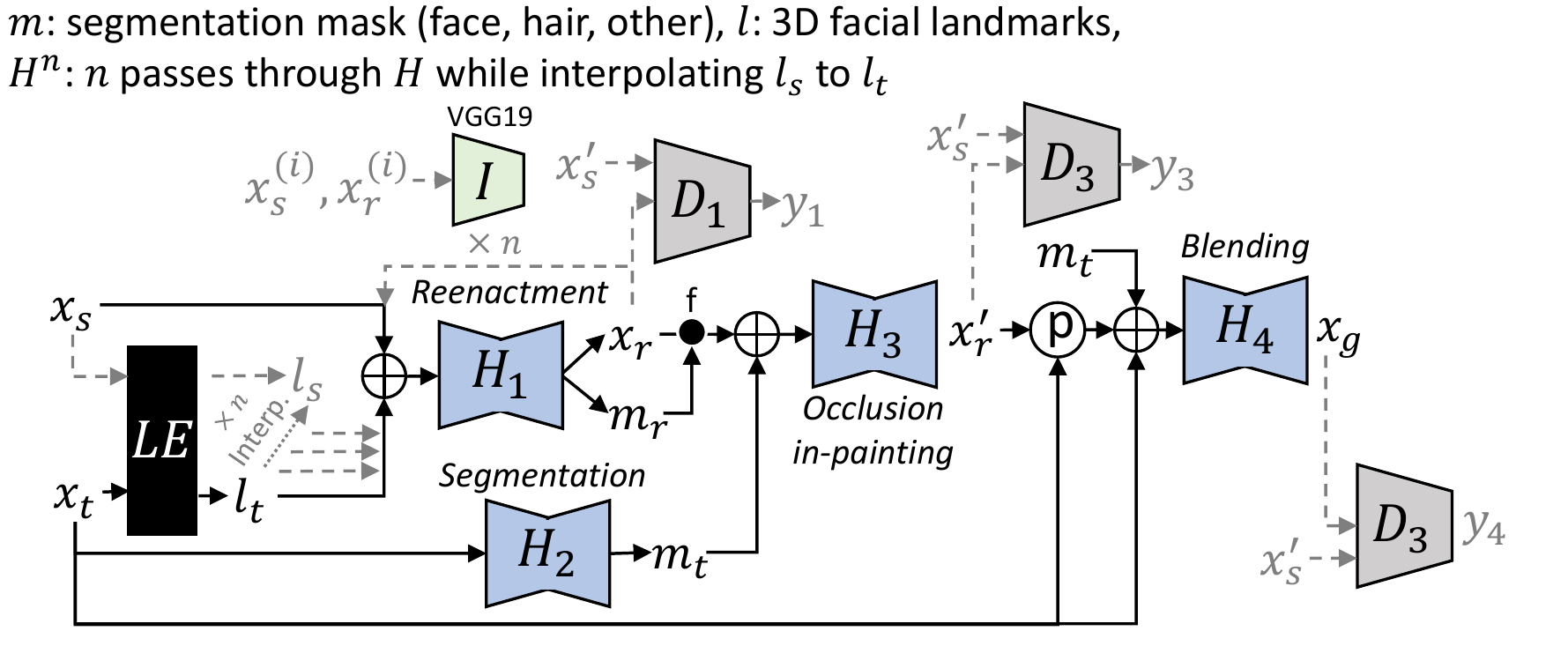}\vspace{0em}
		\includegraphics[width=\textwidth]{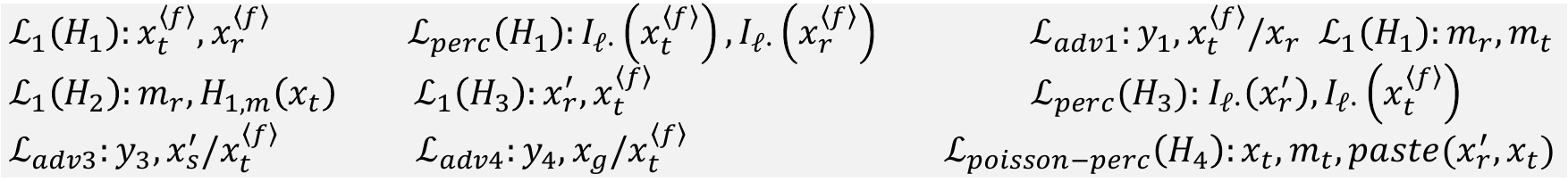}
\end{subfigure}
\begin{subfigure}[t]{.49\textwidth}	
	\centering
	\hfill
\end{subfigure}
\vspace{-.5em}
	\caption{Architectural schematics of the \textbf{replacement networks} with their generation and training dataflows.}\label{fig:schem_rep}
\end{figure*}

		The network schematics and summary of works for replacement deepfakes can be found in Fig. \ref{fig:schem_rep} and Table \ref{tab:replacement} respectively.
	
	\subsection{Swap}\label{subsec:swap}
	
	At first, face swapping was a manual process accomplished using tools such as Photoshop. More automated systems first appeared between 2004-08 in \cite{blanz2004exchanging} and \cite{bitouk2008face}. Later, fully automated methods were proposed in \cite{vlasic2006face,dale2011video,kemelmacher2016transfiguring} and \cite{nirkin2018face} using methods such as warping and reconstructed 3D morphable face models.

	\subsubsection{\textbf{One-to-One} (Identity to Identity)} 
\begin{figure}[t] 
	\centering
		\includegraphics[width=.8\textwidth]{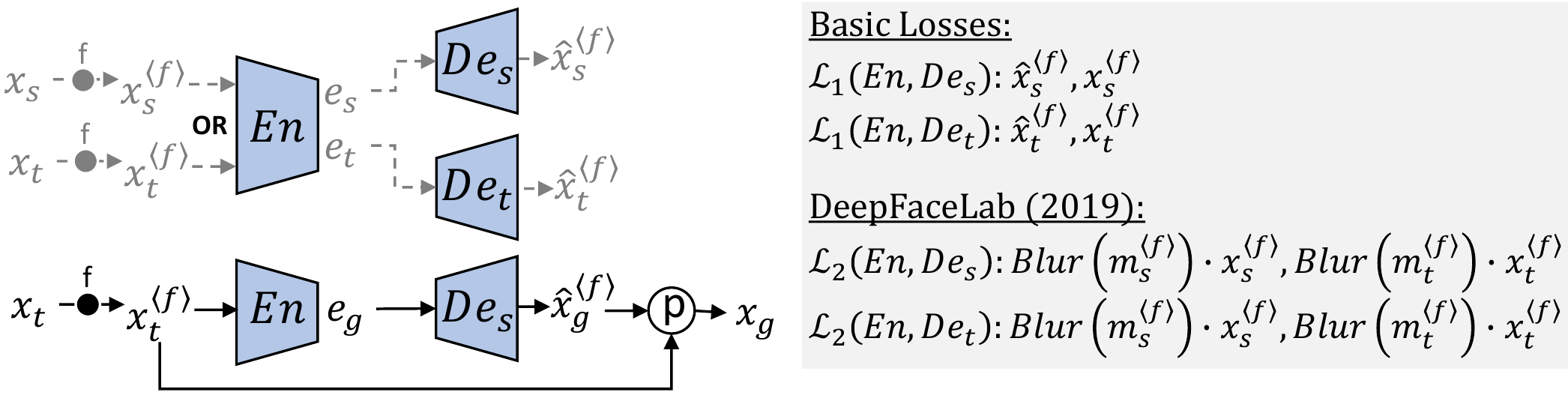}
		\vspace{-.8em}
\caption{The basic schematic for the Reddit \textit{`deepfakes'} model and its variants\cite{iperovDe6:online,deepfake32:online,shaoanlu58:online}.}\label{fig:basic_faceswap}
\vspace{-1em}
\end{figure}

\hfill \\ 
	\textbf{Online Communities.}
	After the Reddit user `deepfakes' was exposed in the media, researchers and online communities began finding improved ways to perform face swapping with deep neural networks. 
	The original deepfake network, published by the Reddit user, is an ED network (visualized in Fig. \ref{fig:basic_faceswap}). The architecture consists of one encoder $En$ and two decoders $De_s$ and $De_t$. The components are trained concurrently as two autoencoders: $De_s(En(x_s))=\hat{x}_s$ and $De_t(En(x_t))=\hat{x}_t$, where $x$ is a cropped face image. As a result, $En$ learns to map $s$ and $t$ to a shared latent space, such that \looseness=-1
	\begin{equation}
		De_s(En(x_t))=x_g
	\end{equation} 
	Currently, there are a number of open source face swapping tools on GitHub based on the original network. One of the most popular is DeepFaceLab \cite{iperovDe6:online}. Their current version offers a wide variety of model configurations, including adversarial training, residual blocks, a style transfer loss, and masked loss to improve the quality of the face and eyes. 
	To help the network map the target's identity into arbitrary face shapes, the training set is augmented with random face warps. 
	
	Another tool called FaceSwap-GAN \cite{shaoanlu58:online} follows a similar architecture, but uses a denoising autoencoder with a self-attention mechanisms, and offers cycle-consistency loss \blue{which can reduce the identity leakage and increase the image fidelity.} The decoders in FaceSwap-GAN also generate segmentation masks \blue{which helps the model handle occlusions and is used to blend $x_g$ back into the target frame.} 
	Finally, \cite{deepfake32:online} is another open source tool that provides a GUI. Their software comes with 10 popular implementations, including that of \cite{iperovDe6:online}, and multiple variations of the original Redit user's code. 

	\subsubsection{\textbf{One-to-Many} (Single Identity to Multiple Identities)}\hfill \\ 
	In \cite{korshunova2017fast}, the authors use a modified style transfer with CNN, 
	where the content is $x_t$ and the style is the identity of $x_s$. The process is (1) align $x_t$ to a reference $x_s$, (2) transfer the identity of $s$ to the image using a multi scale CNN, trained with style loss on images of $s$, and (3) align the output to $x_t$ and blend the face back in with a segmentation mask. 

	\subsubsection{\textbf{Many-to-Many} (Multiple IDs to Multiple IDs)}\hfill \\ 
	One of the first identity agnostic methods was \cite{olszewski2017realistic}, mentioned in Section \ref{subsubsec:expression:many2many}. \blue{However, to train this CGAN, one needs a dataset of paired faces with different identities having the same expression.}

	\vspace{.5em}\noindent\textbf{Disentanglement with EDs.}
	However, To provide more control over the In \cite{bao2018towards} the authors us an ED to disentangle the identity from the attributes (pose, hair, background, and lighting) during the training process. The identity encodings are the last pooling layer of a face classifier, and the attribute encoder is trained using a weighted L2 loss
	and a KL divergence loss to mitigate identity leakage. The authors also show that they can adjust attributes, expression, and pose via interpolation of the encodings. 	
	\blue{Instead of swapping identities, the authors of \cite{sun2018hybrid} wanted to \textit{variably} obfuscate the target's identity.} To accomplish this, the authors used an ED to predict the 3D head parameters which where either modified or replaced with the source's. Finally a GAN was used to in-paint the face of $x_t$ given the modified head model parameters.

	\looseness=-1
	
	\vspace{.1em}\noindent\textbf{Disentanglement with VAEs.}	
	In \cite{natsume2018rsgan}, the authors propose RSGAN: a VAE-GAN consisting of two VAEs and a decoder. One VAE encodes the hair region and the other encodes the face region, where both are conditioned on a predicted attribute vector $c$ describing $x$. Since VAEs are used, the facial attributes can be edited through $c$.
	
	\blue{In contrast to \cite{natsume2018rsgan}, the authors of \cite{natsume2018fsnet} use a VAE to prepare the content for the generator, and use a network to perform the blending via in-painting.} A single VAE-ED network is run on $x_s$ and then $x_t$ producing encodings for the face of $x_s$ and the landmarks of $x_t$. To perform a face swap, a generator receives the masked portrait of $x_t$ and performs in-painting on the masked face. The generator uses the landmark encodings in its embedding layer. During training, randomly generated faces are used with triplet loss on the encodings to preserve identities.

	\vspace{.5em}\noindent\textbf{Face Occlusions.} 
	FSGAN \cite{nirkin2019fsgan}, mentioned Section \ref{subsubsec:expression:many2many}, is also capable of face swapping and can handle occlusions. After the face reenactment generator produces $x_r$, a second network predicts the target's segmentation mask $m_t$. Then  $(x_{r}^{\langle f \rangle},m_t)$ is passed to a third network that performs in-painting for occlusion correction. Finally a fourth network blends the corrected face into $x_t$ while considering ethnicity and lighting.
	Instead of using interpolation like \cite{nirkin2019fsgan}, the authors of \cite{li2019faceshifter} propose FaceShifter which uses novel Adaptive Attentional Denormalization layers (AAD) to transfer localized feature maps between the faces. \blue{In contrast to \cite{nirkin2019fsgan}, FaceShifter reduces the number of operations by handling the occlusions through a refinement network trained to consider the delta between the original $x_t$ and a reconstructed $\hat{x}_t$.}
	
	
\bgroup
\def\arraystretch{1}
\setlength\tabcolsep{.1em}
\scriptsize
\begin{table*}[t]
	\caption{Summary of Deep Learning Replacement Models}
	\label{tab:replacement}
	\centering
		\begin{tabular}{@{}ccccccccccccccccccccccccc@{}}
			\toprule
			&  &  & \multicolumn{3}{r}{Replacement} & \multicolumn{3}{c}{Retraining for new...} & \multicolumn{4}{c}{Model} & \multicolumn{3}{c}{Repr.} & \multicolumn{6}{c}{Model Training} & \multicolumn{2}{c}{Model Execution} & Model Outp. \\ \midrule
			&  &  &  & \cellcolor[HTML]{EFEFEF}\rotatebox[origin=c]{90}{Transfer\hspace{7em}} & \cellcolor[HTML]{EFEFEF}\rotatebox[origin=c]{90}{Swap\hspace{8em}} & \rotatebox[origin=c]{90}{Source ($s$)\hspace{6em}} & \rotatebox[origin=c]{90}{Target ($t$)\hspace{6.5em}} & \cellcolor[HTML]{C0C0C0}\rotatebox[origin=c]{90}{Identity Agnostic\hspace{3em}} & \rotatebox[origin=c]{90}{Encoders\hspace{7em}} & \rotatebox[origin=c]{90}{Decoders\hspace{7em}} & \rotatebox[origin=c]{90}{Discriminators\hspace{5em}} & \rotatebox[origin=c]{90}{Other Netw.\hspace{6em}} & \rotatebox[origin=c]{90}{3DMM/Rendering\hspace{3em}} & \rotatebox[origin=c]{90}{Segmentation\hspace{5em}} & \rotatebox[origin=c]{90}{Landmark / Keypoint \hspace{1.5em}} & \cellcolor[HTML]{EFEFEF}\rotatebox[origin=c]{90}{Labeling of: ID\hspace{4.5em}} & \cellcolor[HTML]{EFEFEF}\rotatebox[origin=c]{90}{Labeling of: Other\hspace{2.8em}} & \cellcolor[HTML]{DBDBDB}\rotatebox[origin=c]{90}{No Pairing\hspace{6.4em}} & \cellcolor[HTML]{DBDBDB}\rotatebox[origin=c]{90}{Paring within Same Video} & \cellcolor[HTML]{DBDBDB}\rotatebox[origin=c]{90}{Paring ID to Diffr. Actions} & \cellcolor[HTML]{DBDBDB}\rotatebox[origin=c]{90}{Requires Video\hspace{4.7em}} & \rotatebox[origin=c]{90}{Source ($x_s$...)\hspace{5em}} & \rotatebox[origin=c]{90}{Target ($x_t$...)\hspace{5em}} & \rotatebox[origin=c]{90}{Resolution\hspace{6.5em}} \\ \midrule\midrule
			& \cite{deepfake32:online} & 2017 & Deepfakes for All & \cellcolor[HTML]{EFEFEF} & \cellcolor[HTML]{EFEFEF}$\bullet$ & \multicolumn{2}{c}{2k-5k portraits} & \cellcolor[HTML]{C0C0C0} & 1 & 2 & 0-1 & 0 &  &  &  & \cellcolor[HTML]{EFEFEF}$\bullet$ & \cellcolor[HTML]{EFEFEF} & \cellcolor[HTML]{DBDBDB}$\bullet$ & \cellcolor[HTML]{DBDBDB} & \cellcolor[HTML]{DBDBDB} & \cellcolor[HTML]{DBDBDB}$\bullet$ & - & portrait & 256x256 \\
			& \cite{shaoanlu58:online} & 2018 & FaceSwap-GAN & \cellcolor[HTML]{EFEFEF} & \cellcolor[HTML]{EFEFEF}$\bullet$ & \multicolumn{2}{c}{2k-5k portraits} & \cellcolor[HTML]{C0C0C0} & 1 & 2 & 2 & 1 &  &  &  & \cellcolor[HTML]{EFEFEF}$\bullet$ & \cellcolor[HTML]{EFEFEF} & \cellcolor[HTML]{DBDBDB}$\bullet$ & \cellcolor[HTML]{DBDBDB} & \cellcolor[HTML]{DBDBDB} & \cellcolor[HTML]{DBDBDB}$\bullet$ & - & portrait & 256x256 \\
			\multirow{-3}{*}{One-to-One} & \cite{iperovDe6:online} & 2018 & DeepFaceLab & \cellcolor[HTML]{EFEFEF} & \cellcolor[HTML]{EFEFEF}$\bullet$ & \multicolumn{2}{c}{2k-5k portraits} & \cellcolor[HTML]{C0C0C0} & 1 & 2 & 0-1 & 0 &  &  &  & \cellcolor[HTML]{EFEFEF}$\bullet$ & \cellcolor[HTML]{EFEFEF} & \cellcolor[HTML]{DBDBDB}$\bullet$ & \cellcolor[HTML]{DBDBDB} & \cellcolor[HTML]{DBDBDB} & \cellcolor[HTML]{DBDBDB}$\bullet$ & - & portrait & 256x256 \\ \midrule
			One-to-Many & \cite{korshunova2017fast} & 2017 & Fast Face Swap & \cellcolor[HTML]{EFEFEF} & \cellcolor[HTML]{EFEFEF}$\bullet$ & 60 portraits & None & \cellcolor[HTML]{C0C0C0} & 0 & 0 & 0 & 2 &  & $\bullet$ & $\bullet$ & \cellcolor[HTML]{EFEFEF}$\bullet$ & \cellcolor[HTML]{EFEFEF} & \cellcolor[HTML]{DBDBDB}$\bullet$ & \cellcolor[HTML]{DBDBDB} & \cellcolor[HTML]{DBDBDB} & \cellcolor[HTML]{DBDBDB} & portrait & portrait & 256x256 \\ \midrule
			& \cite{natsume2018rsgan} & 2018 & RSGAN & \cellcolor[HTML]{EFEFEF} & \cellcolor[HTML]{EFEFEF}$\bullet$ & None & None & \cellcolor[HTML]{C0C0C0} & 4 & 3 & 2 & 1 &  & $\bullet$ &  & \cellcolor[HTML]{EFEFEF} & \cellcolor[HTML]{EFEFEF} & \cellcolor[HTML]{DBDBDB}$\bullet$ & \cellcolor[HTML]{DBDBDB} & \cellcolor[HTML]{DBDBDB} & \cellcolor[HTML]{DBDBDB} & portrait & portrait & 128x128 \\
			& \cite{natsume2018fsnet} & 2018 & FSNet & \cellcolor[HTML]{EFEFEF} & \cellcolor[HTML]{EFEFEF}$\bullet$ & None & None & \cellcolor[HTML]{C0C0C0} & 3 & 4 & 5 & 0 &  &  & $\bullet$ & \cellcolor[HTML]{EFEFEF}$\bullet$ & \cellcolor[HTML]{EFEFEF}$\bullet$ & \cellcolor[HTML]{DBDBDB}$\bullet$ & \cellcolor[HTML]{DBDBDB} & \cellcolor[HTML]{DBDBDB} & \cellcolor[HTML]{DBDBDB} & portrait & portrait & 128x128 \\
			& \cite{bao2018towards} & 2018 & OSIP-FS & \cellcolor[HTML]{EFEFEF} & \cellcolor[HTML]{EFEFEF}$\bullet$ & None & None & \cellcolor[HTML]{C0C0C0} & 2 & 1 & 2 & 0 &  &  &  & \cellcolor[HTML]{EFEFEF}$\bullet$ & \cellcolor[HTML]{EFEFEF} & \cellcolor[HTML]{DBDBDB}$\bullet$ & \cellcolor[HTML]{DBDBDB} & \cellcolor[HTML]{DBDBDB} & \cellcolor[HTML]{DBDBDB} & portrait & portrait & 128x128 \\
			& \cite{moniz2018unsupervised} & 2018 & DepthNets & \cellcolor[HTML]{EFEFEF}$\bullet$ & \cellcolor[HTML]{EFEFEF} & None & None & \cellcolor[HTML]{C0C0C0}$\bullet$ & 3 & 2 & 2 & 1 &  &  & $\bullet$ & \cellcolor[HTML]{EFEFEF}$\bullet$ & \cellcolor[HTML]{EFEFEF} & \cellcolor[HTML]{DBDBDB} & \cellcolor[HTML]{DBDBDB} & \cellcolor[HTML]{DBDBDB}$\bullet$ & \cellcolor[HTML]{DBDBDB} & portrait & portrait & 80x80 \\
			& \cite{nirkin2019fsgan} & 2018 & FSGAN & \cellcolor[HTML]{EFEFEF} & \cellcolor[HTML]{EFEFEF}$\bullet$ & None & None & \cellcolor[HTML]{C0C0C0}$\bullet$ & 4 & 4 & 3 & 1 &  & $\bullet$ & $\bullet$ & \cellcolor[HTML]{EFEFEF}$\bullet$ & \cellcolor[HTML]{EFEFEF} & \cellcolor[HTML]{DBDBDB} & \cellcolor[HTML]{DBDBDB} & \cellcolor[HTML]{DBDBDB}$\bullet$ & \cellcolor[HTML]{DBDBDB}$\bullet$ & portrait & portrait & 256x256 \\
			& \cite{sun2018hybrid} & 2018 & IO-FR & \cellcolor[HTML]{EFEFEF} & \cellcolor[HTML]{EFEFEF}$\bullet$ & None & None & \cellcolor[HTML]{C0C0C0}$\bullet$ & 1 & 1 & 1 & 1 & $\bullet$ &  &  & \cellcolor[HTML]{EFEFEF}$\bullet$ & \cellcolor[HTML]{EFEFEF} & \cellcolor[HTML]{DBDBDB}$\bullet$ & \cellcolor[HTML]{DBDBDB} & \cellcolor[HTML]{DBDBDB} & \cellcolor[HTML]{DBDBDB} & portrait & portrait & 256x256 \\
			& \cite{shaoanlu29:online} & 2019 & FS Face Trans. & \cellcolor[HTML]{EFEFEF} & \cellcolor[HTML]{EFEFEF}$\bullet$ & None & None & \cellcolor[HTML]{C0C0C0}$\bullet$ & 1 & 1 & 2 & 2 &  & $\bullet$ &  & \cellcolor[HTML]{EFEFEF}$\bullet$ & \cellcolor[HTML]{EFEFEF} & \cellcolor[HTML]{DBDBDB}$\bullet$ & \cellcolor[HTML]{DBDBDB} & \cellcolor[HTML]{DBDBDB} & \cellcolor[HTML]{DBDBDB} & portraits & portrait & 128x128 \\
			& \cite{xiao2019identity} & 2019 & IHPT & \cellcolor[HTML]{EFEFEF}$\bullet$ & \cellcolor[HTML]{EFEFEF} & None & None & \cellcolor[HTML]{C0C0C0}$\bullet$ & 2 & 1 & 2 & 0 &  &  &  & \cellcolor[HTML]{EFEFEF} & \cellcolor[HTML]{EFEFEF}$\bullet$ & \cellcolor[HTML]{DBDBDB} & \cellcolor[HTML]{DBDBDB}$\bullet$ & \cellcolor[HTML]{DBDBDB} & \cellcolor[HTML]{DBDBDB}$\bullet$ & cropped & cropped & 128x128 \\
			\multirow{-9}{*}{Many-to-Many} & \cite{li2019faceshifter} & 2019 & FaceShifter & \cellcolor[HTML]{EFEFEF} & \cellcolor[HTML]{EFEFEF}$\bullet$ & None & None & \cellcolor[HTML]{C0C0C0}$\bullet$ & 3 & 3 & 3 & 0 &  &  &  & \cellcolor[HTML]{EFEFEF}$\bullet$ & \cellcolor[HTML]{EFEFEF} & \cellcolor[HTML]{DBDBDB}$\bullet$ & \cellcolor[HTML]{DBDBDB} & \cellcolor[HTML]{DBDBDB} & \cellcolor[HTML]{DBDBDB} & portrait & portrait & 256x256 \\ \bottomrule\bottomrule
		\end{tabular}%
	\vspace{-1.5em}
\end{table*}

\egroup
		
	\subsubsection{\textbf{Few-Shot Learning}}
	The same author of FaceSwap-GAN \cite{shaoanlu58:online} also hosts few-shot approach online dubbed ``One Model to Swap Them All'' \cite{shaoanlu29:online}. In this version the generator receives $(x_{s}^{\langle f \rangle},x_{t}^{\langle f \rangle},m_t)$ where its encoder is conditioned on VGGFace2 features of $x_t$ using FC-AdaIN layers, and its decoder is conditioned on $x_t$ and the face structure $m_t$ via layer concatenations and SPADE-ResBlocks respectively. Two discriminators are used: one on image quality given the face segmentation and the other on the identities. 

	
	\captionsetup[subfigure]{labelfont=bf,textfont=normalfont,singlelinecheck=off,justification=raggedright}

	\renewcommand{\thesubfigure}{\arabic{subfigure}}

	\subsection{Transfer}
	Although face transfers precede face swaps, today there are very few works that use deep learning for this task. However, we note that face a transfer is equivalent to performing \textit{self-reenactment} on a face swapped portrait. Therefore, high quality face a transfers can be achieved by combining a method from Section \ref{subsec:expression} and Section \ref{subsec:swap}.

	In 2018, the authors of \cite{moniz2018unsupervised} proposed DepthNets: an unsupervised network for capturing facial landmarks and translating the pose from one identity to another. The authors use a Siamese network to predict a transformation matrix that maps the $x_s$'s 3D facial landmarks to the corresponding 2D landmarks of $x_t$. A 3D renderer (OpenGL) is then used to warp $x_{s}^{\langle f \rangle}$ to the source pose $l_t$, and the composition is refined using a CycleGAN. \blue{Since warping is involved, the approach is sensitive to occlusions.}

	Later in 2019, the authors of \cite{xiao2019identity} proposed a self-supervised network which can change the identity of an object within an image. Their ED disentangles the identity from an objects pose using a novel disentanglement loss. Furthermore to handle misaligned poses, an L1 loss is computed using a pixel mapped version of $x_g$ to $x_s$ (using the weights of the identity encoder). Similarly, the authors of \cite{li2019mixnmatch} proposed a method disentangled identity transfer. However neither \cite{xiao2019identity} or \cite{li2019mixnmatch} were explicitly performed on faces.


	\section{Countermeasures}\label{sec:countermeasures}

In general, countermeasures to malicious deepfakes can be categorized as either detection or prevention. We will now briefly discuss each accordingly.
A summary and systematization of the deepfake detection methods can be found in Table \ref{tab:detection}.

\subsection{Detection} 
The subject of image forgery detection is a well researched subject \cite{zheng2019survey}.
In our review of detection methods, we will focus on works which specifically deal with detecting deepfakes of humans.

\subsubsection{Artifact-Specific}
Deepfakes often generate artifacts which may be subtle to humans, but can be easily detected using machine learning and forensic analysis. Some works identify deepfakes by searching for specific artifacts. We identify seven types of artifacts: Spatial artifacts in blending, environments, and forensics; temporal artifacts in behavior, physiology, synchronization, and coherence.

\vspace{.5em}\noindent\textbf{Blending (\textit{spatial}).}
Some artifacts appear where the generated content is blended back into the frame. To help emphasize these artifacts to a learner, researchers have proposed edge detectors, quality measures, and frequency analysis \cite{agarwal2017swapped,zhang2017automated,akhtarcomparative,mo2018fake,durall2019unmasking}. In 	\cite{li2020face} the authors \blue{follow a more explicit approach to detecting the boundary.} They trained a CNN network to predict an image's blending boundary and a label (real or fake). Instead of using a deepfake dataset, the authors trained their network on a dataset of face swaps generated by splicing similar faces found through facial landmark similarity. \blue{By doing so, the model has the advantage that is focuses on the blending boundary and not other artifacts caused by the generative model.}

\vspace{.5em}\noindent\textbf{Environment (\textit{spatial}).}
The content of a fake face can be anomalous in context to the rest of the frame. For example, residuals from face warping processes \cite{li2019exposing,danmohah54:online,li2019celeb}, lighting \cite{straub2019using}, and varying fidelity \cite{korshunov2018deepfakes} can indicate the presence of generated content.
In \cite{li2020fighting}, the authors follow a different approach by contrasting the generated foreground to the (untampered) background using a patch and pair CNN. \blue{The authors of \cite{nirkin2020deepfake} also contrast the fore/background but enable a network to identify the distinguishing features automatically.}
They accomplish this by (1) encoding the face and context (hair and background) with an ED and (2) passing the difference between the encodings with the complete image (encoded) to a classifier.

\vspace{.5em}\noindent\textbf{Forensics (\textit{spatial}).}
Several works detect deepfakes by analyzing subtle features and patterns left by the model. 
In \cite{yu2019attributing} and \cite{marra2019gans}, the authors found that GANs leave unique fingerprints and show how it is possible to classify the generator given the content, even in the presence of compression and noise. In \cite{koopman2018detection} the authors analyze a camera's unique sensor noise (PRNU) to detect pasted content.
To focus on the residuals, the authors of \cite{masi2020two} use a two stream ED to encode the color image and a frequency enhanced version using
``Laplacian of Gaussian layers'' (LoG). The two encodings are then fed through an LSTM which then classifies the video based on a sequence of frames. 

\blue{Instead of searching for residuals, the authors of \cite{yang2019exposing} search for imperfections} and found that deepfakes tend to have inconsistent head poses. Therefore, they detect deepfakes by predicting and monitoring facial landmarks. 
The authors of \cite{wang2020cnn} had a different approach by training classifier to focus on the imperfections instead of the residuals. This was accomplished by using a dataset generated using a ProGAN instead of other GANs since the ProGAN's images contain the least amount of frequency artifacts. 
In contrast to \cite{wang2020cnn}, the authors in \cite{guo2020fake} use a network to \textit{emphasize} the residuals and suppress the imperfections in a preprocessing step for a classifier. Their network uses adaptive convolutional layers that predict residuals to maximize the artifacts' influence. \blue{Although this approach may help the network identify artifacts better, it may not generalize as well to new types of artifacts.}

\vspace{.5em}\noindent\textbf{Behavior (\textit{temporal}).}
With large amounts of data on the target, mannerisms and other behaviors can be monitored for anomalies. For example, in \cite{agarwal2019protecting} the authors protect world leaders from a wide variety of deepfake attacks by modeling their recorded stock footage. \blue{Recently, the authors of \cite{mittal2020emotions} showed how behavior can be used with no reference footage of the target.} The approach is to detect discrepancies in the perceived emotion extracted from the clip's audio and video content. The authors use a custom Siamese network to consider the audio and video emotions when contrasted to real and fake videos.

\vspace{.5em}\noindent\textbf{Physiology (\textit{temporal}).}
In 2014, researchers hypothesized that generated content will lack physiological signals and identified computer generated faces by monitoring their heart rate \cite{conotter2014physiologically}. Regarding deepfakes, \cite{ciftci2019fakecatcher} monitored blood volume patterns (pulse) under the skin, and \cite{li2018ictu} \blue{took a more robust approach} by monitoring irregular eye blinking patterns. 
Instead of detecting deepfakes, the authors of \cite{ciftci2020hearts} use the pulse signal to help determine the model used to create the deepfake. 

\vspace{.5em}\noindent\textbf{Synchronization (\textit{temporal}).}
Inconsistencies are also a revealing factor. In \cite{korshunov2018speaker} and \cite{korshunov2019tampered}, the authors noticed that video dubbing attacks can be detected my correlating the speech to landmarks around the mouth. Later, in \cite{agarwal2020detecting}, the authors \blue{refined the approach} by detecting when visemes (mouth shapes) are inconsistent with the spoken phonemes (utternaces). In particular, they focus on phonemes where the mouth is fully closed (B, P, M) since deepfakes in the wild tend to fail in generating these visemes.

\vspace{.5em}\noindent\textbf{Coherence (\textit{temporal}).}
As noted in Section \ref{subsec:expression}, realistic temporal coherence is challenging to generate, and some authors capitalize on the resulting artifacts to detect the fake content. For example, \cite{guera2018deepfake} uses an RNN to detect artifacts such as flickers and jitter, and \cite{sabir2019recurrent} uses an LSTM on the face region only. In \cite{chan2019everybody} a classifier is trained pairs of sequential frames and in \cite{amerini2019deepfake} the authors \blue{refine the network's focus} by monitoring the frames' optical flow. Later the same authors use an LSTM to predict the next frame, and expose deepfakes when the reconstruction error is high \cite{amerini2020exploiting}. \looseness=-1

\subsubsection{Undirected Approaches} 
Instead of focusing on a specific artifact, some authors train deep neural networks as generic classifiers, and let the network decide which features to analyze. In general, researchers have taken one of two approaches: classification or anomaly detection.

\vspace{.5em}\noindent\textbf{Classification.}
In \cite{marra2018detection,rossler2019faceforensics++,nguyen2019capsule}, it was shown that deep neural networks tend to perform better than traditional image forensic tools on compressed imagery. Various authors then demonstrated how standard CNN architectures can effectively detect deepfake videos \cite{afchar2018mesonet,do2018forensics,tariq2018detecting,ding2019swapped}. In \cite{hsu2020deep}, the authors train the CNN as a Siamese network using contrasting examples of real and fake images. 	
\blue{In \cite{fernando2019exploiting}, the authors were concerned that a CNN can only detect the attacks on which they trained. To close this gap,} the authors propose using Hierarchical Memory Network (HMN) architecture which considers the contents of the face and previously seen faces. The network encodes the face region which is then processed using a bidirectional GRU while applying an attention mechanism. The final encoding is then passed to a memory module, which compares it to recently seen encodings and makes a prediction.
Later, in \cite{rana2020deepfakestack}, the authors use an ensemble approach and leverage the predictions of seven deepfake CNNs by passing their predicitons to a meta classifer. \blue{Doing so produces results which are more robust (fewer false positives) than using any single model.}
In \cite{de2020deepfake}, the authors 
tried a variety of different classic spatio-temproal networks and feature extractors as a baseline for temporal deepfake detection. They found that a 3D CNN, which looks at multiple frames at once, out performs both recurrent networks and a the state of the art ID3 architecture. 

To localize the tampered areas, some works train networks to predicting masks learned from a ground truth dataset, or by mapping the neural activations back to the raw image   \cite{nguyen2019multi,du2019towards,stehouwer2019detection,li2019zooming}. 

In general, we note that the use of classifiers to detect deepfakes is problematic since an attacker can evade detection via adversarial machine learning. We will discuss this issue further in Section \ref{subsec:arms_race}.

\bgroup
\def\arraystretch{.8}
\setlength\tabcolsep{.1em}
\begin{table*}[t]
	\centering
	\caption{Summary of Deepfake Detection Models}
	\vspace{-.5em}
	\label{tab:detection}
	\scriptsize	
	\resizebox{.8\textwidth}{!}{%
\begin{tabular}{@{}cccccccccccccccccccccccccc@{}}
			\toprule
			&  &  & \multicolumn{2}{c}{Type} & \multicolumn{3}{c}{Modality} & \multicolumn{4}{c}{Content} & \multicolumn{3}{c}{Method} & \multicolumn{8}{c}{Eval. Dataset} & \multicolumn{3}{c}{Performance*} \\ \midrule
			&  &  & \cellcolor[HTML]{EFEFEF}\rotatebox[origin=c]{90}{Reenactment} & \cellcolor[HTML]{EFEFEF}\rotatebox[origin=c]{90}{Replacement} & \cellcolor[HTML]{DBDBDB}\rotatebox[origin=c]{90}{Image} & \cellcolor[HTML]{DBDBDB}\rotatebox[origin=c]{90}{Video} & \cellcolor[HTML]{DBDBDB}\rotatebox[origin=c]{90}{Audio} & \cellcolor[HTML]{C0C0C0}\rotatebox[origin=c]{90}{Feature} & \cellcolor[HTML]{C0C0C0}\rotatebox[origin=c]{90}{Body Part} & \cellcolor[HTML]{C0C0C0}\rotatebox[origin=c]{90}{Face} & \cellcolor[HTML]{C0C0C0}\rotatebox[origin=c]{90}{Image} & \rotatebox[origin=c]{90}{Model} & \rotatebox[origin=c]{90}{Indicates Affected Area} & \rotatebox[origin=c]{90}{Input Resolution} & \cellcolor[HTML]{EFEFEF}\rotatebox[origin=c]{90}{DeepfakeTIMIT \cite{korshunov2018deepfakes}} & \cellcolor[HTML]{EFEFEF}\rotatebox[origin=c]{90}{DFFD \cite{stehouwer2019detection}} & \cellcolor[HTML]{EFEFEF}\rotatebox[origin=c]{90}{FaceForensics \cite{rossler2018faceforensics}} & \cellcolor[HTML]{EFEFEF}\rotatebox[origin=c]{90}{FaceForensics++ \cite{rossler2019faceforensics++}} & \cellcolor[HTML]{EFEFEF}\rotatebox[origin=c]{90}{FFW \cite{khodabakhsh2018fake}} & \cellcolor[HTML]{EFEFEF}\rotatebox[origin=c]{90}{Celeb-DF \cite{li2019celeb}} & \cellcolor[HTML]{EFEFEF}\rotatebox[origin=c]{90}{Other Deepfake DB} & \cellcolor[HTML]{EFEFEF}\rotatebox[origin=c]{90}{Custom} & \rotatebox[origin=c]{90}{ACC} & \rotatebox[origin=c]{90}{EER} & \rotatebox[origin=c]{90}{AUC} \\
\end{tabular}%
	}
\vspace{-2em}
\end{table*}

\egroup

\vspace{.5em}\noindent\textbf{Anomaly Detection.}
In contrast to classification, anomaly detection models are trained on the normal data and then detect outliers during deployment. \blue{By doing so, these methods do not make assumptions on how the attacks look and thus generalize better to unknown creation methods.} The authors of \cite{wang2019fakespotter} follow this approach by measuring the neural activation (coverage) of a face recognition network. \blue{By doing so, the model is able to overcome noise and other distortions, by obtaining a stronger signal from than just using the raw pixels.} Similarly, in \cite{khalid2020oc} a one-class VAE is trained to used to reconstruct real images. Then, for new images, an anomaly score is computed by taking the MSE between mean component of the encoded image and the mean component of the reconstructed image.
Alternatively, the authors of \cite{bao2018towards} measure an input's embedding distance to real samples using an ED's latent space. \blue{The difference between these works is that \cite{wang2019fakespotter} and \cite{khalid2020oc} rely on a model's inability to process unknown patterns while \cite{bao2018towards} contrasts the model's representations.} 

Instead of using a neural network directly, the authors of \cite{fernandes2020detecting} use a state of the art attribution based confidence metric (ABC). To detect a fake image, the ABC is used to determine if the image fits the training distribution of a pretrained face recognition network (e.g., VGG).


\subsection{Prevention \& Mitigation} 
\vspace{.5em}\noindent\textbf{Data Provenance.}
To prevent deepfakes, some have suggested that data provenance of multimedia should be tracked through distributed ledgers and blockchain networks \cite{fraga2019leveraging}. In \cite{chen2019trusting} the authors suggest that the content should be ranked by participants and AI. In contrast, \cite{hasan2019combating} proposes that the content should authenticated and managed as a global file system over Etherium smart contracts. 

\vspace{.5em}\noindent\textbf{Counter Attacks.}
To combat deepfakes, the authors of \cite{li2019hiding} show how adversarial machine learning can be used to disrupt and corrupt deepfake networks. The authors perform adversarial machine learning to add crafted noise perturbations to $x$, which prevents deepfake technologies from locating a proper face in $x$. In a different approach, the authors of \cite{shan2020fawkes} use adversarial noise to change the identity of the face so that web crawlers will not be able find the image of $t$ to train their model.

	\section{Discussion}\label{sec:discussion}
	
		\subsection{The Creation of Deepfakes}
		
		\subsubsection{Trade-offs Between the Methodologies.}
		In general, there is a different cost and payoff for each deepfake creation method. However, the most effective and threatening deepfakes are those which are (1) the most practical to implement [Training \textit{Data}, Execution \textit{Speed}, and \textit{Accessibility}] and (2) are the most believable to the victim [\textit{Quality}]:
		
		\begin{description}[leftmargin=.5cm]
			\item[Data vs Quality.]  Models trained on numerous samples of the target often yield better results (e.g., \cite{chan2019everybody, fried2019text, iperovDe6:online, jalalifar2018speech, kumar2017obamanet, liu2019video, suwajanakorn2017synthesizing, wu2018reenactgan}). For example, in 2017, \cite{suwajanakorn2017synthesizing} produced an extremely believable reenactment of Obama which exceeds the quality of recent works. However, these models require many hours footage for training, and are therefore are only suitable for exposed targets such as actors, CEOs, and political leaders. An attacker who wants to commit defamation, impersonation, or a scam on an arbitrary individual will need to use a many-to-many or few-shot approach. On the other hand, most of these methods rely on a single reference of $t$ and are therefore prone to generating artifacts. This is because the model must `imagine' missing information (e.g., different poses and occlusions). Therefore, approaches which provide the model with a limited number of reference samples \cite{gu2020flnet, MarioNETte:AAAI2020, tran2018representation, wang2019fewshotvid2vid, wiles2018x2face, yu2019improving, zakharov2019few} strike the best balance between data and quality.
			
			\item[Speed vs Quality.] The trade-off between these aspects depends on whether the attack is online (interactive) or offline (stored media). Social engineering attacks involving deepfakes are likely to be online and thus require real-time speeds. However, high resolution models have many parameters and sometimes use several networks (e.g., \cite{fu2019high}) and some process multiple frames to provide temporal coherence (e.g., \cite{bansal2018recycle,kim2018deep,wang2018vid2vid}). Other methods may be slowed down due to their pre/post-processing steps, such as warping \cite{geng2019warp,gu2020flnet,zhang2019one}, UV mapping or segmentation prediction \cite{cao20193d,olszewski2017realistic,nagano2018pagan,yu2019improving}, and the use of refinement networks \cite{chan2019everybody,geng2019warp,kim2018deep,li2019faceshifter,moniz2018unsupervised,thies2019neural}. To the best of our knowledge, \cite{jamaludin2019you,nirkin2019fsgan,nagano2018pagan,korshunova2017fast} and \cite{NIPS2019_8935} are the only papers which claim to generate real time deepfakes, yet they subjectively tend to be blurry or distort the face. Regardless, a victim is likely fall for an imperfect deepfake in a social engineering attack when placed under pressure in a false pretext \cite{workman2008wisecrackers}. Moreover, it is likely that an attacker will implement a complex method at a lower resolution to speed up the frame rate. In which case, methods that have texture artifacts would be preferred over those which produce shape or identity flaws (e.g., \cite{NIPS2019_8935} vs \cite{zakharov2019few}). For attacks that are not real-time (e.g, fake news), resolution and fidelity is critical. In these cases, works that produce high quality images and videos with temporal coherence are the best candidates (e.g., \cite{MarioNETte:AAAI2020,wang2018vid2vid}). 

			\item[Availability vs Quality.] We also note that availability and reproducibility are key factors in the proliferation of new technologies. Works that publish their code and datasets online (e.g., \cite{wiles2018x2face, wu2018reenactgan, sanchez2018triple, tulyakov2018mocogan,kefalas2019speech,NIPS2019_8935}) are more likely to be used by researchers and criminals compared to those which are unavailable \cite{kim2018deep, pham2018generative,zhang2019faceswapnet,wang2020imaginator,MarioNETte:AAAI2020, fried2019text,thies2019neural,aberman2019deep,nirkin2019fsgan} or require highly specific or private datasets \cite{yu2019improving,ganin2016deepwarp,nagano2018pagan}. This is because the payoff in implementing a paper is minor compared to using a functional and effective method available online. Of course, this does not include state-actors who have plenty of time and funding.
		\end{description}
	
		We have also observed that approaches which augment a network's inputs with synthetic ones produce better results in terms of quality and stability. For example, by rotating limbs \cite{liu2019video,zhou2019dance}, refining rendered heads \cite{nagano2018pagan,fried2019text,yu2019mining,balakrishnan2018synthesizing,thies2019neural,wang2018high}, providing warped imagery \cite{zablotskaia2019dwnet,neverova2018dense,moniz2018unsupervised,geng2019warp} and UV maps \cite{cao20193d,kim2018deep,otberdout2019dynamic,zablotskaia2019dwnet,gu2020flnet}. This is because the provided contextual information reduces the problem's complexity for the neural network.

		Given these considerations, in our opinion, the most significant and available deepfake technologies today are \cite{NIPS2019_8935} for facial reenactment because of it's efficiency and practicality; \cite{chen2019hierarchical} for mouth reenactment because of its quality; and \cite{iperovDe6:online} for face replacement because its high fidelity and wide spread use. However, this is a subjective opinion based on the samples provided online and in the respective papers. A comparative research study, where the methods are trained on the same dataset and evaluated by a number of people is necessary to determine the best quality deepfake in each category.

		\subsubsection{Research Trends}
		Over the last few years there has been a shift towards identity agnostic models and high resolution deepfakes. Some notable advancements include (1) unpaired self-supervised training techniques to reduce the amount of initial training data, (2) one/few-shot learning which enables identity theft with a single profile picture, (3) improvements of face quality and identity through AdaIN layers, disentanglement, and pix2pixHD network components, (4) fluid and realistic videos through temporal discriminators and optical flow prediction, and (5) the mitigation of boundary artifacts by using secondary networks to blend composites into seamless imagery (e.g., \cite{wang2018high,fried2019text,thies2019neural}). 
		\looseness=-1
				
		Another large advancement in this domain was the use of perceptual loss on a pre-trained VGG Face recognition network. The approach boosts the facial quality significantly, and as a result, has been adopted in popular online deepfake tools \cite{deepfake32:online,shaoanlu58:online}. 
		Another advancement being adopted is the use of a network pipeline. Instead of enforcing a set of global losses on a single network, a pipeline of networks is used where each network is tasked with a different responsibility (conversion, generation, occlusions, blending, etc.) This give more control over the final output and has been able to mitigate most of the challenges mention in Section \ref{subsec:challenges}.
		
		\subsubsection{Current Limitations}
		Aside from quality, there are a few limitations with the current deepfake technologies. First, for reenactment, content is always driven and generated with a frontal pose. This limits the reenactment to a very static performance. Today, this is avoided by face swapping the identity onto a lookalike's body, but a good match is not always possible and this approach has limited flexibility. Second, reenactments and replacements depend on the driver's performance to deliver the identity's personality. We believe that next generation deepfakes will utilize videos of the target to stylize the generated content with the expected expressions and mannerisms. This will enable a much more automatic process of creating believable deepfakes. Finally, a new trend is real-time deepfakes. Works such as \cite{jamaludin2019you,nirkin2019fsgan} have achieved real-time deepfakes at 30fps. Although real-time deepfakes are an enabler for phishing attacks, the realism is not quite there yet. Other limitations include the coherent rendering of hair, teeth, tongues, shadows, and the ability to render the target's hands (especially when touching the face).
		Regardless, deepfakes are already very convincing \cite{rossler2019faceforensics++} and are improving at a rapid rate. Therefore, it is important that we focus on effective countermeasures.

		\subsection{The Deepfake Arms Race}\label{subsec:arms_race}
		Like any battle in cyber security, there is an arms race between the attacker and defender. In our survey, we observed that the majority deepfake detection algorithms assume a static game with the adversary: They are either focused on identifying a specific artifact, or do not generalize well to new distributions and unseen attacks \cite{cozzolino2018forensictransfer}.  
		Moreover, based on the recent benchmark of \cite{li2019celeb}, we observe that the performance of state-of-the-art detectors are decreasing rapidly as the quality of the deepfakes improve. Concretely, the three most recent benchmark datasets (DFD by Google \cite{GoogleAI34:online}, DFDC by Facebook \cite{dolhansky2019deepfake}, and Celeb-DF by \cite{li2019celeb}) were released within one month of each other at the end of 2019. However, the deepfake detectors only achieved an AUC of $0.86$,$0.76$, and $0.66$ on each of them respectively. Even a false alarm rate of $0.001$ is far too low considering the millions of images published online daily. 
		\looseness=-1

		\vspace{.5em}\noindent\textbf{Evading Artifact-based Detectors.} 
		To evade an artifact-based detector, the adversary only needs to mitigate a single flaw to evade detection. For example, $G$ can generate the biological signals monitored by \cite{li2018ictu,ciftci2019fakecatcher} by adding a discriminator which monitors these signals. To avoid anomalies in extensive the neuron activation \cite{wang2019fakespotter}, the adversary can add a loss which minimizes neuron coverage. Methods which detect abnormal poses and mannerisms \cite{agarwal2019protecting} can be evaded by reenacting the entire head and by learning the mannerisms from the same databases. Models which identify blurred content \cite{mo2018fake} are affected by noise and sharpening GANs \cite{jalalifar2018speech,kim2016accurate}, and models which search for the boundary where the face was blended in \cite{li2019face,agarwal2017swapped,zhang2017automated,akhtarcomparative,mo2018fake,durall2019unmasking} do not work on deepfakes passed through refiner networks, which use in-painting, or those which output full frames (e.g., \cite{nagano2018pagan,kim2018deep,zhou2019dance,natsume2018fsnet,nirkin2019fsgan,li2019faceshifter,liu2019neural,zablotskaia2019dwnet}). Finally, solutions which search for forensic evidence \cite{yu2019attributing,marra2019gans,koopman2018detection} can be evaded (or at least raise the false alarm rate) by passing $x_g$ through filters, or by performing physical replication or compression. 
		
		\vspace{.5em}\noindent\textbf{Evading Deep Learning Classifiers.}
		There are a number of detection methods which apply deep learning directly to the task of deepfake detection (e.g., \cite{afchar2018mesonet,do2018forensics,tariq2018detecting,ding2019swapped,fernando2019exploiting}). However, an adversary can use adversarial machine learning to evade detection by adding small perturbations to $x_g$. Advances in adversarial machine learning has shown that these attacks transfer across multiple models regardless of the training data used \cite{papernot2016transferability}. Recent works have shown how these attacks not only work on deepfakes classifiers \cite{neekhara2020adversarial} but also work with no knowledge of the classifier or it's training set \cite{carlini2020evading}.
		\looseness=-1

		\vspace{.5em}\noindent\textbf{Moving Forward.}
		Nevertheless, deepfakes are still imperfect, and these methods offer a modest defense for the time being. Furthermore, these works play an important role in understanding the current limitations of deepfakes, and raise the difficulty threshold for malicious users. At some point, it may become too time-consuming and resource-intensive a common attacker to create a good-enough fake to evade detection. However, we argue that solely relying on the development of content-based countermeasures is not sustainable and may lead to a reactive arms-race.
		Therefore, we advocate for more out-of-band approaches for detecting a preventing deepfakes. For example, the establishment of content provenance and authenticity frameworks for online videos \cite{fraga2019leveraging,chen2019trusting,hasan2019combating}, and proactive defenses such as the use of adversarial machine learning to protect content from tampering \cite{li2019hiding}. \looseness=-1
		
		 
		\vspace{-.3em}
		\subsection{Deepfakes in other Domains}\label{subsec:df_otherdomains}
		In this survey, we put a focus on human reenactment and replacement attacks; the type of deepfakes which has made the largest impact so far \cite{hall2018deepfake,antinori2019terrorism}. However, deepfakes extend beyond human visuals, and have spread many other domains. In healthcare, the authors of \cite{mirsky2019ct} showed how deepfakes can be used to inject tor remove medical evidence in CT and MRI scan for insurance fraud, disruption, and physical harm. In \cite{jia2018transfer} it was shown how one's voice can be cloned with only five seconds of audio, and in Sept. 2019 a CEO was scammed out of \$250K via a voice clone deepfake \cite{AVoiceDe76:online}. The authors of \cite{bontrager2018deepmasterprints} have shown how deep learning can generate realistic human fingerprints that can unlock multiple users' devices. In \cite{schreyer2019adversarial} it was shown how deepfakes can be applied to financial records to evade the detection of auditors. Finally, it has been shown how deepfakes of news articles can be generated \cite{NIPS2019_9106} and that deepfake tweets exist as well \cite{fagni2020tweepfake}.		
		
		These examples demonstrate that deepfakes are not just attack tools for misinformation, defamation, and propaganda, but also sabotage, fraud, scams, obstruction of justice, and potentially many more.\looseness=-1

		\subsection{What's on the Horizon}
		We believe that in the coming years, we will see more deepfakes being weaponized for monetization. The technology has proven itself in humiliation, misinformation, and defamtion attacks. Moreover, the tools are becoming more practical \cite{deepfake32:online} and efficient \cite{jia2018transfer}. Therefore, is seems natural that malicious users will find ways to use the technology for a profit. As a result, we expect to see an increase in deepfake phishing attacks and scams targeting both companies and individuals.
		
		As the technology matures, real-time deepfakes will become increasingly realistic. Therefore, we can expect that the technology will be used by hacking groups to perform reconnaissance as part of an APT, and by state actors to perform espionage and sabotage by reenacting of officials or family members.
		
		To keep ahead of the game, we must be proactive and consider the adversary's next step, not just the weaknesses of the current attacks. We suggest that more work be done on evaluating the theoretical limits of these attacks. For example, by finding a bound on a model's delay can help detect real-time attacks such as \cite{jia2018transfer}, and determining the limits of GANs like \cite{agarwal2019limits} can help us devise the appropriate strategies. As mentioned earlier, we recommend further research on solutions which do not require analyzing the content itself. 
		Moreover, we believe it would be beneficial for future works to explore the weaknesses and limitations of current deepfakes detectors. By identifying and understanding these vulnerabilities, researchers will be able to develop stronger countermeasures.

	\section{Conclusion}

	Not all deepfakes are malicious. However, because the technology makes it so easy to create believable media, malicious users are exploiting it to perform attacks. These attacks are targeting individuals and causing psychological, political, monetary, and physical harm. As time goes on, we expect to see these malicious deepfakes spread to many other modalities and industries.
	\looseness=-1
	
	In this survey we focused on reenactment and replacement deepfakes of humans. We provided a deep review of how these technologies work, the differences between their architectures, and what is being done to detect them. We hope this information will be helpful to the community in understanding and preventing malicious deepfakes.
	
	\bibliographystyle{ACM-Reference-Format}
	\bibliography{paper}

\end{document}